%% file: iclr2025_conference.tex
\documentclass{article} 
\usepackage{iclr2025_conference,times}

\input{math_commands.tex}

\usepackage{algorithm}
\usepackage{algpseudocode}
\usepackage{amsmath}

\usepackage{hyperref}
\usepackage{url}

\usepackage{comment}
\usepackage{caption}

\usepackage{booktabs}        
\usepackage{multirow}        
\usepackage{graphicx}        
\usepackage{array}

\usepackage{colortbl}
\usepackage{pifont}
\newcommand{\cmark}{\ding{51}}  
\newcommand{\xmark}{\ding{55}}  
\usepackage{varwidth} 

\usepackage{xcolor}
\definecolor{lightblue}{RGB}{204,229,255}
\definecolor{lightgreen}{RGB}{204,255,204}

\usepackage[framemethod=TikZ]{mdframed}

\newcounter{keyfindingcounter}

\newcommand{\keyfindingbox}[1]{%
  \begin{mdframed}[
    linecolor=green!60,
    backgroundcolor=green!3,
    roundcorner=10pt,
    innertopmargin=10pt,
    innerbottommargin=10pt,
    innerleftmargin=10pt,
    innerrightmargin=10pt,
    skipabove=\topsep,
    skipbelow=\topsep
  ]
  \stepcounter{keyfindingcounter}%
  \textbf{Key Finding \thekeyfindingcounter:} #1
  \end{mdframed}
}

\definecolor{ourscolor}{RGB}{28, 128, 62}     
\definecolor{baselinecolor}{RGB}{192, 57, 43} 
\definecolor{errorcolor}{RGB}{211, 47, 47}    
\definecolor{rewardgreen}{RGB}{28, 128, 62} 
\newcommand{\rewardkeyword}[1]{\textcolor{rewardgreen}{\textbf{#1}}}

\title{Incentivizing Consistent, Effective and Scalable Reasoning Capability in Audio LLMs via Reasoning Process Rewards}

\author{Jiajun Fan$^{1,2}$\thanks{Work done as intern at Amazon} , Roger Ren$^1$, Jingyuan Li$^1$, Rahul Pandey$^1$, Prashanth Gurunath Shivakumar$^1$ \AND Ivan Bulyko$^1$, Ankur Gandhe$^1$, Ge Liu$^2$,  Yile Gu$^1$ \\ \\
$^1$ Amazon\\
$^1$ \{rogerren, jylii, rpandyn, psshvak,   ibbulyko, aggandhe, yilegu\}@amazon.com \\
$^2$ Siebel School of Computing and Data Science, University of Illinois Urbana-Champaign\\ 
$^2$ \{jiajunf3, geliu\}@illinois.edu 
\vspace{-0.15in} 
}

\iclrfinalcopy 
\begin{document}

\maketitle
\begin{abstract}
The role of reasoning in Audio Large Language Models remains widely underexplored, as introducing a reasoning process often degrades rather than improves performance during inference, a phenomenon we term test-time inverse scaling, where longer reasoning chains yield progressively worse results. We demonstrate that this stems not from fundamental limitations of reasoning itself, but from inadequate training: models without proper guidance for the reasoning process produce hallucinatory, inconsistent reasoning that accumulates errors over longer chains. To address these challenges, we introduce CESAR (Consistent, Effective, and Scalable Audio Reasoners), shifting from outcome verification to rewarding the reasoning process. Our online reinforcement learning framework employs Group Relative Policy Optimization with a multi-faceted reward suite that incentivizes not only correctness and format but also consistency, structured analytical patterns, causal reasoning, domain-knowledge integration, and calibrated reasoning depth. CESAR resolves test-time inverse scaling, transforming reasoning from  detriments into gains while revealing model-specific ``reasoning sweet spots", where performance peaks during test-time scaling. We achieve state-of-the-art results on MMAU Test-mini, substantially outperforming Gemini 2.5 Pro and GPT-4o Audio, and near-human-level performance on MMSU reasoning tasks. Through AI-as-judge evaluations and qualitative comparisons, we provide both quantitative and qualitative validation of our improved reasoning quality. Importantly, enhanced reasoning creates synergistic effects, simultaneously improving multimodal reasoning and perception capabilities. Overall, CESAR establishes a principled method for developing robust and scalable reasoning in Audio LLMs.
\end{abstract}


\section{Introduction}

The advent of Audio Large Language Models (Audio LLMs) has opened a new frontier in multimodal AI, promising sophisticated understanding of complex acoustic environments \citep{gong2023ltu, tang2023salmonn, xu2025qwen25omni}. Yet, a critical paradox emerges when these models are asked to reason: while chain-of-thought (CoT) prompting is a proven catalyst for reasoning in text-based domains \citep{wei2022chain, openai2024o1, deepseekai2025deepseekr1}, in audio it often backfires, underperforming non-reasoning versions. We are the first to systematically identify and diagnose this phenomenon as a \textbf{test-time inverse scaling problem} in Audio LLMs, where reasoning processes not only fail to improve performance but actively degrade it during inference, with longer reasoning chains yielding progressively worse results—often underperforming their direct answering versions that bypass reasoning entirely (Fig.~\ref{fig: general framework}). This test-time inverse scaling might lead to the premature conclusion that reasoning is inherently harmful for Audio LLMs \citep{r1_aqa}, but our investigation reveals the true culprit: models produce hallucinatory, inconsistent, and logically unsound reasoning processes when forced to ``think" without proper training on \textit{how} to reason.

Current methodologies are fundamentally ill-equipped to solve this problem. The dominant approach—supervised fine-tuning (SFT) on CoT datasets \citep{ma2025audiocot, liu2025audioreasoner}—teaches models to merely memorize and mimic reasoning templates rather than developing genuine analytical capabilities. While recent reinforcement learning with verifiable rewards (RLVR) methods \citep{r1_aqa, ke_omni_r} represent progress, they remain constrained by outcome-only reward structures that exclusively value final answer correctness and format compliance. This shallow supervision fails to address the root cause: poor reasoning processes that accumulate errors over longer chains, allowing models to generate final answers through flawed or irrelevant logic while perpetuating the very issues of inconsistency and hallucination that lead to test-time inverse scaling.

We address these limitations by introducing \textbf{CESAR} (Consistent, Effective and Scalable Audio Reasoners), representing a fundamental \textbf{paradigm shift from outcome verification to rewarding the reasoning process}. Our framework leverages Group Relative Policy Optimization (GRPO) \citep{shao2024grpo} not merely as a verifier, but as a mechanism to explicitly cultivate reasoning as a controllable, trainable skill. At its core lies a multi-faceted reward suite that provides granular feedback on the reasoning process itself, systematically incentivizing answer \textbf{correctness}, \textbf{format} compliance,  \textbf{consistency} between thoughts and answers, \textbf{structured} analytical patterns, \textbf{causal} reasoning, \textbf{domain} knowledge integration, and \textbf{calibrated} reasoning depth that avoids catastrophic overthinking. Through this process-centric approach, we transform reasoning from an unpredictable liability into a reliable, scalable asset that enables effective test-time scaling through discovered ``reasoning sweet spots." In summary, our approach makes several key contributions:

\begin{figure}[!t]
\centering
\includegraphics[width=\linewidth]{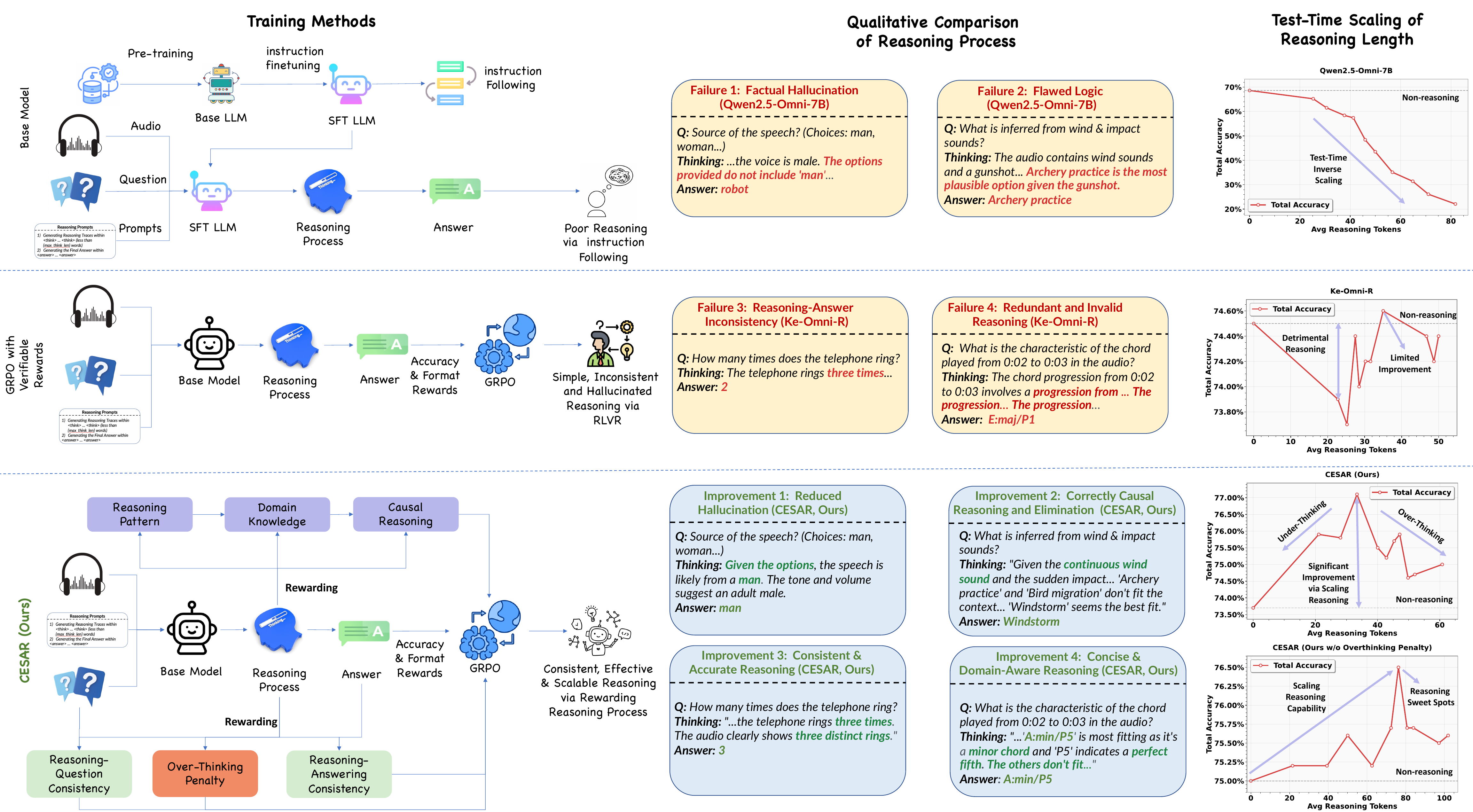}
\caption{
   General Framework of Different Training Methods for Audio Reasoning Models.
}
\label{fig: general framework}
\end{figure}

\begin{enumerate}
    \item We identify and diagnose the \textbf{test-time inverse scaling phenomenon} in Audio LLMs, where reasoning processes degrade performance during inference due to hallucination, inconsistency, and unstructured thought patterns. We demonstrate this stems from inadequate training of reasoning processes rather than unsolvable limitations of reasoning itself.
    
    \item We propose \textbf{CESAR}, a framework employing reasoning process rewards that incentivize consistency, structured analytical patterns, domain knowledge integration, and calibrated reasoning depth to extend current outcome-only RLVR methods. With GRPO, CESAR explicitly cultivates robust reasoning, resolving the test-time inverse scaling problem.

    \item We demonstrate that cultivating robust reasoning capability by \textbf{CESAR unlocks effective test-time scaling}: while poorly trained models suffer catastrophic degradation with longer reasoning chains, our models discover optimal ``reasoning sweet spots" for substantial training-free performance gains, validating our scalable reasoning capability.
    
    \item Our method achieves \textbf{SOTA}  on MMAU Test-mini, surpassing GPT-4o Audio and Gemini 2.5 Pro, and \textbf{near human-level performance} on MMSU reasoning tasks. Most importantly, we observe that enhanced reasoning creates synergistic effects that simultaneously improve multimodal reasoning and  perception capabilities.

    \item We introduce a novel \textbf{AI-as-judge} evaluation framework and comprehensive \textbf{qualitative analysis} that rigorously validate our enhanced reasoning quality, demonstrating commanding win rates against strong baselines and concrete reductions in hallucination while improving logical coherence and reasoning-answer consistency.
\end{enumerate}

\section{Related Work}

\paragraph{Audio Large Language Models.}
The development of Audio Large Language Models (Audio LLMs) has rapidly progressed from foundational audio-to-text tasks to sophisticated multimodal systems. Early work \citep{elizalde2022clap, wu2021wav2clip} established cross-modal understanding through contrastive learning. This paved the way for decoder-based models capable of open-ended generation \citep{deshmukh2023pengi, gong2023ltu}. The current generation of models, such as SALMONN \citep{tang2023salmonn}, the Qwen-Audio series \citep{qwen1_audio, qwen2_audio}, and Audio Flamingo \citep{kong2024audio, kong2025audioflamingo3}, have demonstrated increasingly comprehensive capabilities through large-scale pre-training and instruction tuning. As state-of-the-art models like Qwen2.5-Omni \citep{xu2025qwen25omni}, GPT-4o Audio \citep{gpt4oaudio} and Gemini 2.5 \citep{gemini25} achieve near-human audio understanding, the research frontier has shifted towards a more profound challenge: enabling these models to genuinely \textit{reason} about the acoustic world.

\paragraph{The Limits of Supervised Reasoning.}
Chain-of-thought (CoT) prompting has been transformative for eliciting reasoning in text-based LLMs \citep{wei2022chain}. Naturally, this paradigm was extended to the audio domain through supervised fine-tuning (SFT) on CoT datasets \citep{ma2025audiocot, liu2025audioreasoner, kong2025audioflamingo3}. However, these SFT-based approaches share a fundamental limitation: they teach models to \textbf{imitate reasoning templates, not to develop genuine analytical skill}. This results in models that can produce syntactically plausible reasoning traces but which are often brittle, fail to generalize to complex, unseen problems, and do not address the underlying causes of reasoning failure we identify, such as hallucination and inconsistency.

\paragraph{The Untapped Potential of Reinforcement Learning.}
Reinforcement learning (RL) offers a promising alternative to supervised imitation, as demonstrated by the success of models like OpenAI's o1 \citep{openai2024o1} and DeepSeek-R1 \citep{deepseekai2025deepseekr1} in the text domain. These works show that sophisticated reasoning can be cultivated directly through reward optimization, with methods like Group Relative Policy Optimization (GRPO) \citep{shao2024grpo} proving particularly effective. Early attempts to apply these techniques to audio, such as R1-AQA \citep{r1_aqa} and Ke-Omni-R \citep{ke_omni_r}, have shown initial success. However, they are fundamentally constrained by an \textbf{outcome-oriented reward paradigm}, optimizing solely for the correctness of the final answer. This shallow supervision signal is insufficient; it fails to penalize logical fallacies or reward coherent analytical processes, thereby directly contributing to the poor reasoning capability that causes the test-time inverse scaling problem. Our work addresses this gap by moving from outcome verification to a granular, process-oriented reward system.

\section{Methodology}
\label{sec:methodology}

Existing audio reasoning methods like Ke-Omni-R \citep{ke_omni_r} suffer from reasoning-answer inconsistency, unstructured reasoning, and test-time inverse scaling (Fig. \ref{fig: general framework}) caused by outcome-only verifiable rewards and uncontrolled reasoning emergence. In this paper, we propose CESAR to transform reasoning into a controllable skill through comprehensive reasoning process rewards that incentivize consistency, structured reasoning, and optimal reasoning depth—while discovering model-specific ``reasoning sweet spots" where performance peaks during test-time scaling.

\subsection{Problem Formulation}

Let $\mathcal{D} = {(a_i, q_i, \mathcal{C}_i, y_i)}_{i=1}^N$ denote the audio question-answering dataset, where $a_i$ represents the audio input, $q_i$ is the question, $\mathcal{C}_i = {c_1, c_2, c_3, c_4}$ is the set of multiple-choice options, and $y_i \in \mathcal{C}_i$ is the ground truth answer. Our goal is to train an Audio LLM $\pi_\theta$ that can generate both a reasoning process $t_i$ and a final answer $\hat{y}_i$ given the input $(a_i, q_i, \mathcal{C}_i)$.  The model output follows a structured format where we have:

\begin{equation}
    \pi_\theta(a_i, q_i, \mathcal{C}_i) = \langle \text{think} \rangle t_i \langle /\text{think} \rangle \langle \text{answer} \rangle \hat{y}_i \langle /\text{answer} \rangle
\end{equation}

Here, $t_i$ represents the CoT reasoning process and $\hat{y}_i$ is the predicted answer. This structured output allows us to separately evaluate both the reasoning quality and the final answer correctness. 

Reinforcement learning fine-tuning seeks to optimize the audio LLMs to maximize rewards:
\begin{equation}
\pi^* = \arg\max_\pi \mathbb{E}[R(s_i)],
\end{equation}
where $s_i = (t_i, \hat{y}_i)$ represents the complete model output. Current approaches like R1-AQA and Ke-Omni-R employ outcome-only rewards based solely on answer correctness and format compliance: $R_{\text{RLVR}}(s_i) = \mathbb{I}[\hat{y}_i = y_i] + \mathbb{I}[\text{ValidFormat}(s_i)]$. This impoverished signal leads to three critical failure modes: (1) \textbf{Random Emergence} of reasoning patterns without effective control; (2) \textbf{Reasoning-Answer Inconsistency} where models generate answers inconsistent with their reasoning logic; (3) \textbf{Lack of Structured Reasoning} strategies like elimination or multi-step deduction.

Our fundamental insight is that genuine reasoning capability requires explicit process-oriented incentivization rather than spontaneous emergence. We achieve this through a multi-faceted reward suite that provides granular feedback on reasoning quality, consistency, and structure during training, transforming reasoning from an unpredictable phenomenon into a controllable, trainable skill.

\subsection{From Outcome-Based to Process-Oriented Reasoning Control}

Current RLVR approaches fundamentally fail to distinguish between genuine reasoning and fortunate guessing, leading to random emergence of reasoning behaviors that cannot be systematically controlled or guaranteed. Our framework introduces a novel paradigm that transforms reasoning from an unpredictable emergent phenomenon into a controllable, trainable capability through comprehensive process supervision.

Our total reward $R_{\text{total}}(s_i)$ decomposes into two complementary components that address distinct aspects of reasoning quality:

\begin{equation}
\label{equ: tot rewards}
     \underbrace{\alpha_1 R_{\text{acc}}(s_i) + \alpha_2 R_{\text{format}}(s_i)}_{\text{Verifiable Rewards}} + \underbrace{\alpha_3 R_{\text{consistency}}(s_i) + \alpha_4 R_{\text{keywords}}(s_i) + \alpha_5 R_{\text{overthinking penalty}}(s_i)}_{\text{Reasoning Process Rewards}}
\end{equation}

The verifiable rewards maintain essential correctness constraints and structural integrity, while our \textbf{Reasoning Process Rewards} explicitly shape reasoning quality, consistency, and conciseness. Here $s_i = (t_i, \hat{y}_i)$ represents the complete model output encompassing both reasoning trace and final answer, and $\{\alpha_j\}_{j=1}^{5}$ are weight coefficients that balance answer correctness with reasoning refinement. In practice, we set $\alpha_1=5.0$ for accuracy and $\alpha_{2-5}=1.0$ for other components.

\subsubsection{Foundation: Verifiable Correctness and Structural Integrity}

While transcending traditional RLVR limitations, our framework maintains rigorous grounding in verifiable outcomes. The \textbf{Accuracy Reward} $R_{\text{acc}}(s_i) = \mathbb{I}[\hat{y}_i = y_i]$ establishes the fundamental correctness constraint that ensures reasoning improvements do not come at the expense of answer accuracy. This binary signal prevents the optimization process from learning elaborate but incorrect reasoning patterns, anchoring all process improvements in empirical validity.

The \textbf{Format Reward} enforces structural compliance and prevents the model from bypassing the reasoning framework:

\begin{equation}
    R_{\text{format}}(s_i) = \mathbb{I}[\text{ValidFormat}(s_i)]
\end{equation}

This reward ensures the model produces outputs with proper XML tag structure, specifically requiring both $\langle \text{think} \rangle t_i \langle /\text{think} \rangle$ and $\langle \text{answer} \rangle \hat{y}_i \langle /\text{answer} \rangle$ components. This creates a disciplined reasoning environment where models must engage with the reasoning process rather than circumventing it through format violations.

\subsubsection{Semantic Coherence and Reasoning-Answer Alignment}

A critical challenge in current reasoning approaches is the pervasive problem of reasoning-answer inconsistency, where models generate correct answers despite fundamentally flawed or irrelevant reasoning processes. Additionally, when reasoning traces are unrelated to the question and choices, models are prone to hallucination. The \textbf{Reasoning Consistency Reward} introduces explicit semantic supervision that ensures reasoning traces genuinely support their corresponding conclusions:
\begin{equation}
    R_{\text{consistency}}(s_i) = \text{Sim}_{\text{semantic}}(t_i, \hat{y}_i) + \text{Sim}_{\text{semantic}}(t_i, Q_i),
\end{equation}
where $Q_i = (q_i, \mathcal{C}_i)$ represents the complete question context including both the question text and available choices. This dual-alignment formulation addresses two critical failure modes. The answer-alignment component $\text{Sim}_{\text{semantic}}(t_i, \hat{y}_i)$ prevents reasoning processes from becoming disconnected from their conclusions, which would render the reasoning ineffective. The question-alignment component $\text{Sim}_{\text{semantic}}(t_i, Q_i)$ ensures the reasoning remains focused on the posed question and available choices, preventing hallucination and off-topic elaboration.

We implement semantic similarity using concept overlap (e.g., via overlapped words):
\begin{equation}
    \text{Sim}_{\text{semantic}}(x, y) = \frac{\text{ConceptOverlap}(x, y)}{\max(|\text{Concepts}(x)|, |\text{Concepts}(y)|)}
\end{equation}
where the normalization ensures bounded similarity scores in $[0,1]$.  This approach represents a departure from outcome-only optimization, introducing explicit supervision signals that distinguish between reasoning processes that accidentally arrive at correct answers and those that systematically derive conclusions through valid analytical pathways.

\subsubsection{Incentivizing Structured Reasoning and Penalizing Overthinking}

To explicitly shape reasoning quality, our framework employs a two-pronged strategy: we positively \textbf{incentivize structured reasoning} while simultaneously \textbf{penalizing inefficient overthinking}. The primary mechanism for structured reasoning is the \textbf{Keywords Reward}, which acts as a cognitive scaffold to transform random emergent thoughts into controlled, sophisticated analytical behaviors:

\begin{equation}
    R_{\text{keywords}}(s_i) = R_{\text{pattern}}(s_i) + R_{\text{logic}}(s_i) + R_{\text{domain}}(s_i)
\end{equation}

This tri-component design addresses three fundamental aspects of structured reasoning: structured analytical patterns, logical rigor, and domain expertise integration.

\textbf{Structured Analytical Patterns.} The pattern recognition component systematically rewards models for developing structured reasoning architectures rather than relying on intuitive leaps: $R_{\text{pattern}}(s_i) = \sum_{p \in \mathcal{P}} \cdot \mathbb{I}[\text{Pattern}_p \text{ detected in } t_i]$. The pattern set $\mathcal{P}$ captures sophisticated reasoning architectures through key categories such as sequential organization, comparative analysis, systematic evaluation, and explicit justification. Complete pattern specifications are detailed in App.~\ref{app:detailed_keywords}.

\textbf{Logical Rigor and Causal Reasoning.} The reasoning indicators component cultivates sophisticated logical thinking by rewarding linguistic markers that indicate deep analytical processes: $R_{\text{logic}}(s_i) = \sum_{l \in \mathcal{L}} \cdot \mathbb{I}[\text{Keyword}_l \text{ detected in } t_i]$. The reasoning logic taxonomy $\mathcal{L}$ strategically targets distinct logical functions including formal deduction markers, premise establishment, hypothetical reasoning, and evidential conclusions. These linguistic signatures promote sophisticated logical progression from premises to conclusions (complete taxonomy in App.~\ref{app:detailed_keywords}).

\textbf{Domain Knowledge Integration.} We also incentivize the use of domain knowledge, where the domain component rewards models for incorporating audio-specific expertise rather than generic reasoning patterns: $R_{\text{domain}}(s_i) = \sum_{d \in \mathcal{D}} w_d \cdot \mathbb{I}[\text{Term}_d \text{ detected in } t_i]$. The domain vocabulary $\mathcal{D}$ encompasses specialized terminology across acoustic properties, musical concepts, speech analysis, and environmental audio understanding. This encourages models to ground their reasoning in signal-specific expertise rather than superficial pattern matching (complete vocabulary in App.~\ref{app:detailed_keywords}).

\textbf{Overthinking Penalty.} The necessary counterpart to rewarding structured thought is penalizing its inefficient opposite. The \textbf{Overthinking Penalty} addresses a critical failure mode: the tendency for models to engage in redundant, verbose reasoning that accumulates errors rather than improving analysis quality. This component actively discourages overthinking by penalizing excessively long reasoning traces:

\begin{equation}
\label{equ: overthink penalty}
    R_{\text{overthinking penalty}}(s_i) = f_{\text{length}}(|t_i|) = 1 - \frac{|t_i|}{L_{\text{max\_output}}}
\end{equation}

where $f_{\text{length}}(l)$ is a linear penalty function that decreases as reasoning length $|t_i|$ increases, normalized by the maximum output length $L_{\text{max\_output}}$ (we set as 256 in practice). This design specifically targets common failure modes including circular reasoning, repetitive analysis, and tangential elaboration. By learning to terminate reasoning at an appropriate depth, models develop a meta-cognitive awareness that prevents hallucination accumulation while maintaining analytical rigor.

\subsection{Cultivating Reasoning Capability via Online RL}

Our framework operationalizes process-oriented reasoning control through Group Relative Policy Optimization (GRPO) \citep{shao2024grpo}, systematically cultivating reasoning capabilities rather than relying on random emergence. For each training sample $(a_i, q_i, \mathcal{C}_i, y_i)$, we sample $K$ responses $\{s_i^{(k)}\}_{k=1}^K \sim \pi_\theta(\cdot | a_i, q_i, \mathcal{C}_i)$ and optimize the objective:

\begin{equation}
    \mathcal{L}_{\text{GRPO}} = \mathcal{L}_{PG}^{\text{multi-faceted}} + \beta \cdot \mathcal{L}_{KL},
\end{equation}

where the policy gradient loss $\mathcal{L}_{PG}^{\text{multi-faceted}} = -\mathbb{E} \left[ \sum_{k=1}^K A(s^{(k)}) \cdot \log \pi_\theta(s^{(k)} | a, q, \mathcal{C}) \right]$ provides granular feedback on both analytical processes and final outcomes. The advantage function $A(s_i^{(k)}) = R_{\text{total}}(s_i^{(k)}) - \frac{1}{K} \sum_{j=1}^K R_{\text{total}}(s_i^{(j)})$ enables models to distinguish between high-quality and low-quality reasoning processes, while KL regularization $\mathcal{L}_{KL} = \mathbb{E} \left[ \text{KL}(\pi_\theta || \pi_{\text{ref}}) \right]$ maintains training stability.

To enhance model robustness against linguistic variance, we employ systematic \textbf{data augmentation} that expands our training corpus $\mathcal{D}$ into an augmented version $\mathcal{D}'$ by generating multiple linguistic variations for each question while preserving ground-truth answers. For each instance $(a_i, q_i, \mathcal{C}_i, y_i) \in \mathcal{D}$, we apply answer-invariant transformation templates $\mathcal{T} = \{T_1, \ldots, T_M\}$, where each transformation $T_k$ generates $q'_{i,k} = T_k(q_i, \mathcal{C}_i)$, creating training samples $(a_i, q'_{i,k}, \mathcal{C}_i, y_i)$ with unchanged audio and answers. This forces the model to learn underlying reasoning patterns rather than superficial textual correlations. Complete template specifications are provided in App.~\ref{app: Experimental Details}.

\subsection{Unlocking Reasoning Capability via Test-Time Scaling}

To understand the test-time inverse scaling phenomenon and validate our proposed methods, we introduce \textbf{Test-Time Scaling} to systematically analyze reasoning dynamics by evaluating performance across varying maximum thinking lengths $L_{\text{max\_think}}$. We define performance as $P(L_{\text{max\_think}}) = \mathbb{E} \left[ \mathbb{I}[\hat{y} = y] \mid |t| \leq L_{\text{max\_think}} \right]$ and identify the \textbf{``reasoning sweet spot"} where performance peaks: $L_{\text{sweet}} = \arg\max_{L} P(L)$. Through this simple scaling of reasoning length, CESAR achieves substantial  improvements, with particularly dramatic gains at its reasoning sweet spot, while baseline models show limited improvement or continued degradation (See Fig. \ref{fig:scaling_and_judge_analysis}). This method effectively unlocks reasoning capability at test-time by revealing that our process-oriented training enables models to discover and utilize their optimal reasoning depth for maximum performance.

\section{Experiments}
\label{sec:experiments}

\subsection{Experimental Setup}

We evaluate our framework on  challenging out-of-distribution (OOD) audio reasoning benchmarks: MMAU Test-mini \citep{sakshi2024mmau} with 1k expertly annotated questions spanning speech, sounds, and music requiring 27 distinct reasoning skills, and MMSU \citep{mmsu} with  5k audio-question pairs and granular perception-reasoning task separation. Training uses the AVQA dataset \citep{yang2022avqa} enhanced through systematic data augmentation that generates diverse question phrasings while preserving answer labels. Our experiments employ Qwen2.5-Omni-7B with GRPO, sampling $K=8$ responses per training example. Reward coefficients balance correctness ($\alpha_1=5.0$) with other rewards ($\alpha_{2-5}=1.0$). We compare against base model variants, Ke-Omni-R baseline, proprietary models, and open-source audio models. Unless otherwise specified, all reported scores of our methods are achieved with reasoning. Complete details are in App.~\ref{app: Experimental Details}. 

\begin{figure}[t]
\centering
\begin{minipage}{0.58\textwidth}
\centering
\captionof{table}{MMAU Test-Mini benchmark results. \colorbox{lightblue}{\textbf{Blue}} indicates best performance, \colorbox{lightgreen}{\textbf{green}} indicates second-best. Accuracy (\%) is reported across audio modalities. OP means overthinking penalty. See App. \ref{app:mmau_results} for details.}
\label{tab:mmau_main_results}
\resizebox{\textwidth}{!}{
\begin{tabular}{l|c|ccc|c}
\toprule
\textbf{Method} & \textbf{Reasoning} & \textbf{Sound} & \textbf{Music} & \textbf{Speech} & \textbf{Total Accuracy} \\
\midrule
\multicolumn{6}{c}{\cellcolor{blue!15}\textbf{Our Proposed Methods}} \\
\midrule
\textbf{CESAR} & \cmark & \colorbox{lightblue}{\textbf{83.48}} & \colorbox{lightblue}{\textbf{73.05}} & 74.77  & \colorbox{lightblue}{\textbf{77.10}} \\
\textbf{CESAR} & \xmark & 79.88  & 67.96 & 73.27 & 73.70 \\
\textbf{CESAR w/o OP} & \cmark & \colorbox{lightgreen}{\textbf{81.98}} & 70.06 & \colorbox{lightblue}{\textbf{77.48}} & \colorbox{lightgreen}{\textbf{76.50}} \\
\textbf{CESAR w/o OP} & \xmark & 80.48 & 70.06 & 74.47 & 75.00 \\
\midrule
\multicolumn{6}{c}{\cellcolor{gray!15}\textbf{RL Baseline Methods}} \\
\midrule
\textbf{Ke-Omni-R} & \cmark & 79.28 & 70.06 & 74.47 & 74.60 \\
\textbf{Ke-Omni-R} & \xmark & 78.38 & \colorbox{lightgreen}{\textbf{70.96}} & 74.17 & 74.50 \\
\midrule
\multicolumn{6}{c}{\cellcolor{orange!15}\textbf{Proprietary Models}} \\
\midrule
\textbf{Gemini 2.5 Pro} & - & 75.08 & 68.26 & 71.47 & 71.60 \\
\textbf{Gemini 2.5 Flash} & -  & 73.27 & 65.57 & \colorbox{lightgreen}{\textbf{76.58}} & 71.80 \\
\textbf{GPT-4o Audio} & -  & 64.56 & 56.29 & 66.67 & 62.50 \\
\midrule
\multicolumn{6}{c}{\cellcolor{gray!15}\textbf{Base Models}} \\
\midrule
\textbf{Qwen2.5-Omni-7B} & \cmark & 69.07 & 59.58 & 66.97 & 65.20 \\
\textbf{Qwen2.5-Omni-7B} & \xmark & 72.37 & 64.37 &  69.07 & 68.60 \\
\bottomrule
\end{tabular}
}
\end{minipage}
\hfill
\begin{minipage}{0.38\textwidth}
\centering
\includegraphics[width=\textwidth]{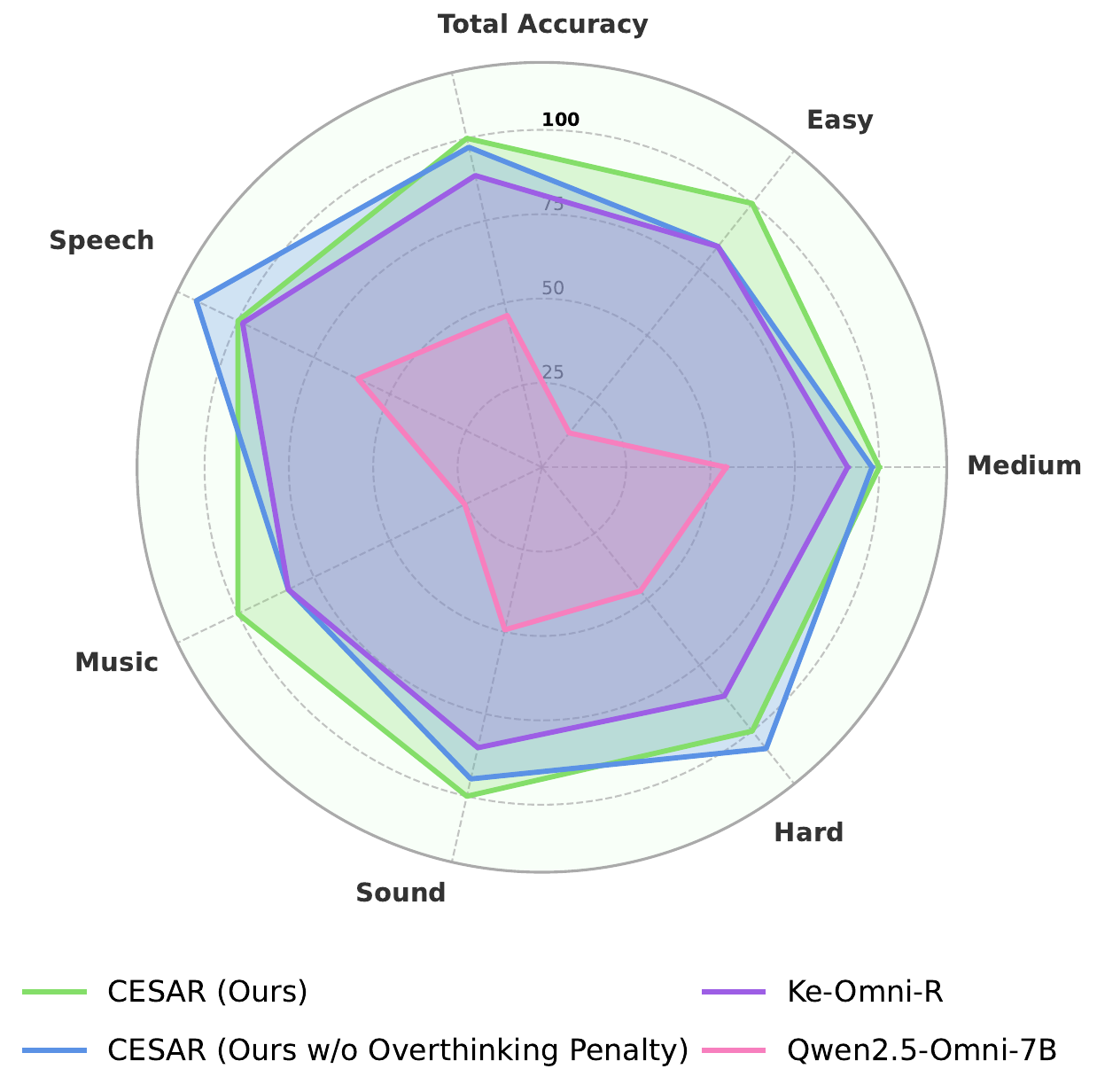}
\caption{Task-wise comparison on the MMAU Test-mini Benchmark (Scores are normalized by CESAR). See App. \ref{app: Beyond Aggregate Scores: A Task-Level Analysis of Controllable Reasoning} for more results.}
\label{fig:radar_analysis main}
\end{minipage}
\end{figure}

\begin{table*}[!t]
\centering
\caption{MMSU Results \citep{mmsu}. Best scores are in \colorbox{lightblue}{blue}, second-best in \colorbox{lightgreen}{green}. Results show accuracy (\%) across perception and reasoning tasks. See App. \ref{app:mmsu_results} for more results.}
\label{tab:mmsu_main_results}
\resizebox{\textwidth}{!}{
\begin{tabular}{l|cccc|cccc|c}
\toprule
\multirow{2}{*}{\textbf{Models}} & \multicolumn{4}{c|}{\textbf{Perception Tasks}} & \multicolumn{4}{c|}{\textbf{Reasoning Tasks}} & \multirow{2}{*}{\textbf{Overall}} \\
\cmidrule{2-9}
& \textbf{Semantics} & \textbf{Phonology} & \textbf{Paralinguistics} & \textbf{Avg} & \textbf{Semantics} & \textbf{Phonology} & \textbf{Paralinguistics} & \textbf{Avg} & \\
\midrule
\textbf{CESAR (Ours)} & \colorbox{lightgreen}{\textbf{60.16}} & 50.16 & 39.50 & \colorbox{lightgreen}{\textbf{48.45}} & \colorbox{lightblue}{\textbf{88.72}} & 80.66 & 57.01 & \colorbox{lightgreen}{\textbf{81.07}} & \colorbox{lightgreen}{\textbf{64.24}} \\
Ke-Omni-R & 58.74 & 46.31 & \colorbox{lightgreen}{\textbf{40.50}} & 47.09 & 86.82 & 74.31 & \colorbox{lightgreen}{\textbf{60.00}} & 78.06 & 62.08 \\
Gemini 1.5 Pro & 57.06 & \colorbox{lightgreen}{\textbf{53.60}} & 31.23 & 46.10 & 79.47 & \colorbox{lightgreen}{\textbf{83.46}} & 46.33 & 76.16 & 60.68 \\
Qwen2.5-Omni & 55.12 & 37.33 & 39.35 & 42.50 & \colorbox{lightgreen}{\textbf{88.00}} & 81.37 & 48.36 & 79.83 & 60.57 \\
GPT-4o Audio & 59.70 & 41.56 & 21.44 & 39.67 & 80.83 & 78.74 & 26.25 & 71.96 & 56.38 \\
\midrule
Human & \colorbox{lightblue}{\textbf{87.10}} & \colorbox{lightblue}{\textbf{94.32}} & \colorbox{lightblue}{\textbf{92.88}} & \colorbox{lightblue}{\textbf{91.24}} & 82.16 & \colorbox{lightblue}{\textbf{87.60}} & \colorbox{lightblue}{\textbf{89.12}} & \colorbox{lightblue}{\textbf{86.77}} & \colorbox{lightblue}{\textbf{89.72}} \\
\bottomrule
\end{tabular}
}
\end{table*}

\subsection{Main Results: State-of-the-Art Performance Across Benchmarks}

\paragraph{MMAU: Significant Performance Gains via Rewarding the Reasoning Process}
As shown in Tab.~\ref{tab:mmau_main_results}, our method establishes new SOTA performance on MMAU Test-mini, decisively surpassing leading proprietary models including GPT-4o Audio and Gemini 2.5 Pro. Most importantly, we demonstrate that process-oriented training delivers synergistic improvements across both reasoning modes: compared to the base model, CESAR achieves substantial gains both with reasoning  and without reasoning, proving that cultivating reasoning processes fundamentally enhances the model's cognitive capabilities. Our framework also significantly outperforms outcome-only RL methods, with reasoning mode delivering larger benefits than the Ke-Omni-R baseline. The radar analysis (Fig.~\ref{fig:radar_analysis main}) reveals controllable reasoning architectures: our two variants exhibit engineered cognitive profiles—CESAR w/o OP excelling on hard tasks through deeper analysis, while full CESAR maintains balanced efficiency across difficulty levels—establishing reasoning as a systematically controllable capability rather than random emergence. See App. \ref{app:mmau_results}, \ref{app: Beyond Aggregate Scores: A Task-Level Analysis of Controllable Reasoning} for more results.

\paragraph{MMSU: Perceptual Improvements with Near-Human Reasoning}
On the MMSU benchmark (Tab.~\ref{tab:mmsu_main_results}), our CESAR achieves dual advances: reasoning capabilities that approach human levels (including super-human performance in semantic reasoning), while simultaneously outperforming larger competitors on perception tasks. This reveals an interesting synergistic effect where cultivating advanced reasoning through our reasoning process rewards also refines foundational auditory perception capabilities. However, while both capabilities advance substantially, the results illuminate a critical asymmetry: reasoning improvements have reached near-human parity, whereas perception performance, despite leading existing models, still exhibits a considerable gap relative to human baselines. This disparity identifies the ``perceptual bottleneck" as a key area for future work in achieving comprehensive human-level audio understanding. See App.~\ref{app:mmsu_results} for more results.

\subsection{Curing Test-Time Inverse Scaling and Unlocking Scalable Reasoning}

In Fig. \ref{fig:scaling_and_judge_analysis}, our Test-Time Scaling analysis reveals the \textbf{test-time inverse scaling} problem, where baseline models exhibit either a catastrophic performance collapse (untrained model) or volatile performance with no clear benefit from longer reasoning (standard RL baseline). In contrast, our methods resolve this issue, transforming reasoning from detriments into gains. As shown in Fig. \ref{fig:scaling_and_judge_analysis} (Left), even without the overthinking penalty, our model's performance steadily climbs to a peak of 76.50\%. Moreover, our full method demonstrates superior calibration; by explicitly penalizing inefficient thought, it discovers a more optimal \textbf{``reasoning sweet spot,"} achieving a higher peak accuracy of 77.1\% with a much shorter reasoning chain of approximately 35-40 tokens. This proves our methods enable consistent, effective reasoning that unlocks scalable capability during inference to achieve performance gains through scaling reasoning lengths. See App. \ref{app:test-time scaling_analysis} for more results.

\subsection{AI-as-Judge Evaluation: Quantifying Reasoning Quality Beyond Accuracy}
\begin{figure}[t!]
\centering
\includegraphics[width=\linewidth]{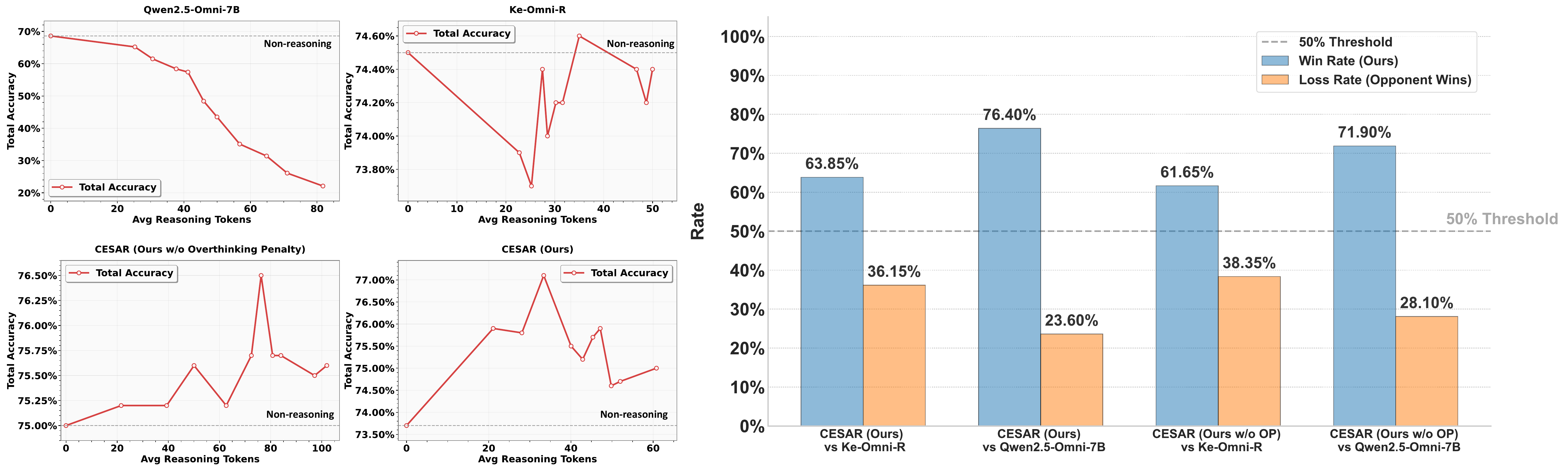}
\caption{
\textbf{Remediating Test-Time Inverse Scaling and Quantifying Reasoning Quality on MMAU Test-mini.} 
\textbf{(Left)} Test-time scaling analysis shows performance when increasing the reasoning tokens by sweeping different maximum thinking lengths from 0 to maximum output length (i.e., 250) in intervals of 25.
\textbf{(Right)} An AI-as-judge evaluation with GPT-4o Audio \citep{gpt4oaudio}  provides quantitative proof of our superior reasoning quality, showing our models achieve commanding win rates against strong baselines. Throughout, OP denotes the overthinking penalty. 
}
\label{fig:scaling_and_judge_analysis}
\end{figure}

To move beyond accuracy and verify our improved reasoning, we introduce an AI-as-Judge for head-to-head comparisons via GPT-4o Audio. As shown in Fig.~\ref{fig:scaling_and_judge_analysis} (Right), our method's reasoning process achieves commanding win rates against both baselines. Notably, even without the Overthinking Penalty, our core rewards still yield a dominant performance, while its inclusion further elevates the win rate. This corroborates the superior performance of our full method in the MMAU  (Tab.~\ref{tab:mmau_main_results}). These results provide direct evidence that our framework generates verifiably superior reasoning, a qualitative leap not captured by accuracy alone. See App.~\ref{app:llm_judge_analysis} for more details and prompts used.

\begin{table*}[!ht]
\centering
\caption{Qualitative Comparison of Reasoning Processes in MMAU Test-Mini. Our method demonstrates consistent and well-structured reasoning processes. \textcolor{rewardgreen}{\textbf{Green text}} highlights cultivated structured reasoning, while \textcolor{errorcolor}{\textit{red italics}} mark reasoning errors. Complete qualitative analysis in App.~\ref{app:Qualitative_Analysis}.}
\label{tab:qualitative_main_examples}
\resizebox{0.8\textwidth}{!}{
\begin{tabular}{p{0.48\linewidth} p{0.48\linewidth}}
\toprule
\multicolumn{2}{l}{\textbf{Example: Logical Deduction from Acoustic Cues}} \\
\multicolumn{2}{l}{\textbf{Question:} What kind of event could be taking place in the audio?} \\
\multicolumn{2}{l}{\textbf{Correct Answer:} A military parade with a marching band} \\
\addlinespace[3pt]
\midrule
\textcolor{ourscolor}{\textbf{CESAR (Correct)}} & \textcolor{baselinecolor}{\textbf{Qwen2.5-Omni-7B (Wrong)}} \\
\midrule
\textbf{Reasoning:} \textit{The presence of marching, \rewardkeyword{music}, and \rewardkeyword{shouting} \rewardkeyword{suggests} a large-scale event. The options are a quiet library, a heated argument, a military parade, and a gentle breeze. A military parade \rewardkeyword{fits the description best}, as it involves a marching band and organized \rewardkeyword{shouting}.}
&
\textbf{Reasoning:} \textit{The audio contains the sound of a man speaking, followed by a loud, \textcolor{errorcolor}{chaotic noise. The chaotic noise could be interpreted as a violent escalation}, so a heated argument is the most likely scenario.} \\
\addlinespace[2pt]
\textbf{Answer:} A military parade with a marching band
&
\textbf{Answer:} \textcolor{errorcolor}{A heated argument escalating to violence} \\
\midrule
\midrule
\multicolumn{2}{l}{\textbf{Example: Superior Reasoning-Answer Consistency}} \\
\multicolumn{2}{l}{\textbf{Question:} How many times does the telephone ring in the audio?} \\
\multicolumn{2}{l}{\textbf{Correct Answer:} 3} \\
\midrule
\textcolor{ourscolor}{\textbf{CESAR (Correct)}} & \textcolor{baselinecolor}{\textbf{Ke-Omni-R (Wrong)}} \\
\midrule
\addlinespace[3pt]
\textbf{Reasoning:} \textit{\rewardkeyword{Considering the options}, the telephone \rewardkeyword{rings} three times. The \rewardkeyword{audio} clearly shows three distinct \rewardkeyword{rings}.}
&
\textbf{Reasoning:} \textit{The telephone rings \textcolor{errorcolor}{three times}, as indicated by the three distinct ring sounds in the audio.} \\
\addlinespace[2pt]
\textbf{Answer:} 3
&
\textbf{Answer:} \textcolor{errorcolor}{2} \\
\bottomrule
\end{tabular}
}
\end{table*}

\subsection{Qualitative Analysis: Concrete Evidence of Reasoning Improvements}

Beyond quantitative improvements, our methods also produce superior reasoning processes. Tab.~\ref{tab:qualitative_main_examples} illustrates two critical failure modes that our process-oriented rewards address. In the military parade example, our model systematically analyzes acoustic cues (``marching, music, and shouting") to reach the correct conclusion, while Qwen2.5-Omni-7B misinterprets organized sounds as ``chaotic noise" and makes unfounded inferences about violence. The telephone counting example reveals an even more fundamental issue: reasoning-answer inconsistency, where Ke-Omni-R correctly identifies ``three rings" in its reasoning trace but inexplicably outputs ``2" as the final answer. Our consistency reward explicitly prevents such disconnects between reasoning processes and conclusions, ensuring that correct reasoning translates to correct answers. See App. \ref{app:Qualitative_Analysis} for more results.

\subsection{Ablation Study: Quantifying Component Contributions}

\begin{table}[!ht]
\centering
\caption{Progressive ablation study on MMAU Test-mini. We start from the full CESAR method and progressively remove components to isolate their individual contributions. All scores are obtained using reasoning at their reasoning sweet spots. See App.~\ref{app:Ablation_Results} for more results.}
\label{tab:ablation_main_results}
\resizebox{\textwidth}{!}{
\begin{tabular}{l|ccccc|c}
\toprule
\textbf{Method} & \textbf{RL Training} & \textbf{Consistency} & \textbf{Keywords} & \textbf{Data Augmentation} & \textbf{Overthinking Penalty} & \textbf{Overall Accuracy (\%)} \\
\midrule
\textbf{Full Method (CESAR)} & \cmark & \cmark & \cmark & \cmark & \cmark & \textbf{77.10} \\
\quad Ablating Overthinking Penalty & \cmark & \cmark & \cmark & \cmark & \xmark & 76.50  \\
\quad Ablating Data Augmentation & \cmark & \cmark & \cmark & \xmark & \xmark & 76.20 \\
\quad Ablating Keywords & \cmark & \cmark & \xmark & \xmark & \xmark & 75.20  \\
\quad Ablating Consistency (Ke-Omni-R) & \cmark & \xmark & \xmark & \xmark & \xmark & 74.60  \\
\quad Ablating RL Training (Base Model) & \xmark & \xmark & \xmark & \xmark & \xmark & 65.20  \\
\bottomrule
\end{tabular}
}
\end{table}

Our progressive ablation study (Tab.~\ref{tab:ablation_main_results}) systematically deconstructs the components of our method. The results confirm the necessity of RL, as its removal triggers a catastrophic performance collapse. Building upon this RL foundation, our process-oriented rewards demonstrate strong synergy. The \textit{Keywords} reward yields the largest single gain over the outcome-only RL baseline (Ke-Omni-R) by sculpting higher-quality, structured reasoning processes. The \textit{Consistency} reward also provides a crucial boost by bridging the critical gap between a model's reasoning and its final output. The final components, \textit{Data Augmentation} and the \textit{Overthinking Penalty}, provide the necessary robustness and calibration to achieve peak performance. Ultimately, the ablation study demonstrates a clear synergistic effect: while each component provides a quantifiable and crucial performance gain, it is their holistic integration within the CESAR framework that unlocks state-of-the-art performance.


\section{Conclusion}

In this paper, we introduce CESAR to address the  test-time inverse scaling problems in Audio LLMs, where CoT reasoning degrades performance due to inadequate optimization of reasoning processes in existing SFT and RLVR methods. Our methods shift from outcome verification to rewarding the reasoning process transforms reasoning from detriments into significant performance gains through GRPO with multi-faceted process rewards. We achieve SOTA results on massive benchmarks, surpassing GPT-4o Audio and Gemini 2.5 Pro, while demonstrating that test-time scaling is a double-edged sword—catastrophic for poorly trained models but enabling substantial gains through discovered ``reasoning sweet spots" for models with strong reasoning capabilities like CESAR. Our comprehensive evaluation across multiple OOD benchmarks reveals synergistic effects where enhanced reasoning improves both multimodal reasoning and perception capabilities, while our AI-as-judge evaluations and qualitative comparisons provide both quantitative and qualitative validation of our improved reasoning quality beyond accuracy. CESAR establishes a principled methodology for developing robust, scalable reasoning in Audio LLMs. See App.~\ref{app: llm usage} for details on our LLM usage.

\bibliography{iclr2025_conference}
\bibliographystyle{iclr2025_conference}

\appendix

\clearpage


\section{Discussion}
\label{app: discussion}

In this paper, we explore how to cultivate robust, scalable, and effective reasoning in Audio Large Language Models. Despite the widespread success of chain-of-thought (CoT) reasoning in domains such as mathematics and coding \citep{gemini25,openai2024o1,deepseekai2025deepseekr1,shao2024grpo}, efforts to introduce reasoning capabilities into Audio Large Language Models \citep{fla1,fla2,kimi} have encountered a central paradox: CoT, a reliable catalyst for reasoning in text, consistently fails in the audio domain. While numerous works have attempted to leverage CoT prompting to enhance audio LLM reasoning and understanding capabilities \citep{ma2025audiocot, liu2025audioreasoner}, several studies including R1-AQA \citep{r1_aqa} have discovered that incorporating reasoning mechanisms not only fails to improve performance but may actually harm it. 

Our systematic investigation reveals a more profound issue that we term \textit{test-time inverse scaling} in Audio LLMs—a phenomenon we are the first to systematically diagnose as a test-time problem where prompting a model to ``think" during inference yields worse results than instinctual, direct answering. When we scale state-of-the-art open-source models such as Qwen2.5-Omni-7B during test-time, their performance counterintuitively degrades as reasoning length increases, often falling below their non-reasoning baselines (Fig.~\ref{fig:scaling_and_judge_analysis}). Similar patterns emerge in Ke-Omni-R \citep{ke_omni_r}, where reasoning during inference frequently underperforms direct answering approaches. \textbf{This test-time inverse scaling manifests in two critical failure modes: (1) any Audio LLM exhibiting worse performance when reasoning is enabled compared to direct answering (as observed in Qwen2.5-Omni-7B and most cases of Ke-Omni-R in Tab.~\ref{tab:mmau_main_results}), and (2) progressive performance degradation as reasoning chain length increases during test-time (as demonstrated in Fig.~\ref{fig:scaling_curves full}).}

Our investigation reveals that this test-time inverse scaling is not a fundamental limitation of reasoning itself, but a symptom of inadequate training: the models possess poor reasoning capability because they have never been properly \textit{taught how} to reason. Our research fundamentally reframes this challenge, moving it from a problem of pattern memorization to one of controllable skill development. We demonstrate that effective reasoning is not an unpredictable emergent phenomenon, but a trainable capability that can be systematically cultivated by directly rewarding the reasoning \textit{process}, thereby transforming reasoning from a liability into a systematic advantage for audio understanding. 

\begin{mdframed}[linecolor=blue!60, backgroundcolor=blue!5, roundcorner=10pt]
\textbf{Key Insight:} The failure of reasoning in Audio LLMs stems not from a fundamental limitation of the models, but from a flawed training paradigm. True reasoning capability is unlocked by shifting focus from supervising outcomes to directly rewarding the intrinsic quality of the reasoning \textit{process}.
\end{mdframed}

Existing methods are hamstrung by this flawed paradigm. Supervised fine-tuning produces brittle mimics, while contemporary reinforcement learning approaches, with their myopic focus on final-answer correctness, inadvertently reinforce the very flaws—inconsistency, hallucination, and unstructured thought—that cause reasoning to fail. Our work pioneers an approach centered on reasoning process rewards, using a multi-faceted reward suite to transform reasoning from a random liability into a reliable asset. The following findings chart a new course for the field.

\keyfindingbox{Test-Time inverse scaling should be reframed not as a fundamental law, but as a diagnostic signal for flawed reasoning processes. This issue is fully solvable with process-oriented supervision.}

Our analysis provides a definitive diagnosis for why unguided reasoning is so detrimental. As vividly demonstrated by the base Qwen2.5-Omni model's catastrophic performance collapse (from 68.60\% down to 65.20\% in Tab.~\ref{tab:all_mmau_results}), allowing an untrained model to ``think" longer provides more opportunities for logical errors and hallucinations to compound. Our framework proves this is not an immutable property. By explicitly rewarding internal consistency, CESAR directly targets the root cause of this degradation, resulting in a complete reversal of the phenomenon. This finding suggests that readers encountering test-time inverse scaling should treat it as a clear signal that a model’s reasoning process requires direct, granular intervention, shifting focus from outcome-only rewards to the quality of the cognitive process itself.

\keyfindingbox{Reasoning can be transformed from an unpredictable emergent property into a controllable and engineerable skill, whose quality can be quantitatively measured beyond simple task accuracy.}

A critical question for any RL method is whether the agent is truly learning a skill or simply exploiting the reward. Our work offers two contributions here. Our multi-faceted rewards, particularly those incentivizing structured and logical patterns (App.~\ref{app:Qualitative_Analysis}), act as a cognitive scaffold to guide the model toward desired analytical behaviors. To validate that this guidance cultivates a genuine skill, our AI-as-judge evaluation provides quantitative proof of superior reasoning quality. The commanding win rates of CESAR introduce a valuable and scalable methodology for the field, enabling researchers to move beyond accuracy to rigorously evaluate the thought process itself. Reasoning, therefore, no longer needs to be a matter of chance; it can become a matter of design.

\keyfindingbox{The optimal reasoning budget is not universal but model-specific. This ``reasoning sweet spot" can be unlocked at inference time, but only after a robust reasoning process has been cultivated during training.}

Our introduction of test-time scaling reveals that the value of increased computation is entirely conditional on the quality of the learned policy. For the base model, more computation is actively harmful; for the outcome-only RL model, it yields volatile gains. In stark contrast, because CESAR has learned a coherent reasoning process—calibrated in part by the `Overthinking Penalty`—test-time scaling becomes a powerful, practical optimization lever. It allows us to identify a distinct performance peak—a ``reasoning sweet spot"—that other models cannot reach. This establishes a critical principle: a model must first learn to \textit{think well} before \textit{thinking more} becomes beneficial. This insight naturally leads to a two-stage best practice for practitioners: first, cultivate robust reasoning through process-oriented training, and then employ test-time scaling as an efficient, training-free strategy to identify the model's optimal computational budget at inference.

\keyfindingbox{Cultivating deliberate, step-by-step reasoning creates a powerful synergistic uplift, enhancing both a model's intuitive answering and its foundational perception.}

This finding reveals a deep connection between different modes of cognition. The rigorous process of learning to reason forces the model to organize its understanding of the world more effectively. This enhanced internal representation sharpens its ``fast," intuitive thinking, evidenced by a massive 5.1\% improvement in its direct-answering capability over the base model (73.70\% vs. 68.60\%). The benefits even cascade to the sensory level, improving foundational perception scores on the MMSU benchmark. Better thinking, it turns out, leads to better hearing.

\keyfindingbox{By elevating reasoning to near-human levels, our work acts as a powerful diagnostic for the field, revealing that the primary barrier to progress is a foundational perceptual bottleneck.}

Perhaps our most significant contribution is diagnostic: by successfully addressing high-level reasoning, our work brings the next major barrier into sharp focus. On the MMSU benchmark (Tab.~\ref{tab:mmsu_main_results}), CESAR achieves near-human and even super-human reasoning capabilities (e.g., 88.72\% vs. human 82.16\% in Semantic Reasoning). This very success allows us to clearly identify the next great challenge. The remaining performance gap to humans can be confidently attributed to a different layer of the system: foundational perception, where our model (48.45\%) still lags far behind human acuity (91.24\%). Our work thus transforms the research landscape, providing a clear, data-driven direction to solve this perceptual bottleneck.

\clearpage

\subsection{Limitations}
\label{app:limitation}
Our investigation also sheds light on several limitations, including a fundamental challenge for the field and method-specific considerations for future work.

\paragraph{The Perceptual Bottleneck.}
The primary limitation we identify is a foundational \textbf{perceptual bottleneck} affecting all current models. This issue is paradoxically highlighted by our own model's success; our results on the MMSU benchmark reveal a stark asymmetry where CESAR achieves super-human reasoning capabilities (e.g., 88.72\% in semantic reasoning) while its foundational perception still significantly lags behind human acuity (48.45\% vs. 91.24\%). This demonstrates that even with near-perfect reasoning, a model's performance is ultimately capped by its ability to perceive a high-fidelity representation of the acoustic world. Resolving this is a critical next step for the entire field.

\paragraph{Computational Requirements.}
The GRPO-based training regimen, which requires sampling multiple responses for each input during online optimization, is computationally intensive. One standard training run of ours requires significant GPU resources, and this computational overhead, while justified by the substantial performance gains, may present a barrier to adoption for research groups with limited hardware resources.

\paragraph{Hyperparameter Tuning.}
Introducing a multi-faceted reward suite inevitably brings the challenge of hyperparameter optimization, specifically in balancing the weights of each reward component. We took steps to mitigate this complexity, for instance by normalizing each reward signal to a consistent [0, 1] range. Furthermore, through empirical investigation, we discovered that giving a higher weight to the accuracy reward while keeping other process-oriented rewards equally weighted yielded the best results. This suggests a potential curriculum learning effect: the model first prioritizes optimizing for accuracy—the most direct path to significant reward gains—and then, upon reaching a performance plateau, begins to refine its policy based on the more nuanced signals from the reasoning process rewards. We believe this is a valuable practical insight and encourage readers applying similar multi-reward frameworks to experiment with prioritizing the primary accuracy reward to guide the initial stages of policy optimization.

\clearpage

\subsection{Future Works}
\label{app: future work}

Our work establishes a principled approach to building robust, controllable reasoning in Audio LLMs, addressing the test-time inverse scaling problem that has plagued the field. Having demonstrated that process-oriented training can reliably improve reasoning capabilities, several promising research directions emerge.

\paragraph{The Perceptual Bottleneck Problem.}
With reasoning capabilities now approaching human levels, our results reveal that perceptual limitations have become the primary constraint on overall performance. The audio encoders used in current systems appear to be the main bottleneck preventing further progress. This suggests that developing more sophisticated audio representations—perhaps through self-supervised learning or novel architectural innovations—should be a priority for the community \citep{varcon}. Our improved reasoning capabilities provide a clear benchmark for evaluating whether perceptual improvements translate to better end-to-end performance.

\paragraph{Cross-Modal Applications.}
The success of process-oriented training in audio raises questions about its broader applicability. Testing whether similar principles work for vision, robotics, or other modalities would help determine if we've uncovered domain-specific insights or more general principles of machine reasoning. Early experiments applying our framework to visual question answering or robotic planning could provide valuable insights into the universality of process-oriented approaches.

\clearpage

\subsection{The Use of Large Language Models (LLMs)}
\label{app: llm usage}

In accordance with the conference guidelines, we acknowledge the use of Large Language Models (LLMs) during the preparation of this manuscript. We utilized LLMs for paper writing assistance, specifically for language polishing and improving the clarity and readability of our work. The LLMs assisted in refining linguistic expression, ensuring proper grammar and academic writing style, and enhancing the overall flow of technical content.

All core research contributions, including the novel methodology, experimental design, theoretical analysis, and scientific insights presented in this work, were developed independently by the authors. The LLMs were used solely as writing assistance tools (i.e., for polishing writing) and did not contribute to the conceptual development, experimental validation, or interpretation of results.

\clearpage

\subsection{Ethics}
Our work aims to enhance multimodal reasoning capabilities in audio LLMs without introducing any additional ethical concerns or resolving existing ones.

\clearpage

\section{Experimental Details}
\label{app: Experimental Details}

\subsection{Baseline Methods}
\label{subsec:baselines}
To validate the superiority of our approach, we compare it against a comprehensive set of baselines that represent different training paradigms and model classes.

\paragraph{Base Model}
Our foundational model is \textbf{Qwen2.5-Omni-7B}, a powerful, unified end-to-end multimodal model capable of perceiving diverse inputs including audio, video, and images, and generating both text and speech responses \citep{xu2025qwen25omni}. We evaluate it in two distinct modes: direct-answering (zero-shot) and CoT-prompted. This crucial comparison allows us to empirically diagnose the test-time inverse scaling problem: by contrasting the performance of a powerful but untrained reasoner with and without a reasoning process, we can isolate the performance degradation caused by unguided ``thinking" and establish a clear baseline from which to measure the absolute gains provided by our RL framework.

\paragraph{RL Baseline}
Our most direct competitor is \textbf{Ke-Omni-R} \citep{ke_omni_r}, the current state-of-the-art  audio reasoning model that shares the same \textbf{Qwen2.5-Omni-7B} base architecture and is also trained using the GRPO algorithm. This makes it the perfect control group for our study. However, Ke-Omni-R relies on a simpler Reinforcement Learning from Verifiable Rewards (RLVR) setup, where rewards are based solely on the correctness of the final answer within a concise reasoning trace of fewer than 50 words \citep{ke_omni_r}. This comparison therefore serves as a direct ablation of our novel, multi-faceted reward suite. By contrasting our process-oriented approach with Ke-Omni-R's outcome-only paradigm, we can effectively measure the performance ceiling of existing RL methods and demonstrate the significant improvements unlocked by rewarding the reasoning process itself.

\paragraph{Other Models}
To situate our work in the broader landscape, we also report scores from other leading models. This includes top-performing proprietary systems such as the Gemini series \citep{gemini25} and GPT-4o Audio \citep{gpt4oaudio}, which represent the state-of-the-art in closed-source multimodal AI. Furthermore, we compare against a wide range of open-source audio LLMs that are primarily trained using supervised fine-tuning (SFT) on CoT datasets, such as Audio-Reasoner \citep{liu2025audioreasoner}. This comprehensive comparison ensures that our results are contextualized against the full spectrum of current approaches, from powerful proprietary APIs to various SFT-based methods. For comprehensive details on these models, we refer the reader to their  original papers and the benchmark papers \citep{mmsu,ghosh2025mmar,sakshi2024mmau}.

\clearpage

\subsection{Evaluation Benchmarks}
\label{subsec:benchmarks}
To rigorously validate the generalization capabilities of our framework, we conduct a comprehensive evaluation on several distinct, challenging, and entirely \textbf{out-of-distribution (OOD)} audio understanding benchmarks. None of the audio clips, questions, or underlying tasks in these benchmarks overlap with our training corpus (AVQA). This strict separation ensures that our evaluation measures genuine, transferable reasoning skill, rather than task-specific memorization or reward hacking, thereby providing a true test of our model's ability to reason in novel acoustic scenarios.

\paragraph{MMAU}
We selected the \textbf{MMAU (Massive Multi-Task Audio Understanding and Reasoning Benchmark)} test-mini split \citep{sakshi2024mmau} as our principal testbed due to its unparalleled breadth and focus on expert-level cognition. Comprising approximately 1000 expertly annotated questions, the benchmark is systematically distributed across the three core audio domains: speech, environmental sounds, and music \citep{sakshi2024mmau}. Its design explicitly targets 27 distinct cognitive skills, which are divided into information extraction and complex reasoning categories \citep{sakshi2024mmau}. The significant challenge of MMAU stems from its demand for expert-level, domain-specific knowledge—such as identifying musical chord progressions or decoding phonological sequences—combined with sophisticated reasoning that moves far beyond simple perception. This comprehensive and demanding nature makes it the ideal environment to validate the general and versatile reasoning capabilities cultivated by CESAR, and explains why our framework achieves state-of-the-art performance on this benchmark.

\paragraph{MMSU}
For a granular, diagnostic analysis of spoken language understanding, we utilize the \textbf{MMSU (Massive Multi-task Spoken Language Understanding and Reasoning Benchmark)} \citep{mmsu}, which serves as a surgical tool for dissecting the relationship between high-level cognition and low-level perception. Containing 5,000 audio-question pairs across 47 distinct tasks grounded in established linguistic theory \citep{mmsu}, its unique value lies in the formal bifurcation of all tasks into foundational \textit{Perception} (e.g., identifying falling tones) and higher-level \textit{Reasoning} (e.g., interpreting sarcasm from prosodic cues) \citep{mmsu}. This explicit separation is strategically vital, as it allows us to provide clear, quantitative evidence for our key discovery: CESAR's ability to achieve near human-level performance on the \textit{Reasoning} tasks validates the effectiveness of our training paradigm. Simultaneously, the significant gap that remains on \textit{Perception} tasks, despite some synergistic improvement, provides definitive proof of the ``perceptual bottleneck," clarifying a critical direction for future research.

\paragraph{MMAR}
To stress-test our model's reasoning capabilities under the most demanding conditions, we include an evaluation on \textbf{MMAR (A Challenging Benchmark for Deep Reasoning)} \citep{ghosh2025mmar}, a benchmark specifically designed to probe the limits of deep, multi-step, and compositional reasoning. Its 1,000 tasks are uniquely characterized by longer audio clips (averaging 20 seconds \citep{ghosh2025mmar}) and complex, real-world \textit{mixed-modality} audio, where overlapping sources like speech, background music, and sound effects must be disentangled \citep{ghosh2025mmar}. The primary difficulty of MMAR lies in its demand for sustained temporal reasoning and the ability to perform multi-hop inferences on composite acoustic scenes, a task that often requires graduate-level domain knowledge \citep{ghosh2025mmar}. We chose MMAR to prove that the reasoning skills cultivated by CESAR are not brittle but robust and scalable. By succeeding here, we demonstrate that our framework builds a durable cognitive capability that holds up under extreme complexity, providing powerful, supplementary evidence of our model's advanced reasoning prowess, with detailed results presented in App. ~\ref{app: Benchmark Results on MMAR}.

\clearpage

\subsection{Training Data and Augmentation Strategy}
\label{subsec:training_data_aug}

\paragraph{Training Data.} The primary training corpus for CESAR is the \textbf{AVQA} dataset. To ensure a fair comparison, our main RL baseline, Ke-Omni-R, also uses AVQA as its foundation. However, it is crucial to note that Ke-Omni-R supplements its training with the specialist \textbf{MusicBench} dataset. Despite not using this in-domain music data, CESAR still outperforms Ke-Omni-R on the music tasks of the MMAU benchmark (73.05\% vs. 70.06\%). This provides strong evidence that cultivating a general, robust reasoning process enhances multimodal generalization, allowing the model to effectively transfer its learned analytical skills to specialized domains even without explicit in-domain training data.

\paragraph{Systematic Data Augmentation via Question Rephrasing.} To enhance model robustness and prevent the learning of superficial textual correlations, we employ a systematic data augmentation scheme. This method expands our training corpus by generating multiple linguistic variations for each question while preserving the ground-truth answer, thereby compelling the model to learn the underlying reasoning task rather than shallow text patterns. Formally, for each instance $(a_i, q_i, \mathcal{C}_i, y_i) \in \mathcal{D}$, we apply a set of answer-invariant transformation templates $\mathcal{T} = \{T_1, \dots, T_M\}$. Each transformation $T_k$ generates a new question $q'_{i,k} = T_k(q_i, \mathcal{C}_i)$, creating a new training sample $(a_i, q'_{i,k}, \mathcal{C}_i, y_i)$.

Our approach uses simple but effective template-based transformations that reframe questions to target specific reasoning capabilities. For instance, an original question like ``What are the main sources of sound in this video?" with choices [motorboat, bus, train, truck] is transformed using capability-specific templates:
\begin{itemize}
    \item \textbf{Temporal Reasoning:} ``Which sound source appears most prominently in the temporal sequence: \{choices\}?"
    \item \textbf{Counting Tasks:} ``Which option has the highest occurrence frequency among: \{choices\}?"
    \item \textbf{Comparative Analysis:} ``Which sound demonstrates the strongest relationship with other audio elements: \{choices\}?"
\end{itemize}
This strategy systematically expands training diversity, forcing the model to develop generalizable reasoning skills that contribute directly to its robust performance.

\clearpage

\subsection{Training Hyperparameters and Prompting Configuration}
\label{subsec:hyperparams}

\paragraph{Training Pipeline and Hyperparameters.} Our training pipeline is built upon the Qwen2.5-Omni-7B model, which we fine-tune using \textbf{Group Relative Policy Optimization (GRPO)} \citep{shao2024grpo}. To ensure a fair comparison and isolate the impact of our proposed reasoning process rewards, our core GRPO hyperparameters (e.g., KL coefficient $\beta$, batch size, learning rate) are kept consistent with those of the RLVR baseline, Ke-Omni-R \citep{ke_omni_r}. This approach prioritizes methodological clarity and reproducibility. For a detailed breakdown of these specific hyperparameter values, we refer the reader to the original Ke-Omni-R work \citep{ke_omni_r}. The optimization process uses the \textbf{AdamW} optimizer with a learning rate of \textbf{1e-5} and a global batch size of \textbf{32}, sampling $\boldsymbol{K=8}$ responses per input for each GRPO step.

\paragraph{Inference Prompts} 
We adopt the prompt template directly from our primary baseline, Ke-Omni-R, to ensure a fair and direct comparison. This shared template instructs the model to follow a strict, two-part output format, namely: (1)~\textit{Generating Reasoning Traces within \texttt{<think>...</think>} (less than \{max\_think\_len\} words)}, and subsequently (2)~\textit{Generating the Final Answer within \texttt{<answer>...</answer>}}. While the prompt structure is shared, the crucial distinction—and a core component of our methodology—lies in its application during evaluation. Whereas Ke-Omni-R reports performance at a fixed, static reasoning length, we leverage the \texttt{\{max\_think\_len\}} parameter to perform our test-time scaling analysis (see Sec. ~\ref{app:test-time scaling_analysis}). By systematically evaluating the model across the full spectrum of values, we are able to not only demonstrate robustness against the test-time inverse scaling problem but also to identify the optimal, model-specific ``reasoning sweet spots" that unlock peak performance, providing a much richer understanding of a model's true capabilities.

\paragraph{AI-as-Judge Evaluation Prompts.} 
To quantitatively assess reasoning quality, we employed an AI-as-judge framework using a SOTA multimodal LLM (GPT-4o Audio). The evaluation prompt instructed the judge to perform a head-to-head comparison between the reasoning traces of two models. The judge's decision was guided by specific criteria, including logical coherence, faithfulness to acoustic evidence, and the overall soundness of the analytical path, with a focus on the process rather than just the final answer's correctness. The complete prompt and detailed methodology are provided in App.~\ref{app:llm_judge_analysis}.

\paragraph{Reward Configuration}
To ensure a fair and direct comparison with the RL baseline, we align our core GRPO training parameters with those of Ke-Omni-R. The critical distinction lies in our reward configuration. After exploring various hyperparameter settings, we identified a simple yet remarkably effective weighting scheme for the components in \eqref{equ: tot rewards}: the accuracy reward weight ($\alpha_{1}$) is set to \textbf{5.0}, while the weights for all other process-oriented rewards (consistency, keywords, overthinking penalty) are set to \textbf{1.0}. This configuration maintains a strong optimization pressure towards generating correct final answers, while the process rewards act as crucial regularizers and fine-grained guides. They shape the reasoning trajectories without overpowering the primary objective of correctness. As substantiated by our ablation study (App.~\ref{app:Ablation_Results}), the thoughtful \textit{design} of these reward functions, rather than their specific weightings, is the primary driver of performance, demonstrating the robustness of our overall framework.

\clearpage

\subsection{Computational Resources}
\label{subsec:compute}
All reinforcement learning experiments were performed on a  high-performance computing cluster equipped with 8 NVIDIA H200 GPUs, each providing 141GB of HBM3e memory. A standard training run for our final model on the augmented AVQA dataset concluded in approximately 2-3 days on this infrastructure.

\clearpage

\subsection{Detailed Keywords}
\label{app:detailed_keywords}

The \textbf{Keywords Reward} ($R_{\text{keywords}}$) is a central component of our process-oriented supervision framework, engineered to guide the model toward generating reasoning traces that are structured, logical, and domain-aware. This reward is calculated as a composite score that aggregates signals from three distinct categories: structured analytical patterns, logical rigor indicators, and domain-specific terminology. To implement this, we programmatically scan each generated reasoning trace for the presence of specific keywords and patterns. The detection mechanism employs a combination of \textbf{simple string matching} for exact phrases (e.g., \texttt{considering the options}, \texttt{is consistent with}) and \textbf{regular expressions} for more flexible patterns (e.g., numbered lists like \texttt{1.}, \texttt{2.}). Each detected term or pattern from our predefined taxonomies contributes positively to the final reward score, thereby explicitly incentivizing the model to construct more sophisticated and coherent reasoning processes. The comprehensive taxonomies of these keywords and phrases, broken down by their function, are detailed in Tables \ref{tab:keywords_pattern}, \ref{tab:keywords_reasoning}, and \ref{tab:keywords_domain}.

\begin{table}[!htbp]
\centering
\caption{Keywords for Structured Analytical Patterns ($R_{\text{pattern}}$).}
\label{tab:keywords_pattern}
\begin{tabular}{@{}l p{0.35\linewidth} p{0.35\linewidth}@{}}
\toprule
\textbf{Category} & \textbf{Description} & \textbf{Example Keywords / Phrases} \\
\midrule
Sequential Organization & Indicates a step-by-step analytical process or temporal ordering. & \texttt{first}, \texttt{second}, \texttt{then}, \texttt{next}, \texttt{finally}, \texttt{step 1}, \texttt{1.}, \texttt{2.} \\
\addlinespace
Comparative Analysis & Phrases used for comparing and contrasting different options or ideas. & \texttt{rather than}, \texttt{compared to}, \texttt{in contrast to}, \texttt{on the other hand} \\
\addlinespace
Systematic Evaluation & Suggests a methodical review and elimination of the provided choices. & \texttt{considering the options}, \texttt{evaluating each choice}, \texttt{among the options} \\
\addlinespace
Explicit Justification & Language that directly justifies the selection of the final answer. & \texttt{most suitable}, \texttt{the best fit}, \texttt{fits the description best} \\
\bottomrule
\end{tabular}
\end{table}

\begin{table}[!htbp]
\centering
\caption{Keywords for Logical Rigor \& Causal Reasoning ($R_{\text{logic}}$).}
\label{tab:keywords_reasoning}
\begin{tabular}{@{}l p{0.35\linewidth} p{0.35\linewidth}@{}}
\toprule
\textbf{Category} & \textbf{Description} & \textbf{Example Keywords / Phrases} \\
\midrule
Premise \& Deduction & Establishes a logical premise and draws a conclusion from it. & \texttt{given}, \texttt{based on}, \texttt{since}, \texttt{therefore}, \texttt{thus}, \texttt{hence}, \texttt{so} \\
\addlinespace
Evidential Support & Links acoustic evidence from the audio signal to an inference. & \texttt{indicates}, \texttt{suggests}, \texttt{is consistent with}, \texttt{as evidenced by} \\
\addlinespace
Hypothetical Reasoning & Terms used for suppositions or stating general principles. & \texttt{assume}, \texttt{suppose}, \texttt{typically}, \texttt{generally}, \texttt{it is likely that} \\
\bottomrule
\end{tabular}
\end{table}

\begin{table}[!htbp]
\centering
\caption{Keywords for Domain Knowledge Integration ($R_{\text{domain}}$).}
\label{tab:keywords_domain}
\begin{tabular}{@{}l p{0.3\linewidth} p{0.3\linewidth}@{}}
\toprule
\textbf{Category} & \textbf{Description} & \textbf{Example Keywords / Phrases} \\
\midrule
Acoustic Properties & Basic terminology related to the physical properties of sound. & \texttt{sound}, \texttt{audio}, \texttt{noise}, \texttt{pitch}, \texttt{volume}, \texttt{timbre}, \texttt{rhythm}, \texttt{frequency} \\
\addlinespace
Environmental \& Animal Sounds & Vocabulary for specific non-speech, non-music sound events. & \texttt{bell}, \texttt{ring}, \texttt{hooves}, \texttt{engine}, \texttt{siren}, \texttt{animal}, \texttt{clip-clop}, \texttt{moo} \\
\addlinespace
Musical Concepts & Specialized terminology for analyzing musical content. & \texttt{chord}, \texttt{note}, \texttt{melody}, \texttt{harmony}, \texttt{instrument}, \texttt{major}, \texttt{minor}, \texttt{P5} \\
\addlinespace
Speech Analysis & Terms used to describe and analyze human vocal characteristics. & \texttt{voice}, \texttt{speech}, \texttt{tone}, \texttt{intonation}, \texttt{male}, \texttt{female}, \texttt{shouting}, \texttt{whisper} \\
\bottomrule
\end{tabular}
\end{table}

\clearpage

\subsection{Reproducibility}
\label{app: Reproducibility}
We have made comprehensive efforts to ensure reproducibility of our work. Our complete methodology is detailed in Section~\ref{sec:methodology}, with step-by-step algorithmic implementation provided in Appendix~\ref{app:pseudocode}. All experimental configurations are thoroughly documented in Section~\ref{sec:experiments}, with hyperparameter settings specified in Section~\ref{subsec:hyperparams}. As Appendix~\ref{app: Experimental Details}, our training pipeline builds upon the open-source codebase (i.e., Ke-Omni-R \citep{ke_omni_r}) using publicly available base models (i.e., Qwen2.5-Omni-7B \citep{xu2025qwen25omni}) and training datasets. Data augmentation procedures are described in Section~\ref{subsec:training_data_aug}. Evaluation benchmarks are all publicly available. Additional implementation details, including computational requirements and reward function specifications, are provided in Appendix~\ref{app: Experimental Details}. All source code and trained models will be made publicly available upon publication to facilitate reproducibility and future research.

\clearpage

\section{Algorithm Pseudocode}
\label{app:pseudocode}

In this section, we  provide the detailed pseudocode for the CESAR framework. Algorithm \ref{alg:cesar_training} outlines the main online reinforcement learning loop using Group Relative Policy Optimization (GRPO). To enhance clarity, we use the superscript 'ex' (e.g., $\mathcal{L}_{\text{GRPO}}^{\text{ex}}$) to denote a value calculated for a single training \textit{example}, distinguishing it from values aggregated over an entire mini-batch. Algorithm \ref{alg:reward_calculation} then specifies the computation of our multi-faceted, process-oriented reward, which is central to cultivating robust reasoning capabilities.

\begin{algorithm}[!ht]
\caption{CESAR Training via Group Relative Policy Optimization (GRPO)}
\label{alg:cesar_training}
\begin{algorithmic}[1]
\State \textbf{Require:} Audio LLM policy $\pi_\theta$ to be fine-tuned, reference policy $\pi_{\text{ref}}$.
\State \textbf{Require:} Training dataset $\mathcal{D} = \{(a_i, q_i, \mathcal{C}_i, y_i)\}_{i=1}^N$.
\State \textbf{Require:} Number of samples per input $K$.
\State \textbf{Require:} Reward weights $\{\alpha_j\}_{j=1}^5$, learning rate $\eta$, KL regularization weight $\beta$.

\State Initialize policy parameters $\theta$ from a pre-trained Audio LLM.
\For{each training iteration}
    \State Sample a mini-batch $B = \{(a, q, \mathcal{C}, y)\}$ from $\mathcal{D}$.
    \State Initialize gradients $\nabla_\theta \mathcal{L} \gets 0$.
    \For{each training example $(a, q, \mathcal{C}, y)$ in $B$}
        \State // Step 1: Sample K responses from the current policy $\pi_\theta$.
        \State Sample a set of $K$ responses $\mathcal{S} = \{s^{(k)} = (t^{(k)}, \hat{y}^{(k)})\}_{k=1}^K \sim \pi_\theta(\cdot | a, q, \mathcal{C})$.
        
        \State // Step 2: Calculate the total reward for each of the K responses.
        \State Initialize a rewards list $R \gets []$.
        \For{$k = 1$ to $K$}
            \State $R_{\text{total}}^{(k)} \gets \Call{CalculateTotalReward}{s^{(k)}, y, q, \mathcal{C}}$ \Comment{See Algorithm \ref{alg:reward_calculation}}
            \State Append $R_{\text{total}}^{(k)}$ to $R$.
        \EndFor
        
        \State // Step 3: Compute the advantage using the mean reward as a baseline.
        \State $\bar{R} \gets \frac{1}{K} \sum_{j=1}^K R_{\text{total}}^{(j)}$.
        \State Initialize policy gradient loss for the example $\mathcal{L}_{\text{PG}}^{\text{ex}} \gets 0$.
        \For{$k = 1$ to $K$}
            \State $A(s^{(k)}) \gets R_{\text{total}}^{(k)} - \bar{R}$. \Comment{Advantage of response $k$}
            \State $\mathcal{L}_{\text{PG}}^{\text{ex}} \gets \mathcal{L}_{\text{PG}}^{\text{ex}} - A(s^{(k)}) \log \pi_\theta(s^{(k)} | a, q, \mathcal{C})$.
        \EndFor
        
        \State // Step 4: Calculate the full loss and accumulate gradients.
        \State $\mathcal{L}_{\text{KL}} \gets \mathbb{E}_{\pi_\theta} \left[ \log \frac{\pi_\theta(\cdot | a, q, \mathcal{C})}{\pi_{\text{ref}}(\cdot | a, q, \mathcal{C})} \right]$.
        \State $\mathcal{L}_{\text{GRPO}}^{\text{ex}} \gets \frac{1}{K}\mathcal{L}_{\text{PG}}^{\text{ex}} + \beta \cdot \mathcal{L}_{\text{KL}}$.
        \State Accumulate gradients: $\nabla_\theta \mathcal{L} \gets \nabla_\theta \mathcal{L} + \nabla_\theta \mathcal{L}_{\text{GRPO}}^{\text{ex}}$.
    \EndFor
    
    \State // Step 5: Update the policy parameters.
    \State $\theta \gets \theta - \eta \cdot \frac{1}{|B|} \nabla_\theta \mathcal{L}$.
\EndFor
\State \textbf{return} Optimized policy parameters $\theta$.
\end{algorithmic}
\end{algorithm}

\begin{algorithm}[!ht]
\caption{Multi-Faceted Reward Calculation}
\label{alg:reward_calculation}
\begin{algorithmic}[1]
\Function{CalculateTotalReward}{$s, y, q, \mathcal{C}$}
    \State \textbf{Input:} A single response $s=(t, \hat{y})$, ground-truth answer $y$, question $q$, choices $\mathcal{C}$.
    \State \textbf{Input:} Reward weights $\{\alpha_j\}_{j=1}^5$.
    
    \State // --- 1. Verifiable Rewards ---
    \State $R_{\text{acc}} \gets \mathbb{I}[\hat{y} = y]$. \Comment{Accuracy}
    \State $R_{\text{format}} \gets \mathbb{I}[\text{ValidFormat}(s)]$. \Comment{XML structure compliance}
    
    \State // --- 2. Reasoning Process Rewards ---
    \State $Q \gets (q, \mathcal{C})$. \Comment{Full question context}
    \State $R_{\text{consistency}} \gets \text{Sim}_{\text{semantic}}(t, \hat{y}) + \text{Sim}_{\text{semantic}}(t, Q)$. \Comment{Semantic alignment}
    
    \State $R_{\text{pattern}} \gets \Call{CalculateKeywordScore}{t, \text{PatternKeywords}}$. \Comment{See Table \ref{tab:keywords_pattern}}
    \State $R_{\text{logic}} \gets \Call{CalculateKeywordScore}{t, \text{LogicKeywords}}$. \Comment{See Table \ref{tab:keywords_reasoning}}
    \State $R_{\text{domain}} \gets \Call{CalculateKeywordScore}{t, \text{DomainKeywords}}$. \Comment{See Table \ref{tab:keywords_domain}}
    \State $R_{\text{keywords}} \gets R_{\text{pattern}} + R_{\text{logic}} + R_{\text{domain}}$.
    
    \State $R_{\text{overthinking}} \gets 1 - \frac{\text{length}(t)}{L_{\text{max\_output}}}$. \Comment{Penalty for verbosity}
    
    \State // --- 3. Compute Total Weighted Reward ---
    \State $R_{\text{total}} \gets \alpha_1 R_{\text{acc}} + \alpha_2 R_{\text{format}} + \alpha_3 R_{\text{consistency}} + \alpha_4 R_{\text{keywords}} + \alpha_5 R_{\text{overthinking}}$.
    
    \State \textbf{return} $R_{\text{total}}$.
\EndFunction
\end{algorithmic}
\end{algorithm}

\clearpage

\section{Additional Experimental Results}
\label{app: add results}

\subsection{Benchmark Results on MMAU}
\label{app:mmau_results}

\begin{table}[!htbp]
\centering
\caption{MMAU Test-mini Benchmark Results. We evaluate our method against state-of-the-art proprietary and open-source audio models. Best scores are highlighted in \colorbox{lightblue}{\textbf{blue}}, second-best scores in \colorbox{lightgreen}{\textbf{green}}. Accuracy (\%) is reported. We report the performance of Qwen2.5-Omni-7B \citep{xu2025qwen25omni} and Ke-Omni-R \citep{ke_omni_r} from our own reproductions under the same protocol; all other baseline results are taken from the MMAU paper \citep{sakshi2024mmau}.}
\label{tab:all_mmau_results}
\resizebox{0.9\textwidth}{!}{
\begin{tabular}{l|c|ccc|c}
\toprule
\textbf{Method} & \textbf{Reasoning} & \textbf{Sound} & \textbf{Music} & \textbf{Speech} & \textbf{Total Accuracy} \\
\midrule
\multicolumn{6}{c}{\cellcolor{blue!15}\textbf{Our Proposed Methods}} \\
\midrule
\textbf{CESAR} & \cmark & \colorbox{lightblue}{\textbf{83.48}} & \colorbox{lightblue}{\textbf{73.05}}  & 74.77 & \colorbox{lightblue}{\textbf{77.10}} \\
\textbf{CESAR} & \xmark & 79.88 & 67.96 & 73.27 & 73.70 \\
\textbf{CESAR w/o OP} & \cmark & \colorbox{lightgreen}{\textbf{81.98}}  & 70.06 & \colorbox{lightblue}{\textbf{77.48}} & \colorbox{lightgreen}{\textbf{76.50}} \\
\textbf{CESAR w/o OP} & \xmark & 80.48 & 70.06 & 74.47 & 75.00 \\
\midrule
\multicolumn{6}{c}{\cellcolor{gray!15}\textbf{RL Baseline Methods}} \\
\midrule
Ke-Omni-R & \cmark & 79.28 & 70.06 & 74.47 & 74.60 \\
Ke-Omni-R & \xmark & 78.38 &  \colorbox{lightgreen}{\textbf{70.96}}  & 74.17 & 74.50 \\
\midrule
\multicolumn{6}{c}{\cellcolor{gray!15}\textbf{Base Models}} \\
\midrule
Qwen2.5-Omni-7B & \cmark & 69.07 & 59.58 & 66.97 & 65.20 \\
Qwen2.5-Omni-7B & \xmark & 72.37 & 64.37 & 69.07 & 68.60 \\
\midrule
\multicolumn{6}{c}{\cellcolor{orange!15}\textbf{Proprietary Models}} \\
\midrule
Gemini 2.5 Pro & - & 75.08 & 68.26 & 71.47 & 71.60 \\
Gemini 2.5 Flash & - & 73.27 & 65.57 & \colorbox{lightgreen}{\textbf{76.58}} & 71.80 \\
Gemini 2.0 Flash & - & 71.17 & 65.27 & 75.08 & 70.50 \\
GPT-4o Audio & - & 64.56 & 56.29 & 66.67 & 62.50 \\
GPT-4o mini Audio & - & 50.75 & 39.22 & 69.07 & 53.00 \\
\midrule
\multicolumn{6}{c}{\cellcolor{red!15}\textbf{Open-Source Audio Models}} \\
\midrule
Kimi-Audio  & - & 75.68 & 66.77 & 62.16 & 68.20 \\
Audio Reasoner & - & 67.87 & 69.16 & 66.07 & 67.70 \\
Phi-4-multimodal & - & 65.47 & 64.37 & 67.27 & 65.70 \\
Audio Flamingo 2  & - & 71.47 &  \colorbox{lightgreen}{\textbf{70.96}} & 44.74 & 62.40 \\
Qwen2-Audio-Instruct & - & 67.27 & 56.29 & 55.26 & 59.60 \\
\bottomrule
\end{tabular}
}
\end{table}

Our evaluation on the MMAU Test-mini benchmark, with comprehensive results presented in Tab.~\ref{tab:all_mmau_results}, not only establishes a new state-of-the-art performance but, more importantly, a deeper analysis of these results uncovers several critical insights into the nature of reasoning in Audio LLMs and the means by which it can be effectively cultivated.

\paragraph{An Insight on Scaling Reasoning Process vs. Scaling Parameters.} The first critical insight from these results emerges from the clear superiority of scaling up the \textit{reasoning process} over simply scaling up model parameters. CESAR, at just 7B parameters, achieves its state-of-the-art 77.10\% accuracy not by possessing a larger architecture, but by effectively scaling its cognitive process at inference time—a latent capability unlocked by our training and fully realized through test-time analysis of reasoning length. This performance decisively surpasses that of proprietary models like the Gemini 2.5 series and GPT-4o Audio, whose primary scaling axis is their vast parameter count. This finding strongly suggests that a new paradigm for performance enhancement is not only viable but superior: instead of relying on brute-force parameter scaling, strategically cultivating and dynamically scaling a model's reasoning process offers a more efficient and effective path to advanced capabilities.

\paragraph{The Symbiotic Rise of Reasoning and Intuition.} Beyond sheer performance, the results reveal a more subtle and perhaps more profound insight into the effects of our training paradigm. Our process-oriented RL training not only enhances the model's explicit, step-by-step reasoning but also substantially elevates its direct, intuitive answering capability. This is evident as our model without reasoning (`CESAR w/o Reasoning`) achieves 73.70\% accuracy, a score far superior to the base model's 68.60\% in the same setting. This suggests that incentivizing high-quality reasoning pathways does more than just teach a model to generate a thinking monologue; it fundamentally refines the model's core representations of the acoustic world. This discovery carries significant implications for practical deployment, as it enables highly efficient inference through fast, direct answers while retaining a powerful, on-demand reasoning faculty for more complex challenges.

\paragraph{Transforming Reasoning from Detriments into Gains.} Finally, the data provides a clear narrative on the evolution of reasoning itself. The base Qwen2.5-Omni-7B model exemplifies the critical problem of uncontrolled reasoning, where performance catastrophically drops by 3.4 points when it is prompted to ``think" (from 68.60\% to 65.20\%). This is a textbook case of the test-time inverse scaling problem. In stark contrast, CESAR systematically reverses this trend, gaining a robust 3.4 points under the same conditions (from 73.70\% to 77.10\%). This transforms the act of reasoning from a high-risk gamble into a reliable and scalable tool for performance enhancement. With this newfound stability, reasoning is no longer an unpredictable behavior but a controllable capability. We therefore suggest that practitioners can now confidently employ reasoning and, by combining it with test-time scaling analysis, identify the model-specific ``sweet spot" to unlock its full, calibrated potential. This marks a pivotal shift, firmly establishing reasoning as a core asset for advancing multimodal understanding.

\clearpage

\subsection{Beyond Aggregate Scores: A Task-Level Analysis of Controllable Reasoning Capability}
\label{app: Beyond Aggregate Scores: A Task-Level Analysis of Controllable Reasoning}

While aggregate scores (Table \ref{tab:all_mmau_results}) establish the state-of-the-art performance of our method, they fundamentally mask the most profound discovery of our work: the systematic emergence of controllable reasoning archetypes. A granular, multi-faceted analysis is essential to understand not merely the quantitative superiority, but the qualitative transformation of reasoning from a randomly emergent phenomenon into a precisely engineerable capability.

\begin{figure}[!htbp]
    \centering
    \includegraphics[width=0.5\textwidth]{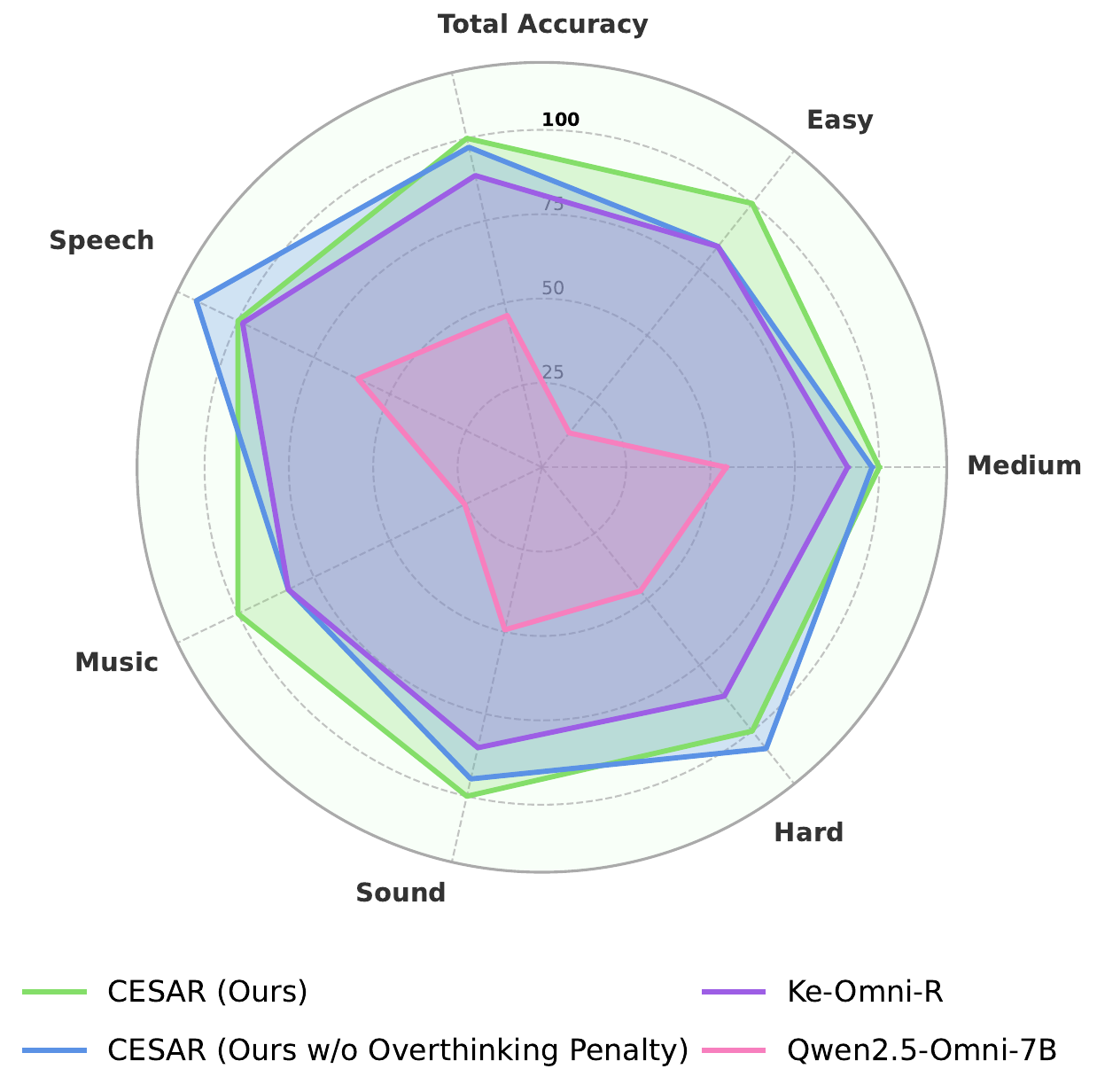}
    \caption{Normalized multi-dimensional performance comparison on the MMAU Test-mini benchmark. Performance is scaled relative to CESAR (Ours), which constitutes the 100\% baseline on each axis. This visualization reveals the emergence of distinct reasoning specializations across task difficulties.}
    \label{fig:radar_analysis}
\end{figure}

\begin{figure}[!htbp]
    \centering
    \includegraphics[width=0.5\textwidth]{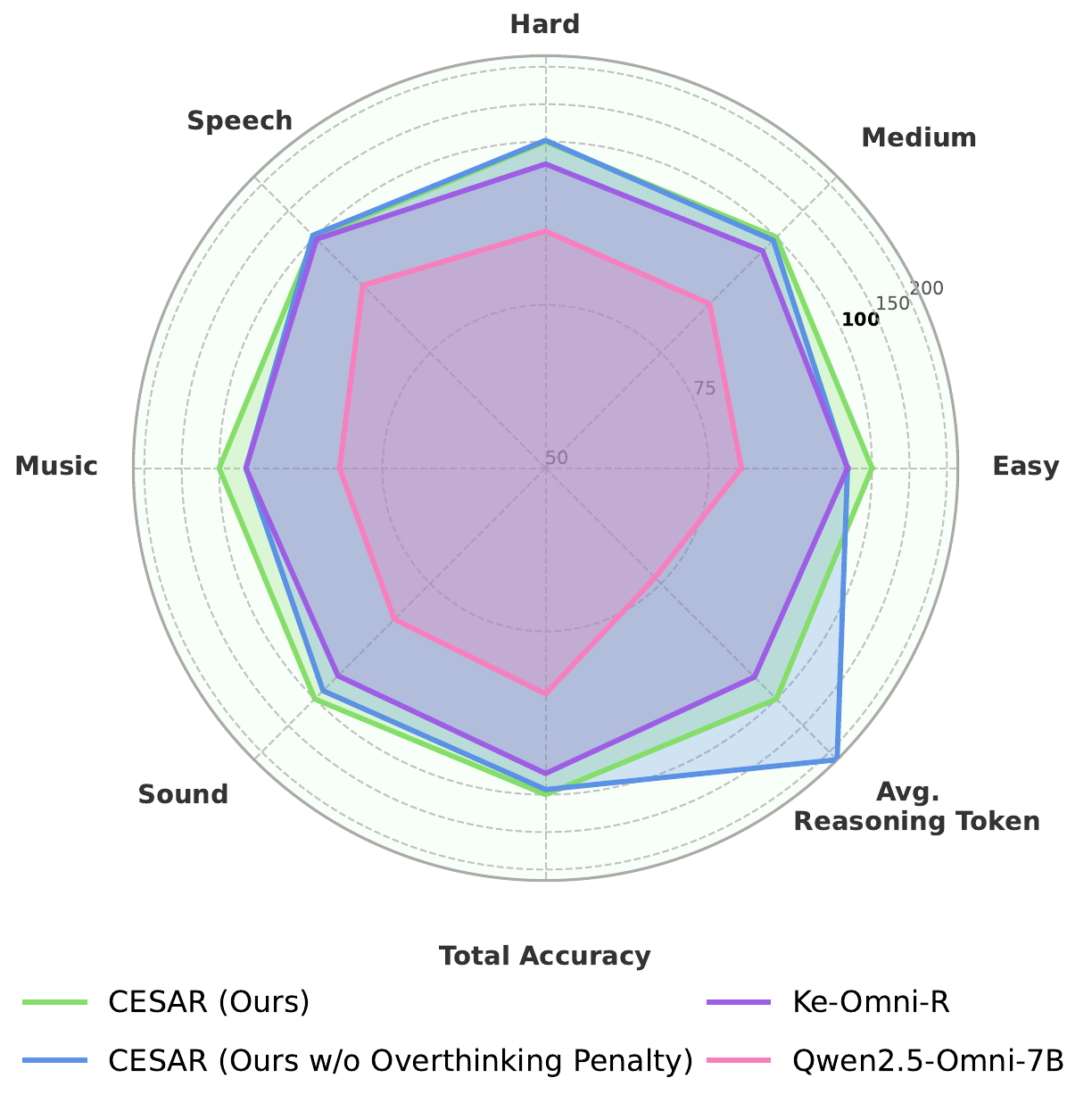}
    \caption{Extended radar analysis including reasoning token efficiency. This enhanced visualization provides clearer visibility of the fundamental trade-offs between reasoning depth and computational cost, unveiling two fundamentally different reasoning archetypes optimized for distinct cognitive scenarios.}
    \label{fig:radar_tokens}
\end{figure}

To reveal this transformation, we present a comprehensive radar chart analysis in Fig. \ref{fig:radar_analysis}, where the performance of our full model, CESAR (Ours), serves as the 100\% baseline across seven key evaluation dimensions. This normalization strategy exposes not merely superior performance, but the emergence of fundamentally different cognitive architectures that our process-oriented framework has systematically cultivated.

\paragraph{The Emergence of Two Reasoning Archetypes: A Tale of Cognitive Specialization.}
The radar charts provide irrefutable visual evidence for our central hypothesis while simultaneously unveiling an unexpected discovery: process-oriented supervision does not simply improve reasoning—it enables the systematic engineering of distinct reasoning archetypes. The performance polygons of our two CESAR variants (green and blue) dominate both visualizations (Fig. \ref{fig:radar_analysis} and Fig. \ref{fig:radar_tokens}), covering significantly larger areas than the strong RL baseline Ke-Omni-R (purple) and the base model Qwen2.5-Omni-7B (pink). This validates our core claim that genuine reasoning emerges only when supervision targets the reasoning \textit{process}, not just final outcomes.

However, the most profound insight emerges from the striking divergence between our two CESAR variants—a divergence that reveals the existence of a fundamental trade-off in reasoning system design. The CESAR (w/o Overthinking Penalty) variant exhibits a distinctive cognitive profile: exceptional performance on \textit{Hard} tasks, consistently exceeding even our 100\% baseline, but at a deliberate cost of efficiency on simpler problems. This model represents what we term a \textbf{depth specialist}—an archetype that favors exhaustive, thorough analysis over computational efficiency.

In stark contrast, the full CESAR (Ours) model demonstrates a fundamentally different cognitive architecture. It forms a perfectly calibrated profile across all difficulty levels, showing exceptional stability on \textit{Easy} and \textit{Medium} tasks while maintaining competitive performance on \textit{Hard} problems. This represents a \textbf{calibrated generalist}—an archetype optimized for consistent, efficient reasoning across diverse problem complexities.

\paragraph{The Fundamental Performance-Depth Trade-off: Efficiency vs. Thoroughness.}
The extended analysis including reasoning token efficiency (Fig. \ref{fig:radar_tokens}) exposes the computational mechanics underlying this cognitive divergence. The depth specialist achieves its superior performance on challenging tasks by investing substantially more reasoning tokens—engaging in extensive, multi-step analytical processes that thoroughly explore problem spaces. Conversely, the calibrated generalist demonstrates remarkable efficiency, achieving comparable overall performance while operating under strict computational constraints imposed by the overthinking penalty.

This discovery challenges conventional assumptions about reasoning optimization and reveals a fundamental principle: \textbf{there exists an inherent tension between reasoning depth and computational efficiency, and optimal performance emerges when models are explicitly trained to navigate this trade-off according to task requirements}. The depth specialist excels precisely because it is willing to invest computational resources in exhaustive analysis when problems demand it. The generalist succeeds by learning to apply just enough analytical effort to solve problems effectively without wasteful over-elaboration.

\paragraph{Engineering Controllable Cognitive Architectures.}
The emergence of these distinct reasoning archetypes represents far more than an interesting experimental observation—it provides definitive proof that reasoning has been transformed from an unpredictable emergent property into a controllable, engineerable capability. The stark differences between our variants are not accidental byproducts of training, but the direct result of our process-oriented reward architecture functioning as precision engineering tools for cognitive behavior.

By systematically modulating a single reward component—the overthinking penalty—we have demonstrated the ability to produce models with predictably different reasoning profiles, each optimized for distinct deployment scenarios. The depth specialist thrives in research environments where thorough analysis justifies computational cost, making it ideal for complex analytical tasks requiring maximum cognitive depth. The calibrated generalist excels in production systems where efficiency and consistency are paramount, delivering reliable performance across diverse problem types without excessive resource consumption.

This unprecedented level of control demonstrates that CESAR transcends being merely a high-performing model—it represents a comprehensive framework for engineering the next generation of controllable audio reasoners with specific, desirable cognitive traits. We have moved decisively beyond the traditional paradigm of hoping for beneficial reasoning patterns to emerge spontaneously, entering an era where cognitive capabilities can be systematically specified, implemented, and validated with the same precision as traditional software systems.

The dual visualization provides definitive empirical proof that our methodology enables the systematic exploration of the reasoning capability space, allowing researchers and practitioners to engineer cognitive systems precisely tailored to their specific requirements and constraints. This work establishes both the theoretical foundation and practical methodology for controllable AI development, where reasoning behavior becomes a design parameter rather than an emergent accident.

\clearpage

\subsection{Benchmark Results on MMSU}
\label{app:mmsu_results}

\begin{table*}[!htbp]
\centering
\caption{MMSU Benchmark Results. We evaluate our method against state-of-the-art audio models across perception and reasoning tasks in speech understanding. Best scores are highlighted in \colorbox{lightblue}{\textbf{blue}}, second-best scores in \colorbox{lightgreen}{\textbf{green}}. All results show accuracy (\%). Human performance is included as an upper bound reference. We report the performance of Ke-Omni-R \citep{ke_omni_r} from our own reproductions under the same protocol; all other baseline results are taken from the MMSU paper \citep{mmsu} (including Qwen2.5-Omni-7B \citep{xu2025qwen25omni}.). The results of Audio Flamingo 3 are taken from their paper \citep{flam3_nvidia}.}
\label{tab:mmsu_results}
\resizebox{\textwidth}{!}{
\begin{tabular}{l|cccc|cccc|c}
\toprule
\multirow{2}{*}{\textbf{Models}} & \multicolumn{4}{c|}{\textbf{Perception Tasks}} & \multicolumn{4}{c|}{\textbf{Reasoning Tasks}} & \multirow{2}{*}{\textbf{Overall}} \\
\cmidrule{2-9}
& \textbf{Semantics} & \textbf{Phonology} & \textbf{Paralinguistics} & \textbf{Avg} & \textbf{Semantics} & \textbf{Phonology} & \textbf{Paralinguistics} & \textbf{Avg} & \\
\midrule
\multicolumn{10}{c}{\cellcolor{blue!15}\textbf{Our Proposed Method}} \\
\midrule
\textbf{CESAR} & \colorbox{lightgreen}{\textbf{60.16}} & 50.16 & 39.50 & \colorbox{lightgreen}{\textbf{48.45}} & \colorbox{lightblue}{\textbf{88.72}} & 80.66 & 57.01 & \colorbox{lightgreen}{\textbf{81.07}} & \colorbox{lightgreen}{\textbf{64.24}} \\
\midrule
\multicolumn{10}{c}{\cellcolor{gray!15}\textbf{Audio RL Baseline}} \\
\midrule
Ke-Omni-R & 58.74 & 46.31 & \colorbox{lightgreen}{\textbf{40.50}} & 47.09 & 86.82 & 74.31 & \colorbox{lightgreen}{\textbf{60.00}} & 78.06 & 62.08 \\
\midrule
\multicolumn{10}{c}{\cellcolor{orange!15}\textbf{Proprietary Models}} \\
\midrule
Gemini 1.5 Pro & 57.06 & \colorbox{lightgreen}{\textbf{53.60}} & 31.23 & 46.10 & 79.47 & \colorbox{lightgreen}{\textbf{83.46}} & 46.33 & 76.16 & 60.68 \\
Qwen2.5-Omni & 55.12 & 37.33 & 39.35 & 42.50 & \colorbox{lightgreen}{\textbf{88.00}} & 81.37 & 48.36 & 79.83 & 60.57 \\
Kimi-Audio & 57.64 & 42.30 & 35.74 & 43.52 & 81.77 & 76.65 & 55.22 & 76.03 & 59.28 \\
GPT-4o Audio & 59.70 & 41.56 & 21.44 & 39.67 & 80.83 & 78.74 & 26.25 & 71.96 & 56.38 \\
Qwen2-Audio-Instruct & 52.14 & 32.87 & 35.56 & 39.02 & 77.62 & 64.81 & 46.67 & 68.90 & 53.27 \\
Gemini 2.0 Flash & 47.17 & 41.30 & 30.62 & 40.83 & 70.69 & 70.69 & 36.16 & 47.83 & 51.03 \\
\midrule
\multicolumn{10}{c}{\cellcolor{green!15}\textbf{Open-Source Audio Models}} \\
\midrule
Audio Flamingo 3 & -- & -- & -- & -- & -- & -- & -- & -- & 62.30 \\
MiniCPM & 56.56 & 34.05 & 36.48 & 40.54 & 80.71 & 74.72 & 46.71 & 73.57 & 56.53 \\
MERA LION & 54.49 & 33.69 & 25.84 & 35.74 & 80.32 & 77.18 & 41.49 & 73.68 & 54.10 \\
Qwen-Audio-Chat & 57.21 & 38.52 & 24.70 & 35.69 & 58.61 & 59.78 & 25.60 & 55.93 & 46.92 \\
DIVA & 44.36 & 33.72 & 27.45 & 33.95 & 62.32 & 74.24 & 40.00 & 65.04 & 48.31 \\
Megrez-3B-Omni & 41.36 & 32.52 & 26.35 & 32.48 & 73.53 & 66.11 & 40.42 & 67.05 & 49.03 \\
Step-Audio & 31.56 & 29.39 & 24.01 & 28.72 & 49.10 & 50.09 & 45.27 & 47.27 & 37.42 \\
BLSP & 31.35 & 20.96 & 23.75 & 28.36 & 47.91 & 42.31 & 42.08 & 44.97 & 35.96 \\
GLM-4-Voice & 27.80 & 24.52 & 27.34 & 26.18 & 46.10 & 48.16 & 44.35 & 46.76 & 35.51 \\
\midrule
\multicolumn{10}{c}{\cellcolor{yellow!15}\textbf{Human Performance (Upper Bound)}} \\
\midrule
Human & \colorbox{lightblue}{\textbf{87.10}} & \colorbox{lightblue}{\textbf{94.32}} & \colorbox{lightblue}{\textbf{92.88}} & \colorbox{lightblue}{\textbf{91.24}} & 82.16 & \colorbox{lightblue}{\textbf{87.60}} & \colorbox{lightblue}{\textbf{89.12}} & \colorbox{lightblue}{\textbf{86.77}} & \colorbox{lightblue}{\textbf{89.72}} \\
\midrule
\multicolumn{10}{c}{\cellcolor{black!10}\textbf{Random Baselines}} \\
\midrule
Most Frequent Choice & 26.20 & 26.04 & 27.83 & 29.83 & 28.30 & 28.30 & 30.10 & 28.41 & 28.06 \\
Random Guess & 24.30 & 25.70 & 26.10 & 24.90 & 23.80 & 25.40 & 25.40 & 25.02 & 25.37 \\
\bottomrule
\end{tabular}
}
\end{table*}

The MMSU benchmark, with its unique split between perception and reasoning tasks, provides a granular lens for a multi-faceted analysis of a model's capabilities. Our examination of the results in Tab.~\ref{tab:mmsu_results} reveals several critical findings, starting with the validation of our method's reasoning capabilities, followed by an exploration of its surprising efficiency and broader impacts, and concluding with an identification of key frontiers for future research.

\paragraph{The Initial Breakthrough: Reasoning Closes to the Human Level.}
The analysis first confirms the efficacy of CESAR in its target domain. The data shows the model achieves superior performance in reasoning tasks, where its average score of 81.07\% not only establishes a new state-of-the-art but also approaches the human benchmark of 86.77\%. This proficiency is particularly pronounced in Semantic Reasoning, where CESAR achieves a super-human score of 88.72\%. This is a direct and powerful validation that our process-oriented training is exceptionally effective at cultivating the kind of deep, nuanced understanding that was previously the domain of human cognition.

\paragraph{A Small Model Can also Win.}
A more compelling finding emerges, however, one that challenges the foundational assumptions of the field. The data in Tab.~\ref{tab:mmsu_results} reveals that CESAR, a 7B model, not only establishes superior performance in reasoning over significantly larger proprietary models like Gemini 1.5 Pro, but also—unexpectedly—surpasses them in average perception tasks (48.45\% vs. 46.10\%). This is a pivotal finding: it demonstrates that a smaller model, when endowed with superior reasoning, can outperform larger competitors across \textit{both} cognitive and perceptual dimensions. This result provides strong evidence for a new, more efficient scaling paradigm.

\paragraph{The Ripple Effect: How Better Thinking Creates Better Hearing.}
The model's unexpected strength in perception suggests a deeper mechanism is at play. The data points to an unanticipated synergy: enhancing reasoning seems to have a beneficial ripple effect on foundational perception. While our rewards explicitly target higher-order thinking, the process of learning to form consistent and logical connections appears to refine the model's underlying representations of the acoustic world. This finding suggests that our methodology improves not just how a model interprets sounds, but how effectively it processes them at a more fundamental level, explaining how a targeted cognitive enhancement can lead to broader, more holistic improvements in multimodal understanding.

\paragraph{A New Vista: Beyond the Perceptual Bottleneck.}
Ultimately, the success in reasoning brings a critical challenge for the field into sharp relief. The asymmetry between our model's near-human reasoning and its still-developing perception (which, while outperforming its peers, still lags far behind the human score of 91.24\%) illuminates a clear ``perceptual bottleneck." This should not be viewed as a limitation of our method, but rather as a key finding that clarifies the path for future research. It reinforces our central thesis: the future lies not in the resource-intensive race of simply scaling up model size, but in the new science of scaling specific \textit{capabilities}. Our work demonstrates that reasoning can be scaled to near-human levels within a compact model; the clear next step is to apply a similarly focused, principled approach to perception, paving the way for a future of AI that is not only more powerful but also dramatically more efficient and accessible.

\clearpage

\subsection{Test-Time Scaling Analysis: From Inverse Scaling to Controllable and Scalable Reasoning}
\label{app:test-time scaling_analysis}

\begin{figure}[!ht]
\centering
\includegraphics[width=0.9\linewidth]{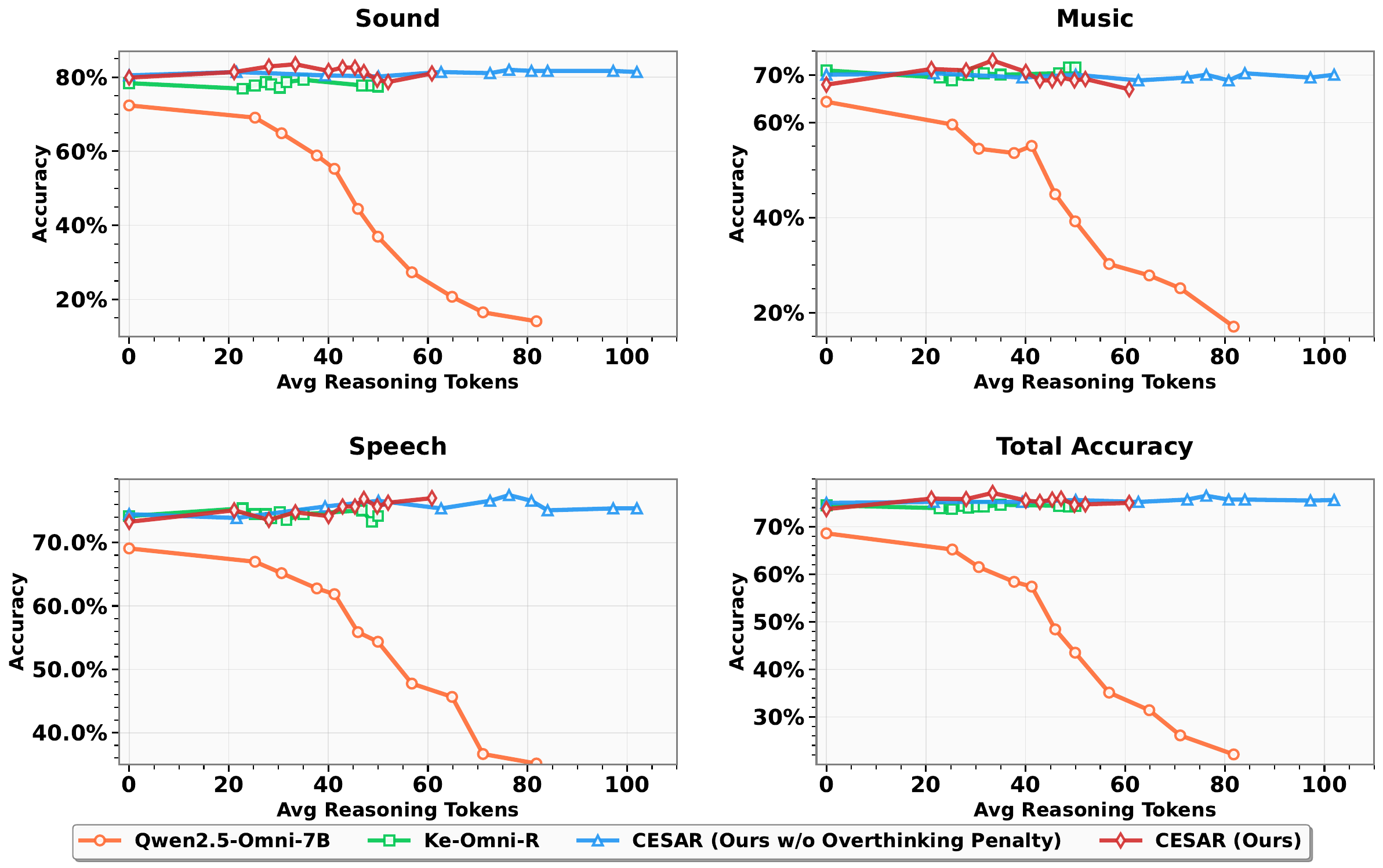}
\includegraphics[width=0.9\linewidth]{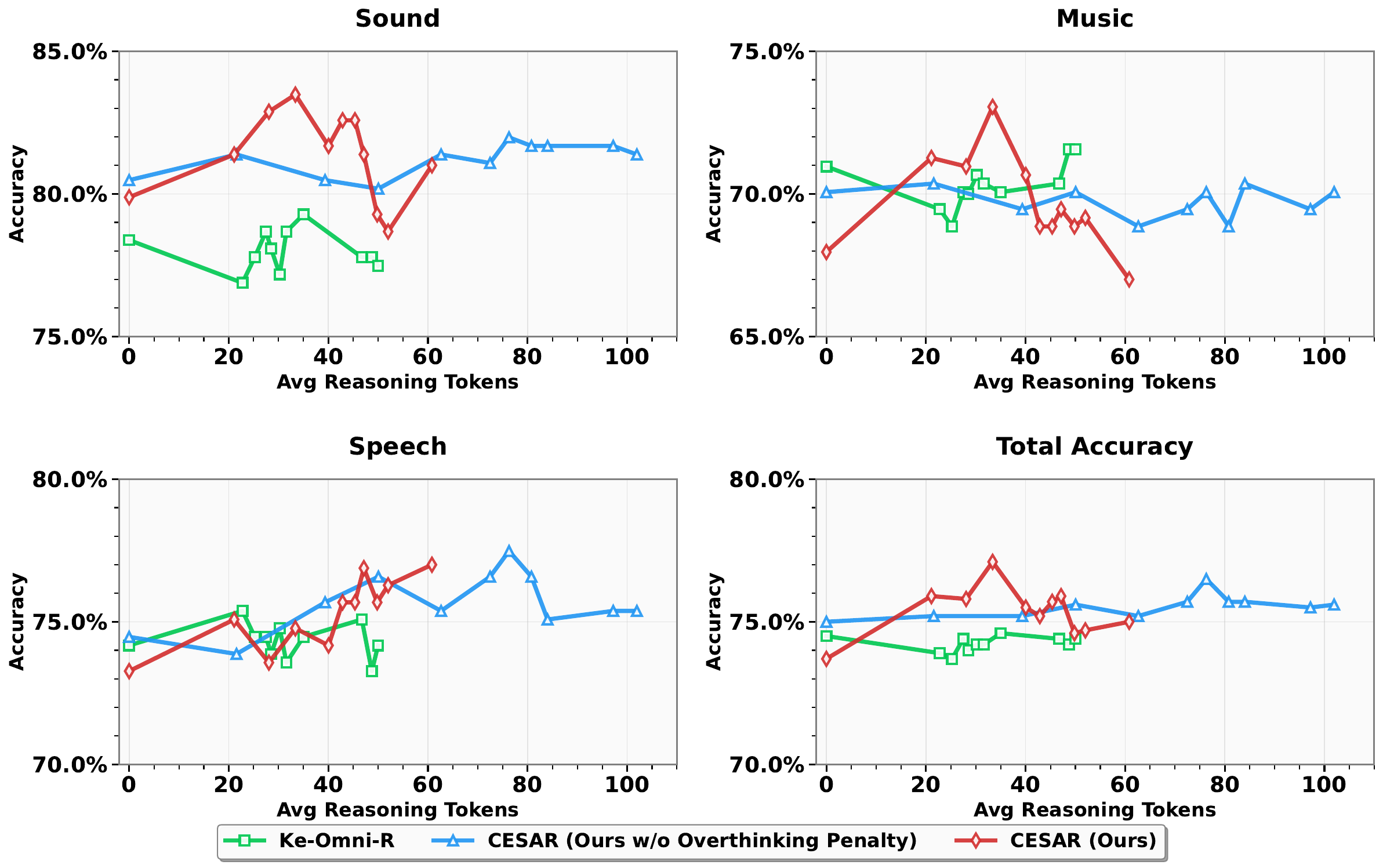}
\caption{
    \textbf{Test-Time Scaling Curves of Reasoning.} Accuracy is plotted against the average length of the reasoning chain (in used tokens). 
    \textbf{(Top Row)} The full comparison reveals a catastrophic performance collapse of the base Qwen2.5-Omni-7B model as it generates longer reasoning chains, empirically demonstrating the test-time inverse scaling problem. In contrast, all RL-trained models remain robust.
    \textbf{(Bottom Row)} A zoomed-in view of the RL models highlights the performance peak of our full method (i.e., CESAR (Ours)), which discovers a ``reasoning sweet spot". It consistently outperforms both the version without the Overthinking Penalty reward (i.e., CESAR (Ours w/o Overthinking Penalty)) and the Ke-Omni-R baseline.
}
\label{fig:scaling_curves full}
\end{figure}

Aggregate accuracy scores, while informative, obscure a critical underlying dynamic: the \textbf{test-time inverse scaling problem} that plagues audio language models lacking explicit reasoning training. This phenomenon is twofold: first, prompting such models to generate a chain of thought often yields worse results than direct, zero-shot answering. Second, their performance degrades precipitously as the reasoning chain lengthens. To rigorously investigate these dynamics and validate our solution, we introduce a \textbf{test-time scaling analysis}.

This training-free inference methodology allows us to probe a model's reasoning capability as a function of its computational budget. We achieve this by systematically varying an upper bound on reasoning length, specifically by adjusting the \texttt{max\_think\_len} parameter within the prompt across a range (e.g., 25, 50, ..., 250) \citep{ke_omni_r}. Critically, this parameter does not force a fixed output length; rather, it provides a ceiling, allowing the model to autonomously determine an appropriate reasoning depth based on the problem's demands and its own intrinsic capabilities. We then plot accuracy against the \textit{actual average number of reasoning tokens} generated. As illustrated in Fig.~\ref{fig:scaling_curves full}, this analysis provides a granular view into each model's reasoning behavior, reveals profound differences in their underlying skills, and demonstrates how to fully unlock the latent reasoning potential cultivated by our framework.

\paragraph{The Peril of Untrained Reasoning: A Case of Test-Time Inverse Scaling.}
The most dramatic finding, shown in the top row of Fig.~\ref{fig:scaling_curves full}, is the \textbf{catastrophic performance collapse} of the base Qwen2.5-Omni-7B model. While it begins with a respectable accuracy at zero reasoning tokens (i.e., direct answering), its performance enters a free fall as it is prompted to generate longer reasoning processes. This provides powerful empirical evidence for the test-time inverse scaling problem: for a model that has not been explicitly trained \textit{how} to reason, ``thinking more" is actively harmful. Each additional token of unguided ``reasoning" introduces new opportunities for error accumulation and hallucination, transforming a decent zero-shot guesser into a demonstrably poor reasoner.

\paragraph{Curing Test-Time Inverse Scaling through Process-Oriented RL.}
In stark contrast, all models trained with our process-oriented reinforcement learning framework are completely immune to this collapse. The bottom row of Fig.~\ref{fig:scaling_curves full} shows that our models (CESAR and its variant) and the Ke-Omni-R baseline all maintain remarkable stability as reasoning length increases. This result establishes a fundamental principle: \textbf{robust training cultivates robust scaling}. By receiving granular feedback on the reasoning \textit{process}, our models learn to generate coherent, logically sound, and self-consistent reasoning processes that do not derail. This learned robustness is the key to curing the test-time inverse scaling fragility observed in the base model, transforming reasoning from detriments into gains.

\paragraph{Unlocking Calibrated Reasoning and Model-Specific ``Sweet Spots".}
The zoomed-in analysis reveals the final and most profound layer of insight. While the Ke-Omni-R baseline is stable, its performance is noisy and fails to consistently benefit from longer reasoning. This comparison highlights the limitations of outcome-only rewards. Our process-oriented rewards, however, cultivate distinct and controllable reasoning styles. The CESAR (w/o Overthinking Penalty) variant, for instance, exhibits a preference for longer reasoning chains, showing sustained high performance as token count increases. 

Most significantly, our full method, CESAR, demonstrates a form of learned metacognition. It discovers a model-specific \textbf{``reasoning sweet spot,"} where its performance actively \textit{peaks} at an optimal reasoning depth (around 30-40 tokens) before gracefully stabilizing. This behavior is a direct consequence of the `Overthinking Penalty` reward, which trains the model to balance analytical sufficiency with conciseness. It learns not only \textit{how} to reason, but also \textit{how much} to reason. This finding challenges the monolithic ``longer is better" assumption, proving that reasoning is a trainable skill. The test-time scaling analysis is therefore not merely an evaluation tool; it is the key to unlocking these cultivated, model-specific reasoning capabilities at inference time, allowing each model to achieve its peak performance.

\clearpage

\subsection{Quantifying Reasoning Quality beyond Accuracy: An AI-as-Judge Framework}
\label{app:llm_judge_analysis}

\begin{figure}[!ht]
\centering
\includegraphics[width=0.8\linewidth]{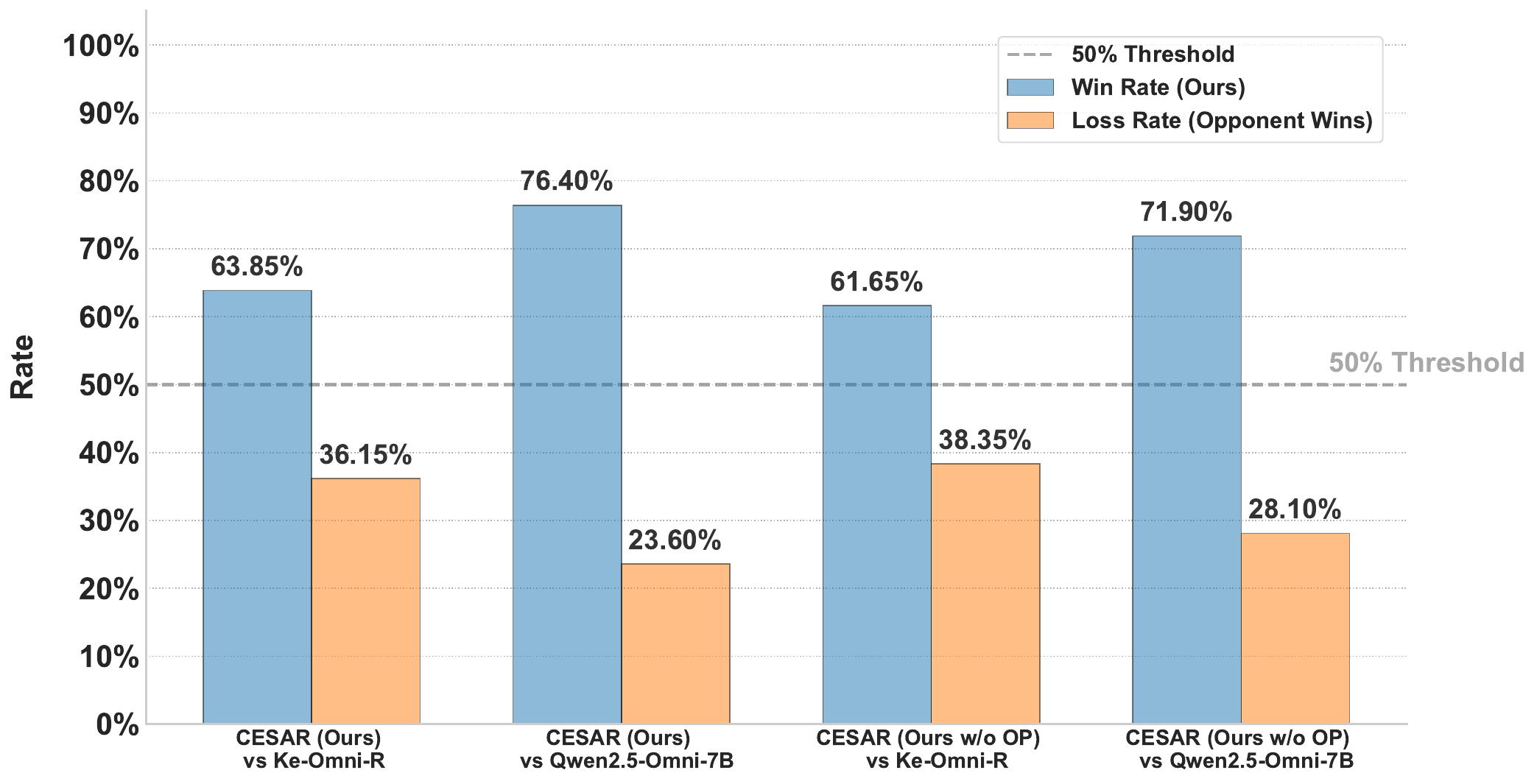}
\caption{
    \textbf{Quantitative Analysis of Reasoning Quality via AI-as-Judge.} Win rate of our model's reasoning process when compared head-to-head against the base model (Qwen2.5-Omni-7B) and the Ke-Omni-R baseline. The dominant win rates provide strong quantitative evidence of superior reasoning quality, a metric that accuracy scores alone cannot capture. Throughout, OP denotes the overthinking penalty defined in \eqref{equ: overthink penalty}.
}
\label{fig: win_rate}
\end{figure}

Evaluating the quality of a model's reasoning process—distinct from the correctness of its final answer—poses a significant challenge in multimodal AI. Current evaluation paradigms are insufficient: outcome-based accuracy is a coarse proxy that can reward correct answers derived from flawed or fallacious logic, while anecdotal qualitative analysis is inherently unscalable and subjective. This methodological gap makes it difficult to determine whether a model is developing genuine analytical skill or simply becoming a more effective test-taker.

To address this, we introduce a scalable and rigorous framework for the quantitative evaluation of reasoning quality itself. We employ a powerful, state-of-the-art multimodal model, akin to GPT-4o Audio \citep{gpt4oaudio}, as an expert adjudicator. For each comparison, after providing the judge with the full context (audio, question, choices, and correct answer) and the two reasoning traces, we use the following direct prompt for evaluation: \textit{Given the audio context and two reasoning processes from Model A and Model B, try to determine which process is superior. A superior process is more logical, faithful to the audio, and follows a clearer analytical path. Focus on the quality of the reasoning, not just the final answer's correctness, and conclude with 'Model A Wins', 'Model B Wins', or 'Tie'.} This approach ensures a consistent and targeted assessment focused squarely on the analytical process. 

Noting that to enhance the clarity and readability of the final results of AI-as-Judge, we distribute 'Tie' outcomes equally for the final win-rate calculation (e.g., an initial outcome where Model A wins 40\%, Model B wins 20\%, and 40\% are ties is converted to a final win rate of 60\% for A and 40\% for B).

The results of AI-as-Judge, presented in Fig.~\ref{fig: win_rate}, are decisive. Against the base Qwen2.5-Omni-7B model, CESAR's reasoning is judged superior in a commanding \textbf{76.4\%} of comparisons. This offers the first quantitative proof that our process-centric rewards cultivate a fundamentally more robust and logical reasoning architecture, rather than merely improving final-answer accuracy. Even more critically, when pitted against Ke-Omni-R—a strong baseline also trained with reinforcement learning but with outcome-only rewards—our model's reasoning prevails in a significant \textbf{63.85\%} of cases. This result starkly illustrates the limitations of simplistic, outcome-based RL and validates the necessity of our multi-faceted reward suite for shaping genuinely high-quality reasoning. By filling a critical methodological gap, our work provides the field with a scalable tool to assess reasoning quality directly, proving that our process-centric framework cultivates verifiably superior analytical capabilities.

\clearpage

\subsection{Qualitative Analysis: The Anatomy of Cultivated Reasoning}
\label{app:Qualitative_Analysis}

Beyond quantitative benchmarks, a granular examination of the reasoning traces reveals the concrete mechanisms through which CESAR transforms a model's cognitive behavior. The following head-to-head comparisons are not merely anecdotal evidence of better performance; they are a microscopic validation of our central thesis. They illustrate precisely how our process-oriented rewards eradicate the critical failure modes—hallucination, logical fallacies, and inconsistency—that plague models trained without such guidance, thereby cultivating reasoning that is demonstrably more \textbf{E}ffective, \textbf{C}onsistent, and \textbf{L}ogical.

\subsubsection{Comparison with Base Model (Qwen2.5-Omni-7B): Untrained vs. Cultivated Reasoning}
This comparison starkly contrasts the brittle, uncontrolled output of a powerful base model against the robust, structured thought process instilled by CESAR. The examples demonstrate that merely prompting a model to ``think" is insufficient; without explicit training on \textit{how} to reason, the process itself becomes a source of error, leading to failures that range from high-level factual invention to low-level execution mistakes.

\begin{table*}[!ht]
\centering
\begin{tabular}{p{0.48\linewidth} p{0.48\linewidth}}
    \multicolumn{2}{c}{\textbf{Example 1: Robustness against Hallucination}} \\
    \toprule
    \multicolumn{2}{c}{
        \parbox{\linewidth}{
            \raggedright
            \textbf{Question:} Based on the given audio, identify the source of the speech. \vspace{1mm}\newline
            \textbf{Choices:} man, woman, child, robot \vspace{1mm}\newline
            \textbf{Correct Answer:} man \vspace{1mm}\newline
            \textbf{Task ID:} 72fb5481-73ae-409d-8e16-c94ac48d2ee4
        }
    } \\ 
    \midrule 
    \textcolor{ourscolor}{\textbf{CESAR (Correct)}} & \textcolor{baselinecolor}{\textbf{Qwen2.5-Omni-7B (Wrong)}} \\
    \midrule
    \textbf{Reasoning:} \textit{\rewardkeyword{Given the options}, the speech is likely from a man. The tone and volume \rewardkeyword{suggest} an adult male. The other options are less probable.} 
    &
    \textbf{Reasoning:} \textit{The audio is a speech, and the voice is male. \textcolor{errorcolor}{The options provided do not include 'man', so the closest match is 'robot'}.} \\
    \addlinespace[2pt]
    \textbf{Answer:} man 
    & 
    \textbf{Answer:} \textcolor{errorcolor}{robot} \\
    \bottomrule
\end{tabular}
\end{table*}

\paragraph{Grounded Reasoning vs. Factual Hallucination.} The base model often fails to ground its reasoning in the provided context. In Example 1, after correctly identifying the acoustic evidence (``the voice is male"), it invents a false premise—that 'man' is absent from the choices—and proceeds to an illogical conclusion. This reveals a fundamental inability to tether its internal state to the problem's explicit constraints. CESAR, by contrast, exhibits a learned, systematic methodology. Its trace begins with a grounding step (\rewardkeyword{Given the options}), explicitly acknowledging the context, and then connects acoustic cues to an inference using relational language (\rewardkeyword{suggests})—showcasing an \textbf{effective} and sound analytical process directly instilled by our `Keywords` reward.

\begin{table*}[!ht]
\centering
\begin{tabular}{p{0.48\linewidth} p{0.48\linewidth}}
    \multicolumn{2}{c}{\textbf{Example 2: Logical Deduction from Acoustic Cues}} \\
    \toprule
    \multicolumn{2}{c}{
        \parbox{\linewidth}{
            \raggedright
            \textbf{Question:} What kind of event could be taking place in the audio? \vspace{1mm}\newline
            \textbf{Choices:} A quiet library during study hours, A heated argument escalating to violence, A military parade with a marching band, A gentle breeze blowing through trees \vspace{1mm}\newline
            \textbf{Correct Answer:} A military parade with a marching band \vspace{1mm}\newline
            \textbf{Task ID:} 3690d238-8902-4217-a068-80a56972842e
        }
    } \\ 
    \midrule 
    \textcolor{ourscolor}{\textbf{CESAR (Correct)}} & \textcolor{baselinecolor}{\textbf{Qwen2.5-Omni-7B (Wrong)}} \\
    \midrule
    \textbf{Reasoning:} \textit{The presence of \rewardkeyword{marching, music, and shouting suggests} a large-scale event. The options are a quiet library, a heated argument, a military parade, and a gentle breeze. A military parade \rewardkeyword{fits the description best}, as it involves a marching band and organized \rewardkeyword{shouting}.}
    &
    \textbf{Reasoning:} \textit{The audio contains the sound of a man speaking, followed by a loud, \textcolor{errorcolor}{chaotic noise}. The options are a quiet library, a heated argument, a military parade, and a gentle breeze. The \textcolor{errorcolor}{chaotic noise could be interpreted as a violent escalation}, so a heated argument is the most likely scenario.} \\
    \addlinespace[2pt]
    \textbf{Answer:} A military parade with a marching band
    & 
    \textbf{Answer:} \textcolor{errorcolor}{A heated argument escalating to violence} \\
    \bottomrule
\end{tabular}
\end{table*}

\paragraph{Logical Deduction vs. Flawed Inference.} The base model struggles to interpret complex acoustic scenes. In Example 2, it oversimplifies a rich soundscape into a generic, low-information label (\textcolor{errorcolor}{``chaotic noise"}), from which it makes a spurious logical leap. This highlights a core weakness: the inability to decompose and analyze multi-element audio. CESAR demonstrates a far more sophisticated, multi-step \textbf{logical} process. It correctly identifies the constituent acoustic parts (\rewardkeyword{marching, music, and shouting}), synthesizes them into a high-level hypothesis (``a large-scale event"), and systematically evaluates the choices to find the one that \rewardkeyword{fits the description best}. This is a true deductive capability, a direct outcome of rewarding structured analytical patterns.

\begin{table*}[!ht]
\centering
\begin{tabular}{p{0.48\linewidth} p{0.48\linewidth}}
    \multicolumn{2}{c}{\textbf{Example 3: Reasoning-Answer Consistency}} \\
    \toprule
    \multicolumn{2}{c}{
        \parbox{\linewidth}{
            \raggedright
            \textbf{Question:} What was the order of the sounds in the sequence? \vspace{1mm} \newline
            \textbf{Choices:} ['light\_switch\_clicking', 'boiling\_water', 'doorbell\_ringing', 'clock\_ticking'], ['boiling\_water', 'light\_switch\_clicking', 'clock\_ticking', 'doorbell\_ringing'], ['clock\_ticking', 'doorbell\_ringing', 'boiling\_water', 'light\_switch\_clicking'], ['doorbell\_ringing', 'clock\_ticking', 'light\_switch\_clicking', 'boiling\_water'] \vspace{1mm} \newline
            \textbf{Correct Answer:} ['light\_switch\_clicking', 'boiling\_water', 'doorbell\_ringing', 'clock\_ticking'] \vspace{1mm}\newline
            \textbf{Task ID:} bbabe360-0573-43d4-b2e6-6892150cbdcd
        }
    } \\ 
    \midrule 
    \textcolor{ourscolor}{\textbf{CESAR (Correct)}} & \textcolor{baselinecolor}{\textbf{Qwen2.5-Omni-7B (Wrong)}} \\
    \midrule
    \textbf{Reasoning:} \textit{\rewardkeyword{Considering the sequence}, the light switch clicking is \rewardkeyword{followed by} boiling water, \rewardkeyword{then} a doorbell rings, and \rewardkeyword{finally}, a clock ticks. This order fits the given options.}
    &
    \textbf{Reasoning:} \textit{First, there's a light switch click, then boiling water sound, followed by a doorbell ringing, and finally, a clock ticking.} \\
    \addlinespace[2pt]
    \textbf{Answer:} ['light\_switch\_clicking', 'boiling\_water', 'doorbell\_ringing', 'clock\_ticking']
    &
    \textbf{Answer:} ['light\_switch\textcolor{errorcolor}{\_click\_click}', 'boiling\_water', 'doorbell\_ringing', 'clock\_ticking'] \\
    \bottomrule
\end{tabular}
\end{table*}

\paragraph{Precise Execution vs. Minor Inconsistency.} Example 3 reveals a subtle but critical failure in precision. While the base model's reasoning is perceptually and logically correct, it stumbles at the final execution step, introducing a fatal typo (\textcolor{errorcolor}{\_click\_click}) into its answer. This shows that a sound thought process is nullified if the model cannot reliably translate it into a correct final output. CESAR's end-to-end process rewards, which value not just the reasoning trace but its faithful conversion into a final answer, ensure this level of precision. Its reasoning trace also exhibits the sequential pattern keywords (\rewardkeyword{Considering the sequence... followed by... then... finally}) that our framework explicitly incentivizes.

\subsubsection{Comparison with RL Baseline (Ke-Omni-R): The Necessity of Process Rewards}
This comparison isolates the crucial contribution of our multi-faceted, process-oriented reward suite against a baseline also trained with RL but guided only by the correctness of the final answer. These examples provide causal evidence that optimizing for outcomes alone is insufficient, as it permits and may even reinforce broken cognitive processes.

\begin{table*}[!ht]
\centering
\begin{tabular}{p{0.48\linewidth} p{0.48\linewidth}}
    \multicolumn{2}{c}{\textbf{Example 4: Superior Reasoning-Answer Consistency}} \\
    \toprule
    \multicolumn{2}{c}{
        \parbox{\linewidth}{
            \raggedright
            \textbf{Question:} How many times does the telephone ring in the audio? \vspace{1mm}\newline
            \textbf{Choices:} 2, 4, 5, 3 \vspace{1mm}\newline
            \textbf{Correct Answer:} 3 \vspace{1mm}\newline
            \textbf{Task ID:} b132f501-53cd-4e78-84e3-ac65c5588260
        }
    } \\ 
    \midrule 
    \textcolor{ourscolor}{\textbf{CESAR (Correct)}} & \textcolor{baselinecolor}{\textbf{Ke-Omni-R (Wrong)}} \\
    \midrule
    \textbf{Reasoning:} \textit{\rewardkeyword{Considering the options}, the telephone \rewardkeyword{rings three times}. The \rewardkeyword{audio} clearly shows three distinct \rewardkeyword{rings}.} 
    &
    \textbf{Reasoning:} \textit{The telephone rings \textcolor{errorcolor}{three} times, as indicated by the \textcolor{errorcolor}{three} distinct ring sounds in the audio.} \\
    \addlinespace[2pt]
    \textbf{Answer:} 3 
    & 
    \textbf{Answer:} \textcolor{errorcolor}{2} \\
    \bottomrule
\end{tabular}
\end{table*}

\paragraph{Enforcing Reasoning-Answer Consistency.} Example 4 is a textbook failure of reasoning-answer consistency. Ke-Omni-R correctly identifies the acoustic fact in its reasoning (``rings three times") but then inexplicably provides a contradictory answer (\textcolor{errorcolor}{2}). An outcome-only reward system is blind to this internal contradiction; on trials where the model might guess the correct final answer despite such a flaw, it would be rewarded, reinforcing this broken cognitive process. CESAR's `Reasoning Consistency Reward` directly targets this failure mode, ensuring the final answer is a direct and faithful consequence of the thought process. This makes the reasoning verifiably \textbf{consistent} and trustworthy.

\begin{table*}[!ht]
\centering
\begin{tabular}{p{0.48\linewidth} p{0.48\linewidth}}
    \multicolumn{2}{c}{\textbf{Example 5: Concise vs. Convoluted Reasoning}} \\
    \toprule
    \multicolumn{2}{c}{
        \parbox{\linewidth}{
            \raggedright
            \textbf{Question:} What is the characteristic of the chord played from 0:02 to 0:03 in the audio? \vspace{1mm}\newline
            \textbf{Choices:} A:min/P5, E:maj/P1, D:maj/M3, G:maj/P1 \vspace{1mm}\newline
            \textbf{Correct Answer:} A:min/P5 \vspace{1mm}\newline
            \textbf{Task ID:} 7333dffb-368f-44cc-adb4-b4e9805164a3
        }
    } \\ 
    \midrule 
    \textcolor{ourscolor}{\textbf{CESAR (Correct)}} & \textcolor{baselinecolor}{\textbf{Ke-Omni-R (Wrong)}} \\
    \midrule
    \textbf{Reasoning:} \textit{\rewardkeyword{Considering the options}, 'A:min/P5' seems the \rewardkeyword{most fitting} as it's a \rewardkeyword{minor chord} and 'P5' indicates a \rewardkeyword{perfect fifth}. The others don't fit the context as well.}
    &
    \textbf{Reasoning:} \textit{The chord progression from 0:02 to 0:03 involves a progression from a lower note to a higher one... \textcolor{errorcolor}{The progression is from a lower note to a lower one... The progression is from a lower note to a lower one... The progression is from a lower note to a higher one...}} \\
    \addlinespace[2pt]
    \textbf{Answer:} A:min/P5
    &
    \textbf{Answer:} \textcolor{errorcolor}{E:maj/P1} \\
    \bottomrule
\end{tabular}
\end{table*}

\paragraph{Efficient Reasoning vs. Convoluted Overthinking.} Example 5 highlights the problem of unconstrained, inefficient reasoning. The Ke-Omni-R baseline engages in a convoluted and self-contradictory monologue (\textcolor{errorcolor}{a loop of opposing statements}). This ``overthinking" is a direct symptom of a reward function that does not value efficiency. CESAR, shaped by our Overthinking Penalty and Keywords rewards, produces an \textbf{effective} and parsimonious trace. It efficiently uses domain knowledge (\rewardkeyword{minor chord}, \rewardkeyword{perfect fifth}) and structured evaluation (\rewardkeyword{most fitting}) to arrive at the answer, demonstrating a learned ability to allocate cognitive effort purposefully.

\begin{table*}[!ht]
\centering
\begin{tabular}{p{0.48\linewidth} p{0.48\linewidth}}
    \multicolumn{2}{c}{\textbf{Example 6: Robustness against Factual Hallucination}} \\
    \toprule
    \multicolumn{2}{c}{
        \parbox{\linewidth}{
            \raggedright
            \textbf{Question:} Given that an infant is crying, name a famous person known for their association with the animal heard in the background. \vspace{1mm}\newline
            \textbf{Choices:} George Orwell, Ernest Hemingway, Charles Darwin, Charles Dickens \vspace{1mm}\newline
            \textbf{Correct Answer:} Ernest Hemingway \vspace{1mm}\newline
            \textbf{Task ID:} 255742c0-cc69-4be5-891e-a3c796904934
        }
    } \\ 
    \midrule 
    \textcolor{ourscolor}{\textbf{CESAR (Correct)}} & \textcolor{baselinecolor}{\textbf{Ke-Omni-R (Wrong)}} \\
    \midrule
    \textbf{Reasoning:} \textit{\rewardkeyword{Among the options}, Ernest Hemingway is known for his love of animals, particularly cats. The animal sound could be a cat. George Orwell and Charles Dickens are not known for their association with cats. Charles Darwin is known for his work on evolution, not for cats.} 
    &
    \textbf{Reasoning:} \textit{Considering the context of the question, the animal in the background is a pig. \textcolor{errorcolor}{Ernest Hemmanway is famously associated with pigs in his works, particularly in ``The Old Man and the Sea."}} \\
    \addlinespace[2pt]
    \textbf{Answer:} Ernest Hemingway
    & 
    \textbf{Answer:} \textcolor{errorcolor}{Ernest Hemmanway} \\
    \bottomrule
\end{tabular}
\end{table*}

\paragraph{Grounded Knowledge vs. Factual Hallucination.} Finally, Example 6 reveals a spectacular, multi-layered failure of factual grounding from the RL baseline. Ke-Omni-R not only misidentifies the animal sound as a ``pig," but proceeds to invent a completely false piece of literary trivia to justify its choice, incorrectly linking Hemingway to pigs in ``The Old Man and the Sea." This shows how an outcome-only agent can generate syntactically plausible reasoning that is semantically untethered from reality. The failure is compounded by an execution error, as it ultimately misspells the author's name (\textcolor{errorcolor}{Hemmanway}) in the final answer. CESAR's reasoning, by contrast, is sound. It correctly identifies a plausible animal (cat), recalls Hemingway's true association with it, and systematically eliminates other options. Our holistic reward suite prevents this kind of unconstrained hallucination, ensuring the reasoning process is both internally and externally valid.

\subsubsection{Efficient Reasoning vs. Pathological Overthinking: The Critical Role of the Overthinking Penalty}
This final comparison isolates the specific contribution of our `Overthinking Penalty` by presenting a case where its absence leads to outright failure. Here, the full CESAR model arrives at the correct answer through a clear and evidence-based analytical path. The `CESAR w/o OP` variant, however, initially identifies the correct acoustic evidence but then engages in a long, convoluted reasoning process. This overthinking causes it to fixate on an irrelevant, imagined context, ultimately leading it to discard the correct answer in favor of a flawed one. This example provides dramatic proof that overthinking is not a benign inefficiency but a pathological failure mode that can derail the entire analytical process. It demonstrates that the `Overthinking Penalty` is not merely a regularizer for efficiency, but a critical component for ensuring robustness and correctness.

\begin{table*}[!ht]
\centering
\begin{tabular}{p{0.48\linewidth} p{0.48\linewidth}}
    \multicolumn{2}{c}{\textbf{Example 7: Grounded Acoustic Analysis vs. Contextual Hallucination}} \\
    \toprule
    \multicolumn{2}{c}{
        \begin{varwidth}{\linewidth}
            \raggedright
            \textbf{Question:} What is the sound event taking place in the audio? \vspace{1mm}\newline
            \textbf{Choices:} A person crying, A person laughing, A person coughing, A person yawning \vspace{1mm}\newline
            \textbf{Correct Answer:} A person laughing \vspace{1mm}\newline
            \textbf{Task ID:} 01a7e75e-302a-464a-959c-6b3a04213d2f
        \end{varwidth}
    } \\ 
    \midrule 
    \textcolor{ourscolor}{\textbf{CESAR (Correct)}} & \textcolor{baselinecolor}{\textbf{CESAR w/o OP (Wrong)}} \\
    \midrule
    \textbf{Reasoning:} \textit{The \rewardkeyword{audio} contains a series of sharp, rhythmic vocal bursts. The \rewardkeyword{pitch} and cadence are \rewardkeyword{consistent with} joyful laughter. It does not match the sound patterns of crying (more sustained wailing), coughing (a forceful expulsion of air), or yawning (a long, deep inhalation and exhalation). \rewardkeyword{Therefore}, the sound is a person laughing.}
    &
    \textbf{Reasoning:} \textit{The audio clearly contains a human vocalization. It's a series of short, sharp, and somewhat rhythmic sounds. This could be interpreted as either laughing or crying, as both can have a similar staccato pattern. Let's analyze the context. While it sounds like laughter, \textcolor{errorcolor}{this type of abrupt, almost gasping sound can also be associated with someone trying to suppress tears or crying in a very distressed way. Crying is a more primal and urgent emotional expression than laughter. Given the ambiguity, the more intense and emotionally significant possibility is often a safer bet in interpretation, as it might signify a situation requiring attention. The sound has a slightly strained quality that leans more towards distress than pure joy.}} \\
    \addlinespace[2pt]
    \textbf{Answer:} A person laughing
    & 
    \textbf{Answer:} \textcolor{errorcolor}{A person crying} \\
    \bottomrule
\end{tabular}
\end{table*}

\paragraph{Evidence-Based Deduction vs. Over-analysis of Ambiguity.}
The Example 7 provides a textbook case of a model ``thinking itself into a mistake." Both models correctly identify the core acoustic features (``sharp, rhythmic vocal bursts"). The full CESAR model follows a direct and logical path: it compares these features against the prototypes of each option and correctly concludes they are ``consistent with joyful laughter." Its reasoning is grounded entirely in the acoustic evidence.

The `CESAR w/o OP` model, however, falls into a trap of pathological overthinking. After acknowledging the sound could be laughter, it begins to construct an elaborate, unfalsifiable alternative narrative. It introduces irrelevant external concepts like ``\textit{\textcolor{errorcolor}{primal and urgent emotional expression}}" and ``\textit{\textcolor{errorcolor}{safer bet in interpretation}}." It imagines a ``strained quality" to support a ``distress" hypothesis that is not strongly grounded in the audio. By lacking a penalty for this verbose and speculative detour, the model gives undue weight to a complex, imagined scenario over the most direct interpretation of the sound itself. This demonstrates perfectly how the `Overthinking Penalty` is crucial for keeping the model's reasoning tethered to evidence, preventing it from spiraling into contextual hallucinations that corrupt an initially correct perception.

\clearpage

\subsection{Ablation Study: Deconstructing the Sources of Improved Reasoning Capability in CESAR}
\label{app:Ablation_Results}

To precisely quantify the contribution of each component within the CESAR framework, we conduct a progressive ablation study, systematically deconstructing our full model to isolate the impact of each design choice. The results, presented in Tab.~\ref{tab:ablation}, provide a comprehensive validation of our methodology, revealing not only that each component is effective, but also how they synergize to cultivate robust reasoning.

\begin{table*}[!ht]
\centering
\caption{Progressive Ablation Study Results. We systematically remove components from our full method to demonstrate their individual contributions. ``Reasoning" refers to the chain-of-thought reasoning mechanism. Best scores are highlighted in \colorbox{lightblue}{\textbf{blue}}, second-best scores in \colorbox{lightgreen}{\textbf{green}}.}
\label{tab:ablation}
\resizebox{\textwidth}{!}{
\begin{tabular}{l|c|c|c|cccc|cccc}
\toprule
\multirow{2}{*}{\textbf{Method}} & \multirow{2}{*}{\textbf{Ablation Components}} & \multirow{2}{*}{\textbf{Reasoning?}} & \multirow{2}{*}{\textbf{RL Post-training}} & \multicolumn{4}{c|}{\textbf{Technical Components}} & \multicolumn{4}{c}{\textbf{Performance (\%)}} \\
\cmidrule(lr){5-8} \cmidrule(lr){9-12}
& & & & \textbf{Consistency} & \textbf{Key Words} & \textbf{Data Augmentation} & \textbf{Overthinking Penalty} & \textbf{Sound} & \textbf{Music} & \textbf{Speech} & \textbf{Total Accuracy} \\
\midrule
\multicolumn{12}{c}{\cellcolor{gray!20}\textbf{Our Proposed Methods}} \\
\midrule
\multirow{8}{*}{\textbf{CESAR}} 
& \multirow{2}{*}{None (Full Method)} & \xmark & \cmark & \cmark & \cmark & \cmark & \cmark & 79.88 & 67.96 & 73.27 & 73.70 \\
& & \cmark & \cmark & \cmark & \cmark & \cmark & \cmark & \colorbox{lightblue}{\textbf{83.48}} & \colorbox{lightblue}{\textbf{73.05}} & 74.77 & \colorbox{lightblue}{\textbf{77.10}} \\
\cmidrule(lr){2-12}
& \multirow{2}{*}{Overthinking Penalty} & \xmark & \cmark & \cmark & \cmark & \cmark & \xmark & 80.48 & 70.06 & 74.47 & 75.00 \\
& & \cmark & \cmark & \cmark & \cmark & \cmark & \xmark & 81.98 & 70.06 & \colorbox{lightblue}{\textbf{77.48}} & \colorbox{lightgreen}{\textbf{76.50}} \\
\cmidrule(lr){2-12}
& \multirow{2}{*}{Data Augmentation} & \xmark & \cmark & \cmark & \cmark & \xmark & \xmark & 80.18 & 68.86 & 75.38 & 74.80 \\
& & \cmark & \cmark & \cmark & \cmark & \xmark & \xmark & \colorbox{lightgreen}{\textbf{82.28}} & 68.86 & \colorbox{lightblue}{\textbf{77.48}} & 76.20 \\
\cmidrule(lr){2-12}
& \multirow{2}{*}{Key Words} & \xmark & \cmark & \cmark & \xmark & \xmark & \xmark & 80.48 & 66.77 & 74.77 & 74.00 \\
& & \cmark & \cmark & \cmark & \xmark & \xmark & \xmark & 80.78 & 68.86 & \colorbox{lightgreen}{\textbf{75.98}} & 75.20 \\
\midrule
\multicolumn{12}{c}{\cellcolor{gray!20}\textbf{Baseline Methods}} \\
\midrule
\multirow{2}{*}{\textbf{Ke-Omni-R}} & \multirow{2}{*}{Consistency}
& \xmark & \cmark & \xmark & \xmark & \xmark & \xmark & 78.38 & \colorbox{lightgreen}{\textbf{70.96}} & 74.17 & 74.50 \\
& & \cmark & \cmark & \xmark & \xmark & \xmark & \xmark & 79.28 & 70.06 & 74.47 & 74.60 \\
\midrule
\multirow{2}{*}{\textbf{Qwen2.5-Omni-7B}} & \multirow{2}{*}{RL Post-training}
& \xmark & \xmark & \xmark & \xmark & \xmark & \xmark & 72.37 & 64.37 & 69.07 & 68.60 \\
& & \cmark & \xmark & \xmark & \xmark & \xmark & \xmark & 69.07 & 59.58 & 66.97 & 65.20 \\
\bottomrule
\end{tabular}
}
\end{table*}

\paragraph{The Foundational Role of Process-Oriented RL.} The ablation starkly confirms that online reinforcement learning \citep{iclr_rwr,pzero,prance,atari_review,sp_vla} is the core mechanism transforming the model's fundamental capabilities. Removing RL post-training entirely—reverting to the base Qwen2.5-Omni-7B model—causes the most significant performance drop, a staggering 9.4 points in reasoning accuracy (from 74.60\% of Ke-Omni-R to 65.20\%). More critically, this is not merely a quantitative drop but a qualitative reversal: the base model is the only variant that exhibits test-time inverse scaling, where enabling reasoning \textbf{degrades} performance. In contrast, every single RL-trained variant sees a performance \textbf{gain} from reasoning. This demonstrates that RL is not an incremental improvement but the essential catalyst that turns reasoning from detriments into gains, a prerequisite for any further refinement.

\paragraph{The Synergy of Process-Specific Rewards.} Our results clearly show that high-quality reasoning emerges from a synergy of process-oriented rewards, with each component providing a distinct and vital contribution. The most significant gains over the outcome-only RL baseline (Ke-Omni-R) come from our two core process rewards.
First, removing the \textit{Consistency} reward (which effectively reduces our model to the Ke-Omni-R baseline's level of process supervision) leads to a significant performance drop. As confirmed in our qualitative analysis (See App. \ref{app:Qualitative_Analysis}), a model lacking this reward can produce reasoning traces that are completely disconnected from the final answer. If reasoning is not required to be consistent with the answer, it becomes an unreliable, and potentially harmful, cognitive artifact. 
Second, removing the \textit{Key Words} reward causes a further substantial drop of 1.0\% in accuracy (from 76.20\% to 75.20\%). This demonstrates that explicitly rewarding logical structure and domain-specific terminology is a powerful driver of effective analytical strategies. Together, these rewards target distinct but complementary facets of high-quality reasoning—internal coherence and logical structure—proving that a multi-faceted approach is significantly more effective than optimizing for any single aspect alone.

\paragraph{The Importance of Calibrated Training.} The final components, while having a smaller numerical impact, are crucial for calibrating and robustifying the learned skills. Although removing \textit{Data Augmentation} only results in a minor accuracy drop, its role in exposing the model to linguistic diversity is essential for generalization and preventing the learning of superficial correlations in real-world scenarios.
Most interestingly, the \textit{Overthinking Penalty} serves a dual purpose. While its removal leads to a 0.6\% performance drop, it also reveals a deeper insight into the value of reasoning. In our full method, the performance gap between reasoning and non-reasoning modes is 3.4 points (77.10\% vs 73.70\%). Without the penalty, this gap narrows to just 1.5 points (76.50\% vs 75.00\%). This shows that by penalizing inefficient thought, the model learns to better distinguish when a simple, intuitive answer is insufficient and a more deliberate reasoning process is required. This enlarges the space where the cultivated reasoning skill provides a distinct advantage, underscoring the importance of not only knowing how to reason, but also when.

Ultimately, the ablation validates our holistic approach. Our framework comprehensively elevates performance in both non-reasoning and reasoning settings, with each component proving its value in building a model that is not only more accurate but reasons in a more effective, consistent, and logical manner.

\clearpage

\subsection{Benchmark Results on MMAR}
\label{app: Benchmark Results on MMAR}

The MMAR benchmark, designed to test deep, multi-step reasoning on longer audio, exposes the limits of current models. Our analysis of the results in Tab.~\ref{tab:all_mmar_results} reveals a clear fragmentation in reasoning capabilities across the field and highlights distinct challenges for future research.

\begin{table*}[!htbp]
\centering
\caption{MMAR Benchmark Results. We evaluate our method against state-of-the-art audio models across single and mixed audio modalities. Best scores are highlighted in \colorbox{lightblue}{\textbf{blue}}, second-best scores in \colorbox{lightgreen}{\textbf{green}}. All results show accuracy (\%). Mix-S-M: Sound+Music, Mix-S-Sp: Sound+Speech, Mix-M-Sp: Music+Speech, Mix-S-M-Sp: Sound+Music+Speech. We report the performance of Ke-Omni-R \citep{ke_omni_r} from our own reproductions under the same protocol; all other baseline results are taken from the MMAR paper \citep{ghosh2025mmar} (including Qwen2.5-Omni-7B \citep{xu2025qwen25omni}).}
\label{tab:all_mmar_results}
\resizebox{\textwidth}{!}{
\begin{tabular}{l|c|ccc|cccc|c}
\toprule
\textbf{Method} & \textbf{Size} & \textbf{Sound} & \textbf{Music} & \textbf{Speech} & \textbf{Mix-S-M} & \textbf{Mix-S-Sp} & \textbf{Mix-M-Sp} & \textbf{Mix-S-M-Sp} & \textbf{Overall} \\
\midrule
\multicolumn{10}{c}{\cellcolor{blue!15}\textbf{Our Proposed Method}} \\
\midrule
\textbf{CESAR} & \textbf{7B} & \colorbox{lightblue}{\textbf{66.06}} & \colorbox{lightblue}{\textbf{55.83}} & 62.24 & \colorbox{lightgreen}{\textbf{63.64}} & 67.43 & 60.98 & 66.67 & 62.70 \\
\midrule
\multicolumn{10}{c}{\cellcolor{gray!15}\textbf{Base Model}} \\
\midrule
Qwen2.5-Omni-7B & 7B & 58.79 & 40.78 & 59.86 & 54.55 & 61.93 & \colorbox{lightblue}{\textbf{67.07}} & 58.33 & 56.70 \\
\midrule
\multicolumn{10}{c}{\cellcolor{gray!15}\textbf{Audio RL Baseline}} \\
\midrule
Ke-Omni-R & 7B & \colorbox{lightgreen}{\textbf{63.64}} & 47.09 & 62.93 & \colorbox{lightgreen}{\textbf{63.64}} & \colorbox{lightgreen}{\textbf{68.35}} & \colorbox{lightblue}{\textbf{67.07}} & 45.83 & 60.90 \\
\midrule
\multicolumn{10}{c}{\cellcolor{orange!15}\textbf{Proprietary Models}} \\
\midrule
Gemini 2.0 Flash & - & 61.21 & \colorbox{lightgreen}{\textbf{50.97}} & \colorbox{lightblue}{\textbf{72.11}} & \colorbox{lightblue}{\textbf{81.82}} & \colorbox{lightblue}{\textbf{72.48}} & \colorbox{lightgreen}{\textbf{65.85}} & \colorbox{lightgreen}{\textbf{70.83}} & \colorbox{lightblue}{\textbf{65.60}} \\
GPT-4o Audio & - & 53.94 & \colorbox{lightgreen}{\textbf{50.97}} & \colorbox{lightgreen}{\textbf{70.41}} & \colorbox{lightgreen}{\textbf{63.64}} & \colorbox{lightblue}{\textbf{72.48}} & 62.20 & \colorbox{lightblue}{\textbf{75.00}} & \colorbox{lightgreen}{\textbf{63.50}} \\
GPT-4o mini Audio & - & 38.79 & 35.92 & 58.84 & 45.45 & 60.09 & 57.32 & 50.00 & 50.60 \\
Baichuan-Omni-1.5 & 11B & 41.21 & 33.01 & 40.48 & 36.36 & 48.62 & 39.02 & 41.67 & 40.70 \\
\midrule
\multicolumn{10}{c}{\cellcolor{purple!15}\textbf{Large Audio Reasoning Models (LARMs)}} \\
\midrule
Audio-Reasoner & 8.4B & 43.64 & 33.50 & 32.99 & 45.45 & 42.66 & 31.71 & 25.00 & 36.80 \\
Audio-CoT & 8.4B & 35.76 & 25.24 & 34.01 & 9.09 & 30.73 & 30.49 & 37.50 & 31.30 \\
\midrule
\multicolumn{10}{c}{\cellcolor{green!15}\textbf{Large Audio Language Models (LALMs)}} \\
\midrule
Qwen2.5-Omni-3B  & 3B & 53.94 & 46.12 & 53.74 & 36.36 & 60.09 & 57.32 & 58.33 & 53.80 \\
SALAMONN (13B) & 13B & 30.30 & 31.07 & 34.69 & 9.09 & 34.86 & 35.37 & 41.67 & 33.20 \\
SALAMONN (7B) & 7B & 30.91 & 29.61 & 34.35 & 9.09 & 37.61 & 28.05 & 37.50 & 32.80 \\
Audio Flamingo & 2.2B & 32.73 & 21.84 & 24.83 & 18.18 & 30.28 & 24.39 & 25.00 & 26.60 \\
Audio Flamingo 2 & 0.5B & 20.61 & 20.39 & 24.15 & 27.27 & 23.85 & 26.83 & 25.00 & 23.00 \\
Audio Flamingo 2 & 1.5B & 26.67 & 20.87 & 22.79 & 9.09 & 22.94 & 23.17 & 20.83 & 22.90 \\
Audio Flamingo 2 & 3B & 24.85 & 17.48 & 20.75 & 18.18 & 26.61 & 23.17 & 8.33 & 21.90 \\
LTU & 7B & 19.39 & 19.90 & 13.95 & 18.18 & 24.77 & 21.95 & 16.67 & 19.20 \\
LTU-AS & 7B & 20.00 & 14.08 & 19.05 & 9.09 & 20.64 & 28.05 & 12.50 & 19.00 \\
MusiLingo & 7B & 9.09 & 7.28 & 4.08 & 9.09 & 6.88 & 7.32 & 8.33 & 6.60 \\
MU-LLaMa & 7B & 13.94 & 13.59 & 14.97 & 9.09 & 12.39 & 14.63 & 16.67 & 13.90 \\
\midrule
\multicolumn{10}{c}{\cellcolor{red!15}\textbf{Large Reasoning Models (LRMs) + Audio Caption}} \\
\midrule
Caption + OpenAI o3 & - & 49.70 & 41.75 & 63.95 & 36.36 & 60.09 & 52.44 & 54.17 & 54.70 \\
Caption + DeepSeek-R1 & 671B & 46.67 & 49.51 & 62.59 & 45.45 & 58.72 & 56.10 & 54.17 & 55.50 \\
Caption + OpenAI o1 & - & 48.48 & 43.20 & 63.61 & 18.18 & 56.88 & 45.12 & 45.83 & 53.00 \\
Caption + GPT-4o & - & 46.06 & 40.29 & 60.88 & 27.27 & 53.67 & 46.34 & 45.83 & 50.70 \\
Caption + DeepSeek-V3 & 671B & 42.42 & 40.78 & 56.12 & 18.18 & 50.00 & 45.12 & 37.50 & 47.60 \\
\midrule
\multicolumn{10}{c}{\cellcolor{black!10}\textbf{Random Baseline}} \\
\midrule
Random Guess & - & 29.39 & 25.88 & 31.48 & 25.00 & 29.30 & 31.10 & 28.13 & 29.32 \\
\bottomrule
\end{tabular}
}
\end{table*}

\paragraph{The Acoustic-Linguistic Divide.}
The MMAR results first expose a stark divergence between linguistic and acoustic reasoning capabilities across current models. While top proprietary systems show strong performance in tasks dominated by \textit{Speech}, our model, CESAR, achieves the highest scores in the non-linguistic domains of \textit{Sound} (66.06\%) and \textit{Music} (55.83\%). This bifurcation is further illuminated by the ``Caption + LRM" methods; their reliance on text transcripts allows them to perform reasonably well on speech-centric tasks but leaves them unable to compete on acoustic tasks where critical, non-transcribable information is paramount. This demonstrates that advanced audio reasoning is not a monolithic capability and that true progress requires models that can reason over the raw acoustic signal, not just its textual representation.

\paragraph{Superiority of Process-Oriented Reinforcement Learning.}
Second, the benchmark's difficulty serves as a critical test of training methodologies, revealing the limitations of prevalent supervised fine-tuning (SFT) paradigms \citep{con_former,escl,ot_tee_nips23}. Our method, trained with process-oriented reinforcement learning \citep{casa,lbc,cn_rl_entropy,gdi,ppo,pi2}, establishes a commanding lead over other Large Audio Reasoning Models (LARMs) like Audio-Reasoner, with a performance chasm of nearly 26 points (62.70\% vs. 36.80\%). Models trained via SFT on static CoT datasets prove to be brittle, failing to generalize to the complex, multi-hop reasoning required by MMAR. This vast performance gap strongly suggests that robust reasoning skills cannot be effectively learned through imitation alone; they require the interactive, process-focused feedback inherent to our RL framework.

\paragraph{The Frontier of Mixed-Modality Reasoning.}
Finally, MMAR underscores the profound challenge of reasoning over mixed-modality audio streams. Across the board, even top-performing models, including ours and leading proprietary systems, show high variance and struggle for consistent dominance in the ``Mix-" categories. This indicates that while models may handle individual audio types, the compositional understanding and temporal grounding of multiple, overlapping audio sources (e.g., background music, foreground speech, and intermittent sound effects) remains a formidable challenge. This area represents the next clear frontier for the field of audio intelligence.

\clearpage

\end{document}

%% file: math_commands.tex
\usepackage{amsmath,amsfonts,bm}

\def\eqref#1{equation~\ref{#1}}

\def\1{\bm{1}}

\DeclareMathAlphabet{\mathsfit}{\encodingdefault}{\sfdefault}{m}{sl}
\SetMathAlphabet{\mathsfit}{bold}{\encodingdefault}{\sfdefault}{bx}{n}













%% file: iclr2025_conference.bbl
\begin{thebibliography}{43}
\providecommand{\natexlab}[1]{#1}
\providecommand{\url}[1]{\texttt{#1}}
\expandafter\ifx\csname urlstyle\endcsname\relax
  \providecommand{\doi}[1]{doi: #1}\else
  \providecommand{\doi}{doi: \begingroup \urlstyle{rm}\Url}\fi

\bibitem[Balaji et~al.(2023)Balaji, Vunnava, Domingo, Gupta, Gupta, Guest, and Srinivasan]{fla1}
Bharathan Balaji, Venkata Sai~Gargeya Vunnava, Nina Domingo, Shikhar Gupta, Harsh Gupta, Geoffrey Guest, and Aravind Srinivasan.
\newblock Flamingo: Environmental impact factor matching for life cycle assessment with zero-shot machine learning.
\newblock \emph{{ACM} J. Comput. Sustain. Soc.}, 1\penalty0 (2):\penalty0 11:1--11:23, 2023.
\newblock \doi{10.1145/3616385}.
\newblock URL \url{https://doi.org/10.1145/3616385}.

\bibitem[Chu et~al.(2023)Chu, Xu, Zhou, Yang, Zhang, Yan, Zhou, and Zhou]{qwen1_audio}
Yunfei Chu, Jin Xu, Xiaohuan Zhou, Qian Yang, Shiliang Zhang, Zhijie Yan, Chang Zhou, and Jingren Zhou.
\newblock Qwen-audio: Advancing universal audio understanding via unified large-scale audio-language models.
\newblock \emph{CoRR}, abs/2311.07919, 2023.
\newblock \doi{10.48550/ARXIV.2311.07919}.
\newblock URL \url{https://doi.org/10.48550/arXiv.2311.07919}.

\bibitem[Chu et~al.(2024)Chu, Xu, Yang, Wei, Wei, Guo, Leng, Lv, He, Lin, Zhou, and Zhou]{qwen2_audio}
Yunfei Chu, Jin Xu, Qian Yang, Haojie Wei, Xipin Wei, Zhifang Guo, Yichong Leng, Yuanjun Lv, Jinzheng He, Junyang Lin, Chang Zhou, and Jingren Zhou.
\newblock Qwen2-audio technical report.
\newblock \emph{CoRR}, abs/2407.10759, 2024.
\newblock \doi{10.48550/ARXIV.2407.10759}.
\newblock URL \url{https://doi.org/10.48550/arXiv.2407.10759}.

\bibitem[Comanici \& et~al.(2025)Comanici and et~al.]{gemini25}
Gheorghe Comanici and et~al.
\newblock Gemini 2.5: Pushing the frontier with advanced reasoning, multimodality, long context, and next generation agentic capabilities, 2025.
\newblock URL \url{https://arxiv.org/abs/2507.06261}.

\bibitem[DeepSeek{-}AI et~al.(2025)DeepSeek{-}AI, Guo, Yang, Zhang, Song, Zhang, Xu, Zhu, Ma, Wang, Bi, Zhang, Yu, Wu, Wu, Gou, Shao, Li, Gao, Liu, Xue, Wang, Wu, Feng, Lu, Zhao, Deng, Zhang, Ruan, Dai, Chen, Ji, Li, Lin, Dai, Luo, Hao, Chen, Li, Zhang, Bao, Xu, Wang, Ding, Xin, Gao, Qu, Li, Guo, Li, Wang, Chen, Yuan, Qiu, Li, Cai, Ni, Liang, Chen, Dong, Hu, Gao, Guan, Huang, Yu, Wang, Zhang, Zhao, Wang, Zhang, Xu, Xia, Zhang, Zhang, Tang, Li, Wang, Li, Tian, Huang, Zhang, Wang, Chen, Du, Ge, Zhang, Pan, Wang, Chen, Jin, Chen, Lu, Zhou, Chen, Ye, Wang, Yu, Zhou, Pan, and Li]{deepseekai2025deepseekr1}
DeepSeek{-}AI, Daya Guo, Dejian Yang, Haowei Zhang, Junxiao Song, Ruoyu Zhang, Runxin Xu, Qihao Zhu, Shirong Ma, Peiyi Wang, Xiao Bi, Xiaokang Zhang, Xingkai Yu, Yu~Wu, Z.~F. Wu, Zhibin Gou, Zhihong Shao, Zhuoshu Li, Ziyi Gao, Aixin Liu, Bing Xue, Bingxuan Wang, Bochao Wu, Bei Feng, Chengda Lu, Chenggang Zhao, Chengqi Deng, Chenyu Zhang, Chong Ruan, Damai Dai, Deli Chen, Dongjie Ji, Erhang Li, Fangyun Lin, Fucong Dai, Fuli Luo, Guangbo Hao, Guanting Chen, Guowei Li, H.~Zhang, Han Bao, Hanwei Xu, Haocheng Wang, Honghui Ding, Huajian Xin, Huazuo Gao, Hui Qu, Hui Li, Jianzhong Guo, Jiashi Li, Jiawei Wang, Jingchang Chen, Jingyang Yuan, Junjie Qiu, Junlong Li, J.~L. Cai, Jiaqi Ni, Jian Liang, Jin Chen, Kai Dong, Kai Hu, Kaige Gao, Kang Guan, Kexin Huang, Kuai Yu, Lean Wang, Lecong Zhang, Liang Zhao, Litong Wang, Liyue Zhang, Lei Xu, Leyi Xia, Mingchuan Zhang, Minghua Zhang, Minghui Tang, Meng Li, Miaojun Wang, Mingming Li, Ning Tian, Panpan Huang, Peng Zhang, Qiancheng Wang, Qinyu Chen, Qiushi Du, Ruiqi Ge,
  Ruisong Zhang, Ruizhe Pan, Runji Wang, R.~J. Chen, R.~L. Jin, Ruyi Chen, Shanghao Lu, Shangyan Zhou, Shanhuang Chen, Shengfeng Ye, Shiyu Wang, Shuiping Yu, Shunfeng Zhou, Shuting Pan, and S.~S. Li.
\newblock Deepseek-r1: Incentivizing reasoning capability in llms via reinforcement learning.
\newblock \emph{CoRR}, abs/2501.12948, 2025.
\newblock \doi{10.48550/ARXIV.2501.12948}.
\newblock URL \url{https://doi.org/10.48550/arXiv.2501.12948}.

\bibitem[Deshmukh et~al.(2023)Deshmukh, Elizalde, Singh, and Wang]{deshmukh2023pengi}
Soham Deshmukh, Benjamin Elizalde, Rita Singh, and Huaming Wang.
\newblock Pengi: An audio language model for audio tasks.
\newblock In Alice Oh, Tristan Naumann, Amir Globerson, Kate Saenko, Moritz Hardt, and Sergey Levine (eds.), \emph{Advances in Neural Information Processing Systems 36: Annual Conference on Neural Information Processing Systems 2023, NeurIPS 2023, New Orleans, LA, USA, December 10 - 16, 2023}, 2023.
\newblock URL \url{http://papers.nips.cc/paper\_files/paper/2023/hash/3a2e5889b4bbef997ddb13b55d5acf77-Abstract-Conference.html}.

\bibitem[Elizalde et~al.(2023)Elizalde, Deshmukh, Ismail, and Wang]{elizalde2022clap}
Benjamin Elizalde, Soham Deshmukh, Mahmoud~Al Ismail, and Huaming Wang.
\newblock {CLAP} learning audio concepts from natural language supervision.
\newblock In \emph{{IEEE} International Conference on Acoustics, Speech and Signal Processing {ICASSP} 2023, Rhodes Island, Greece, June 4-10, 2023}, pp.\  1--5. {IEEE}, 2023.
\newblock \doi{10.1109/ICASSP49357.2023.10095889}.
\newblock URL \url{https://doi.org/10.1109/ICASSP49357.2023.10095889}.

\bibitem[Fan(2021)]{atari_review}
Jiajun Fan.
\newblock A review for deep reinforcement learning in atari: Benchmarks, challenges, and solutions.
\newblock \emph{CoRR}, abs/2112.04145, 2021.
\newblock URL \url{https://arxiv.org/abs/2112.04145}.

\bibitem[Fan \& Xiao(2022)Fan and Xiao]{gdi}
Jiajun Fan and Changnan Xiao.
\newblock Generalized data distribution iteration.
\newblock In Kamalika Chaudhuri, Stefanie Jegelka, Le~Song, Csaba Szepesv{\'{a}}ri, Gang Niu, and Sivan Sabato (eds.), \emph{International Conference on Machine Learning, {ICML} 2022, 17-23 July 2022, Baltimore, Maryland, {USA}}, volume 162 of \emph{Proceedings of Machine Learning Research}, pp.\  6103--6184. {PMLR}, 2022.
\newblock URL \url{https://proceedings.mlr.press/v162/fan22c.html}.

\bibitem[Fan et~al.(2020)Fan, Ba, Guo, and Hao]{pi2}
Jiajun Fan, He~Ba, Xian Guo, and Jianye Hao.
\newblock Critic {PI2:} master continuous planning via policy improvement with path integrals and deep actor-critic reinforcement learning.
\newblock \emph{CoRR}, abs/2011.06752, 2020.
\newblock URL \url{https://arxiv.org/abs/2011.06752}.

\bibitem[Fan et~al.(2023)Fan, Zhuang, Liu, Hao, Wang, Zhu, Wang, and Xia]{lbc}
Jiajun Fan, Yuzheng Zhuang, Yuecheng Liu, Jianye Hao, Bin Wang, Jiangcheng Zhu, Hao Wang, and Shu{-}Tao Xia.
\newblock Learnable behavior control: Breaking atari human world records via sample-efficient behavior selection.
\newblock In \emph{The Eleventh International Conference on Learning Representations, {ICLR} 2023, Kigali, Rwanda, May 1-5, 2023}. OpenReview.net, 2023.
\newblock URL \url{https://openreview.net/forum?id=FeWvD0L\_a4}.

\bibitem[Fan et~al.(2025)Fan, Shen, Cheng, Chen, Liang, and Liu]{iclr_rwr}
Jiajun Fan, Shuaike Shen, Chaoran Cheng, Yuxin Chen, Chumeng Liang, and Ge~Liu.
\newblock Online reward-weighted fine-tuning of flow matching with wasserstein regularization.
\newblock In \emph{The Thirteenth International Conference on Learning Representations, {ICLR} 2025, Singapore, April 24-28, 2025}. OpenReview.net, 2025.
\newblock URL \url{https://openreview.net/forum?id=2IoFFexvuw}.

\bibitem[Ghosh et~al.(2025)Ghosh, Kong, Kumar, Sakshi, Kim, Ping, Valle, Manocha, and Catanzaro]{fla2}
Sreyan Ghosh, Zhifeng Kong, Sonal Kumar, S.~Sakshi, Jaehyeon Kim, Wei Ping, Rafael Valle, Dinesh Manocha, and Bryan Catanzaro.
\newblock Audio flamingo 2: An audio-language model with long-audio understanding and expert reasoning abilities.
\newblock \emph{CoRR}, abs/2503.03983, 2025.
\newblock \doi{10.48550/ARXIV.2503.03983}.
\newblock URL \url{https://doi.org/10.48550/arXiv.2503.03983}.

\bibitem[Goel et~al.(2025{\natexlab{a}})Goel, Ghosh, Kim, Kumar, Kong, gil Lee, Yang, Duraiswami, Manocha, Valle, and Catanzaro]{kong2025audioflamingo3}
Arushi Goel, Sreyan Ghosh, Jaehyeon Kim, Sonal Kumar, Zhifeng Kong, Sang gil Lee, Chao-Han~Huck Yang, Ramani Duraiswami, Dinesh Manocha, Rafael Valle, and Bryan Catanzaro.
\newblock Audio flamingo 3: Advancing audio intelligence with fully open large audio language models, 2025{\natexlab{a}}.
\newblock URL \url{https://arxiv.org/abs/2507.08128}.

\bibitem[Goel et~al.(2025{\natexlab{b}})Goel, Ghosh, Kim, Kumar, Kong, Lee, Yang, Duraiswami, Manocha, Valle, and Catanzaro]{flam3_nvidia}
Arushi Goel, Sreyan Ghosh, Jaehyeon Kim, Sonal Kumar, Zhifeng Kong, Sang{-}gil Lee, Chao{-}Han~Huck Yang, Ramani Duraiswami, Dinesh Manocha, Rafael Valle, and Bryan Catanzaro.
\newblock Audio flamingo 3: Advancing audio intelligence with fully open large audio language models.
\newblock \emph{CoRR}, abs/2507.08128, 2025{\natexlab{b}}.
\newblock \doi{10.48550/ARXIV.2507.08128}.
\newblock URL \url{https://doi.org/10.48550/arXiv.2507.08128}.

\bibitem[Gong et~al.(2024)Gong, Luo, Liu, Karlinsky, and Glass]{gong2023ltu}
Yuan Gong, Hongyin Luo, Alexander~H. Liu, Leonid Karlinsky, and James~R. Glass.
\newblock Listen, think, and understand.
\newblock In \emph{The Twelfth International Conference on Learning Representations, {ICLR} 2024, Vienna, Austria, May 7-11, 2024}. OpenReview.net, 2024.
\newblock URL \url{https://openreview.net/forum?id=nBZBPXdJlC}.

\bibitem[Hurst et~al.(2024)Hurst, Lerer, Goucher, Perelman, Ramesh, Clark, Ostrow, Welihinda, Hayes, Radford, Madry, Baker{-}Whitcomb, Beutel, Borzunov, Carney, Chow, Kirillov, Nichol, Paino, Renzin, Passos, Kirillov, Christakis, Conneau, Kamali, Jabri, Moyer, Tam, Crookes, Tootoonchian, Kumar, Vallone, Karpathy, Braunstein, Cann, Codispoti, Galu, Kondrich, Tulloch, Mishchenko, Baek, Jiang, Pelisse, Woodford, Gosalia, Dhar, Pantuliano, Nayak, Oliver, Zoph, Ghorbani, Leimberger, Rossen, Sokolowsky, Wang, Zweig, Hoover, Samic, McGrew, Spero, Giertler, Cheng, Lightcap, Walkin, Quinn, Guarraci, Hsu, Kellogg, Eastman, Lugaresi, Wainwright, Bassin, Hudson, Chu, Nelson, Li, Shern, Conger, Barette, Voss, Ding, Lu, Zhang, Beaumont, Hallacy, Koch, Gibson, Kim, Choi, McLeavey, Hesse, Fischer, Winter, Czarnecki, Jarvis, Wei, Koumouzelis, and Sherburn]{gpt4oaudio}
Aaron Hurst, Adam Lerer, Adam~P. Goucher, Adam Perelman, Aditya Ramesh, Aidan Clark, AJ~Ostrow, Akila Welihinda, Alan Hayes, Alec Radford, Aleksander Madry, Alex Baker{-}Whitcomb, Alex Beutel, Alex Borzunov, Alex Carney, Alex Chow, Alex Kirillov, Alex Nichol, Alex Paino, Alex Renzin, Alex~Tachard Passos, Alexander Kirillov, Alexi Christakis, Alexis Conneau, Ali Kamali, Allan Jabri, Allison Moyer, Allison Tam, Amadou Crookes, Amin Tootoonchian, Ananya Kumar, Andrea Vallone, Andrej Karpathy, Andrew Braunstein, Andrew Cann, Andrew Codispoti, Andrew Galu, Andrew Kondrich, Andrew Tulloch, Andrey Mishchenko, Angela Baek, Angela Jiang, Antoine Pelisse, Antonia Woodford, Anuj Gosalia, Arka Dhar, Ashley Pantuliano, Avi Nayak, Avital Oliver, Barret Zoph, Behrooz Ghorbani, Ben Leimberger, Ben Rossen, Ben Sokolowsky, Ben Wang, Benjamin Zweig, Beth Hoover, Blake Samic, Bob McGrew, Bobby Spero, Bogo Giertler, Bowen Cheng, Brad Lightcap, Brandon Walkin, Brendan Quinn, Brian Guarraci, Brian Hsu, Bright Kellogg, Brydon
  Eastman, Camillo Lugaresi, Carroll~L. Wainwright, Cary Bassin, Cary Hudson, Casey Chu, Chad Nelson, Chak Li, Chan~Jun Shern, Channing Conger, Charlotte Barette, Chelsea Voss, Chen Ding, Cheng Lu, Chong Zhang, Chris Beaumont, Chris Hallacy, Chris Koch, Christian Gibson, Christina Kim, Christine Choi, Christine McLeavey, Christopher Hesse, Claudia Fischer, Clemens Winter, Coley Czarnecki, Colin Jarvis, Colin Wei, Constantin Koumouzelis, and Dane Sherburn.
\newblock Gpt-4o system card.
\newblock \emph{CoRR}, abs/2410.21276, 2024.
\newblock \doi{10.48550/ARXIV.2410.21276}.
\newblock URL \url{https://doi.org/10.48550/arXiv.2410.21276}.

\bibitem[Jaech et~al.(2024)Jaech, Kalai, Lerer, Richardson, El{-}Kishky, Low, Helyar, Madry, Beutel, Carney, Iftimie, Karpenko, Passos, Neitz, Prokofiev, Wei, Tam, Bennett, Kumar, Saraiva, Vallone, Duberstein, Kondrich, Mishchenko, Applebaum, Jiang, Nair, Zoph, Ghorbani, Rossen, Sokolowsky, Barak, McGrew, Minaiev, Hao, Baker, Houghton, McKinzie, Eastman, Lugaresi, Bassin, Hudson, Li, de~Bourcy, Voss, Shen, Zhang, Koch, Orsinger, Hesse, Fischer, Chan, Roberts, Kappler, Levy, Selsam, Dohan, Farhi, Mely, Robinson, Tsipras, Li, Oprica, Freeman, Zhang, Wong, Proehl, Cheung, Mitchell, Wallace, Ritter, Mays, Wang, Such, Raso, Leoni, Tsimpourlas, Song, von Lohmann, Sulit, Salmon, Parascandolo, Chabot, Zhao, Brockman, Leclerc, Salman, Bao, Sheng, Andrin, Bagherinezhad, Ren, Lightman, Chung, Kivlichan, O'Connell, Osband, Gilaberte, and Akkaya]{openai2024o1}
Aaron Jaech, Adam Kalai, Adam Lerer, Adam Richardson, Ahmed El{-}Kishky, Aiden Low, Alec Helyar, Aleksander Madry, Alex Beutel, Alex Carney, Alex Iftimie, Alex Karpenko, Alex~Tachard Passos, Alexander Neitz, Alexander Prokofiev, Alexander Wei, Allison Tam, Ally Bennett, Ananya Kumar, Andre Saraiva, Andrea Vallone, Andrew Duberstein, Andrew Kondrich, Andrey Mishchenko, Andy Applebaum, Angela Jiang, Ashvin Nair, Barret Zoph, Behrooz Ghorbani, Ben Rossen, Benjamin Sokolowsky, Boaz Barak, Bob McGrew, Borys Minaiev, Botao Hao, Bowen Baker, Brandon Houghton, Brandon McKinzie, Brydon Eastman, Camillo Lugaresi, Cary Bassin, Cary Hudson, Chak~Ming Li, Charles de~Bourcy, Chelsea Voss, Chen Shen, Chong Zhang, Chris Koch, Chris Orsinger, Christopher Hesse, Claudia Fischer, Clive Chan, Dan Roberts, Daniel Kappler, Daniel Levy, Daniel Selsam, David Dohan, David Farhi, David Mely, David Robinson, Dimitris Tsipras, Doug Li, Dragos Oprica, Eben Freeman, Eddie Zhang, Edmund Wong, Elizabeth Proehl, Enoch Cheung, Eric Mitchell,
  Eric Wallace, Erik Ritter, Evan Mays, Fan Wang, Felipe~Petroski Such, Filippo Raso, Florencia Leoni, Foivos Tsimpourlas, Francis Song, Fred von Lohmann, Freddie Sulit, Geoff Salmon, Giambattista Parascandolo, Gildas Chabot, Grace Zhao, Greg Brockman, Guillaume Leclerc, Hadi Salman, Haiming Bao, Hao Sheng, Hart Andrin, Hessam Bagherinezhad, Hongyu Ren, Hunter Lightman, Hyung~Won Chung, Ian Kivlichan, Ian O'Connell, Ian Osband, Ignasi~Clavera Gilaberte, and Ilge Akkaya.
\newblock Openai o1 system card.
\newblock \emph{CoRR}, abs/2412.16720, 2024.
\newblock \doi{10.48550/ARXIV.2412.16720}.
\newblock URL \url{https://doi.org/10.48550/arXiv.2412.16720}.

\bibitem[KimiTeam et~al.(2025)KimiTeam, Ding, Ju, Leng, Liu, Liu, Shang, Shen, Song, Tan, Tang, Wang, Wei, Xin, Xu, Yu, Zhang, Zhou, Charles, Chen, Chen, Du, He, Hu, Lai, Li, Liu, Sun, Wang, Wang, Wu, Wu, Yang, Yang, Yang, Yang, Yin, Yuan, Zhang, and Zhou]{kimi}
KimiTeam, Ding Ding, Zeqian Ju, Yichong Leng, Songxiang Liu, Tong Liu, Zeyu Shang, Kai Shen, Wei Song, Xu~Tan, Heyi Tang, Zhengtao Wang, Chu Wei, Yifei Xin, Xinran Xu, Jianwei Yu, Yutao Zhang, Xinyu Zhou, Y.~Charles, Jun Chen, Yanru Chen, Yulun Du, Weiran He, Zhenxing Hu, Guokun Lai, Qingcheng Li, Yangyang Liu, Weidong Sun, Jianzhou Wang, Yuzhi Wang, Yuefeng Wu, Yuxin Wu, Dongchao Yang, Hao Yang, Ying Yang, Zhilin Yang, Aoxiong Yin, Ruibin Yuan, Yutong Zhang, and Zaida Zhou.
\newblock Kimi-audio technical report.
\newblock \emph{CoRR}, abs/2504.18425, 2025.
\newblock \doi{10.48550/ARXIV.2504.18425}.
\newblock URL \url{https://doi.org/10.48550/arXiv.2504.18425}.

\bibitem[Kong et~al.(2024)Kong, Goel, Badlani, Ping, Valle, and Catanzaro]{kong2024audio}
Zhifeng Kong, Arushi Goel, Rohan Badlani, Wei Ping, Rafael Valle, and Bryan Catanzaro.
\newblock Audio flamingo: {A} novel audio language model with few-shot learning and dialogue abilities.
\newblock In \emph{Forty-first International Conference on Machine Learning, {ICML} 2024, Vienna, Austria, July 21-27, 2024}. OpenReview.net, 2024.
\newblock URL \url{https://openreview.net/forum?id=WYi3WKZjYe}.

\bibitem[Li et~al.(2025{\natexlab{a}})Li, Liu, Dinkel, Niu, Zhang, and Luan]{r1_aqa}
Gang Li, Jizhong Liu, Heinrich Dinkel, Yadong Niu, Junbo Zhang, and Jian Luan.
\newblock Reinforcement learning outperforms supervised fine-tuning: A case study on audio question answering.
\newblock \emph{arXiv preprint arXiv:2503.11197}, 2025{\natexlab{a}}.
\newblock URL \url{https://github.com/xiaomi-research/r1-aqa; https://huggingface.co/mispeech/r1-aqa}.

\bibitem[Li et~al.(2024)Li, Tang, Meng, Fan, Chai, Ma, Wang, and Zhu]{prance}
Ye~Li, Chen Tang, Yuan Meng, Jiajun Fan, Zenghao Chai, Xinzhu Ma, Zhi Wang, and Wenwu Zhu.
\newblock {PRANCE:} joint token-optimization and structural channel-pruning for adaptive vit inference.
\newblock \emph{CoRR}, abs/2407.05010, 2024.
\newblock \doi{10.48550/ARXIV.2407.05010}.
\newblock URL \url{https://doi.org/10.48550/arXiv.2407.05010}.

\bibitem[Li et~al.(2025{\natexlab{b}})Li, Meng, Sun, Ji, Tang, Fan, Ma, Xia, Wang, and Zhu]{sp_vla}
Ye~Li, Yuan Meng, Zewen Sun, Kangye Ji, Chen Tang, Jiajun Fan, Xinzhu Ma, Shutao Xia, Zhi Wang, and Wenwu Zhu.
\newblock {SP-VLA:} {A} joint model scheduling and token pruning approach for {VLA} model acceleration.
\newblock \emph{CoRR}, abs/2506.12723, 2025{\natexlab{b}}.
\newblock \doi{10.48550/ARXIV.2506.12723}.
\newblock URL \url{https://doi.org/10.48550/arXiv.2506.12723}.

\bibitem[Ma et~al.(2025{\natexlab{a}})Ma, Chen, Wang, Chng, and Chen]{ma2025audiocot}
Ziyang Ma, Zhuo Chen, Yuping Wang, Eng~Siong Chng, and Xie Chen.
\newblock Audio-cot: Exploring chain-of-thought reasoning in large audio language model.
\newblock \emph{CoRR}, abs/2501.07246, 2025{\natexlab{a}}.
\newblock \doi{10.48550/ARXIV.2501.07246}.
\newblock URL \url{https://doi.org/10.48550/arXiv.2501.07246}.

\bibitem[Ma et~al.(2025{\natexlab{b}})Ma, Ma, Zhu, Yang, Chao, Xu, Chen, Chen, Chen, Cong, Li, Li, Li, Li, Li, Lian, Liang, Liu, Niu, Wang, Wang, Wang, Wu, Yang, Yu, Yuan, Zheng, Zhou, Zhu, Xue, Benetos, Yu, Siong, and Chen]{ghosh2025mmar}
Ziyang Ma, Yinghao Ma, Yanqiao Zhu, Chen Yang, Yi{-}Wen Chao, Ruiyang Xu, Wenxi Chen, Yuanzhe Chen, Zhuo Chen, Jian Cong, Kai Li, Keliang Li, Siyou Li, Xinfeng Li, Xiquan Li, Zheng Lian, Yuzhe Liang, Minghao Liu, Zhikang Niu, Tianrui Wang, Yuping Wang, Yuxuan Wang, Yihao Wu, Guanrou Yang, Jianwei Yu, Ruibin Yuan, Zhisheng Zheng, Ziya Zhou, Haina Zhu, Wei Xue, Emmanouil Benetos, Kai Yu, Chng~Eng Siong, and Xie Chen.
\newblock {MMAR:} {A} challenging benchmark for deep reasoning in speech, audio, music, and their mix.
\newblock \emph{CoRR}, abs/2505.13032, 2025{\natexlab{b}}.
\newblock \doi{10.48550/ARXIV.2505.13032}.
\newblock URL \url{https://doi.org/10.48550/arXiv.2505.13032}.

\bibitem[Sakshi et~al.(2025)Sakshi, Tyagi, Kumar, Seth, Selvakumar, Nieto, Duraiswami, Ghosh, and Manocha]{sakshi2024mmau}
S.~Sakshi, Utkarsh Tyagi, Sonal Kumar, Ashish Seth, Ramaneswaran Selvakumar, Oriol Nieto, Ramani Duraiswami, Sreyan Ghosh, and Dinesh Manocha.
\newblock {MMAU:} {A} massive multi-task audio understanding and reasoning benchmark.
\newblock In \emph{The Thirteenth International Conference on Learning Representations, {ICLR} 2025, Singapore, April 24-28, 2025}. OpenReview.net, 2025.
\newblock URL \url{https://openreview.net/forum?id=TeVAZXr3yv}.

\bibitem[Schulman et~al.(2017)Schulman, Wolski, Dhariwal, Radford, and Klimov]{ppo}
John Schulman, Filip Wolski, Prafulla Dhariwal, Alec Radford, and Oleg Klimov.
\newblock Proximal policy optimization algorithms.
\newblock \emph{CoRR}, abs/1707.06347, 2017.
\newblock URL \url{http://arxiv.org/abs/1707.06347}.

\bibitem[Shao et~al.(2024)Shao, Wang, Zhu, Xu, Song, Zhang, Li, Wu, and Guo]{shao2024grpo}
Zhihong Shao, Peiyi Wang, Qihao Zhu, Runxin Xu, Junxiao Song, Mingchuan Zhang, Y.~K. Li, Y.~Wu, and Daya Guo.
\newblock Deepseekmath: Pushing the limits of mathematical reasoning in open language models.
\newblock \emph{CoRR}, abs/2402.03300, 2024.
\newblock \doi{10.48550/ARXIV.2402.03300}.
\newblock URL \url{https://doi.org/10.48550/arXiv.2402.03300}.

\bibitem[Tang et~al.(2024)Tang, Yu, Sun, Chen, Tan, Li, Lu, Ma, and Zhang]{tang2023salmonn}
Changli Tang, Wenyi Yu, Guangzhi Sun, Xianzhao Chen, Tian Tan, Wei Li, Lu~Lu, Zejun Ma, and Chao Zhang.
\newblock {SALMONN:} towards generic hearing abilities for large language models.
\newblock In \emph{The Twelfth International Conference on Learning Representations, {ICLR} 2024, Vienna, Austria, May 7-11, 2024}. OpenReview.net, 2024.
\newblock URL \url{https://openreview.net/forum?id=14rn7HpKVk}.

\bibitem[Wang et~al.(2025{\natexlab{a}})Wang, Wu, Li, Yang, Chen, Zhang, and Meng]{mmsu}
Dingdong Wang, Jincenzi Wu, Junan Li, Dongchao Yang, Xueyuan Chen, Tianhua Zhang, and Helen Meng.
\newblock {MMSU:} {A} massive multi-task spoken language understanding and reasoning benchmark.
\newblock \emph{CoRR}, abs/2506.04779, 2025{\natexlab{a}}.
\newblock \doi{10.48550/ARXIV.2506.04779}.
\newblock URL \url{https://doi.org/10.48550/arXiv.2506.04779}.

\bibitem[Wang et~al.(2022)Wang, Chen, Fan, Huang, Liu, and Liu]{escl}
Hao Wang, Zhichao Chen, Jiajun Fan, Yuxin Huang, Weiming Liu, and Xinggao Liu.
\newblock Entire space counterfactual learning: Tuning, analytical properties and industrial applications.
\newblock \emph{CoRR}, abs/2210.11039, 2022.
\newblock \doi{10.48550/ARXIV.2210.11039}.
\newblock URL \url{https://doi.org/10.48550/arXiv.2210.11039}.

\bibitem[Wang et~al.(2023{\natexlab{a}})Wang, Fan, Chen, Li, Liu, Liu, Dai, Wang, Dong, and Tang]{ot_tee_nips23}
Hao Wang, Jiajun Fan, Zhichao Chen, Haoxuan Li, Weiming Liu, Tianqiao Liu, Quanyu Dai, Yichao Wang, Zhenhua Dong, and Ruiming Tang.
\newblock Optimal transport for treatment effect estimation.
\newblock In Alice Oh, Tristan Naumann, Amir Globerson, Kate Saenko, Moritz Hardt, and Sergey Levine (eds.), \emph{Advances in Neural Information Processing Systems 36: Annual Conference on Neural Information Processing Systems 2023, NeurIPS 2023, New Orleans, LA, USA, December 10 - 16, 2023}, 2023{\natexlab{a}}.
\newblock URL \url{http://papers.nips.cc/paper\_files/paper/2023/hash/1160e7f31d0a74abbbe1bbf7924b949c-Abstract-Conference.html}.

\bibitem[Wang et~al.(2023{\natexlab{b}})Wang, Lian, Wu, Li, Fan, Xu, Li, and Xie]{con_former}
Hao Wang, Jianxun Lian, Mingqi Wu, Haoxuan Li, Jiajun Fan, Wanyue Xu, Chaozhuo Li, and Xing Xie.
\newblock Convformer: Revisiting transformer for sequential user modeling.
\newblock \emph{CoRR}, abs/2308.02925, 2023{\natexlab{b}}.
\newblock \doi{10.48550/ARXIV.2308.02925}.
\newblock URL \url{https://doi.org/10.48550/arXiv.2308.02925}.

\bibitem[Wang et~al.(2025{\natexlab{b}})Wang, Fan, Guo, Nguyen, Ji, and Liu]{pzero}
Ziwen Wang, Jiajun Fan, Ruihan Guo, Thao Nguyen, Heng Ji, and Ge~Liu.
\newblock Proteinzero: Self-improving protein generation via online reinforcement learning.
\newblock \emph{CoRR}, abs/2506.07459, 2025{\natexlab{b}}.
\newblock \doi{10.48550/ARXIV.2506.07459}.
\newblock URL \url{https://doi.org/10.48550/arXiv.2506.07459}.

\bibitem[Wang et~al.(2025{\natexlab{c}})Wang, Fan, Nguyen, Ji, and Liu]{varcon}
Ziwen Wang, Jiajun Fan, Thao Nguyen, Heng Ji, and Ge~Liu.
\newblock Variational supervised contrastive learning.
\newblock \emph{CoRR}, abs/2506.07413, 2025{\natexlab{c}}.
\newblock \doi{10.48550/ARXIV.2506.07413}.
\newblock URL \url{https://doi.org/10.48550/arXiv.2506.07413}.

\bibitem[Wei et~al.(2022)Wei, Wang, Schuurmans, Bosma, Ichter, Xia, Chi, Le, and Zhou]{wei2022chain}
Jason Wei, Xuezhi Wang, Dale Schuurmans, Maarten Bosma, Brian Ichter, Fei Xia, Ed~H. Chi, Quoc~V. Le, and Denny Zhou.
\newblock Chain-of-thought prompting elicits reasoning in large language models.
\newblock In Sanmi Koyejo, S.~Mohamed, A.~Agarwal, Danielle Belgrave, K.~Cho, and A.~Oh (eds.), \emph{Advances in Neural Information Processing Systems 35: Annual Conference on Neural Information Processing Systems 2022, NeurIPS 2022, New Orleans, LA, USA, November 28 - December 9, 2022}, 2022.
\newblock URL \url{http://papers.nips.cc/paper\_files/paper/2022/hash/9d5609613524ecf4f15af0f7b31abca4-Abstract-Conference.html}.

\bibitem[Wu et~al.(2022)Wu, Seetharaman, Kumar, and Bello]{wu2021wav2clip}
Ho{-}Hsiang Wu, Prem Seetharaman, Kundan Kumar, and Juan~Pablo Bello.
\newblock Wav2clip: Learning robust audio representations from clip.
\newblock In \emph{{IEEE} International Conference on Acoustics, Speech and Signal Processing, {ICASSP} 2022, Virtual and Singapore, 23-27 May 2022}, pp.\  4563--4567. {IEEE}, 2022.
\newblock \doi{10.1109/ICASSP43922.2022.9747669}.
\newblock URL \url{https://doi.org/10.1109/ICASSP43922.2022.9747669}.

\bibitem[Xiao et~al.(2021{\natexlab{a}})Xiao, Shi, Fan, and Deng]{casa}
Changnan Xiao, Haosen Shi, Jiajun Fan, and Shihong Deng.
\newblock {CASA:} {A} bridge between gradient of policy improvement and policy evaluation.
\newblock \emph{CoRR}, abs/2105.03923, 2021{\natexlab{a}}.
\newblock URL \url{https://arxiv.org/abs/2105.03923}.

\bibitem[Xiao et~al.(2021{\natexlab{b}})Xiao, Shi, Fan, and Deng]{cn_rl_entropy}
Changnan Xiao, Haosen Shi, Jiajun Fan, and Shihong Deng.
\newblock An entropy regularization free mechanism for policy-based reinforcement learning.
\newblock \emph{CoRR}, abs/2106.00707, 2021{\natexlab{b}}.
\newblock URL \url{https://arxiv.org/abs/2106.00707}.

\bibitem[Xie et~al.(2025)Xie, Lin, Liu, Wu, Yan, and Miao]{liu2025audioreasoner}
Zhifei Xie, Mingbao Lin, Zihang Liu, Pengcheng Wu, Shuicheng Yan, and Chunyan Miao.
\newblock Audio-reasoner: Improving reasoning capability in large audio language models.
\newblock \emph{CoRR}, abs/2503.02318, 2025.
\newblock \doi{10.48550/ARXIV.2503.02318}.
\newblock URL \url{https://doi.org/10.48550/arXiv.2503.02318}.

\bibitem[Xu et~al.(2025)Xu, Guo, He, Hu, He, Bai, Chen, Wang, Fan, Dang, Zhang, Wang, Chu, and Lin]{xu2025qwen25omni}
Jin Xu, Zhifang Guo, Jinzheng He, Hangrui Hu, Ting He, Shuai Bai, Keqin Chen, Jialin Wang, Yang Fan, Kai Dang, Bin Zhang, Xiong Wang, Yunfei Chu, and Junyang Lin.
\newblock Qwen2.5-omni technical report.
\newblock \emph{CoRR}, abs/2503.20215, 2025.
\newblock \doi{10.48550/ARXIV.2503.20215}.
\newblock URL \url{https://doi.org/10.48550/arXiv.2503.20215}.

\bibitem[Yang et~al.(2022)Yang, Wang, Duan, Chen, Hou, Jin, and Zhu]{yang2022avqa}
Pinci Yang, Xin Wang, Xuguang Duan, Hong Chen, Runze Hou, Cong Jin, and Wenwu Zhu.
\newblock Avqa: A dataset for audio-visual question answering on videos.
\newblock In \emph{Proceedings of the 30th ACM International Conference on Multimedia}, pp.\  3480--3491, 2022.

\bibitem[Zhao et~al.(2025)Zhao, Guo, Wen, Xiang, and Zou]{ke_omni_r}
Shuaijiang Zhao, Tingwei Guo, Cheng Wen, Bajian Xiang, and Wei Zou.
\newblock Ke-omni-r: Achieving advanced audio reasoning with a concise 50-words think process.
\newblock \url{https://github.com/shuaijiang/Ke-Omni-R}, 2025.

\end{thebibliography}
